%% file: sparseInertialPoser.tex
\documentclass{egpubl}
\usepackage{eg2017}

\ConferencePaper        
 \electronicVersion 

\ifpdf \usepackage[pdftex]{graphicx} \pdfcompresslevel=9
\else \usepackage[dvips]{graphicx} \fi

\PrintedOrElectronic

\usepackage{t1enc,dfadobe}

\usepackage{egweblnk}
\usepackage{cite}

\title{Sparse Inertial Poser:\\ Automatic 3D Human Pose Estimation from Sparse IMUs}

\author[v. Marcard et al.]
       {T. von Marcard$^{1}$ \, B. Rosenhahn$^{1}$ \, M.\,J. Black$^{2}$ \, G. Pons-Moll$^{2}$
        \\
         $^1$Institut f{\"u}r Informationsverarbeitung (TNT), Leibniz-Universit{\"a}t Hannover, Germany\\
         $^2$Max Planck Institute for Intelligent Systems, T{\"u}bingen, Germany
       }

\usepackage{amsmath,amssymb} 
\graphicspath{{./images/}}
\usepackage{import}
\usepackage{subfigure}

\begin{document}

\teaser{
\centering
\includegraphics[height=4.5 cm]{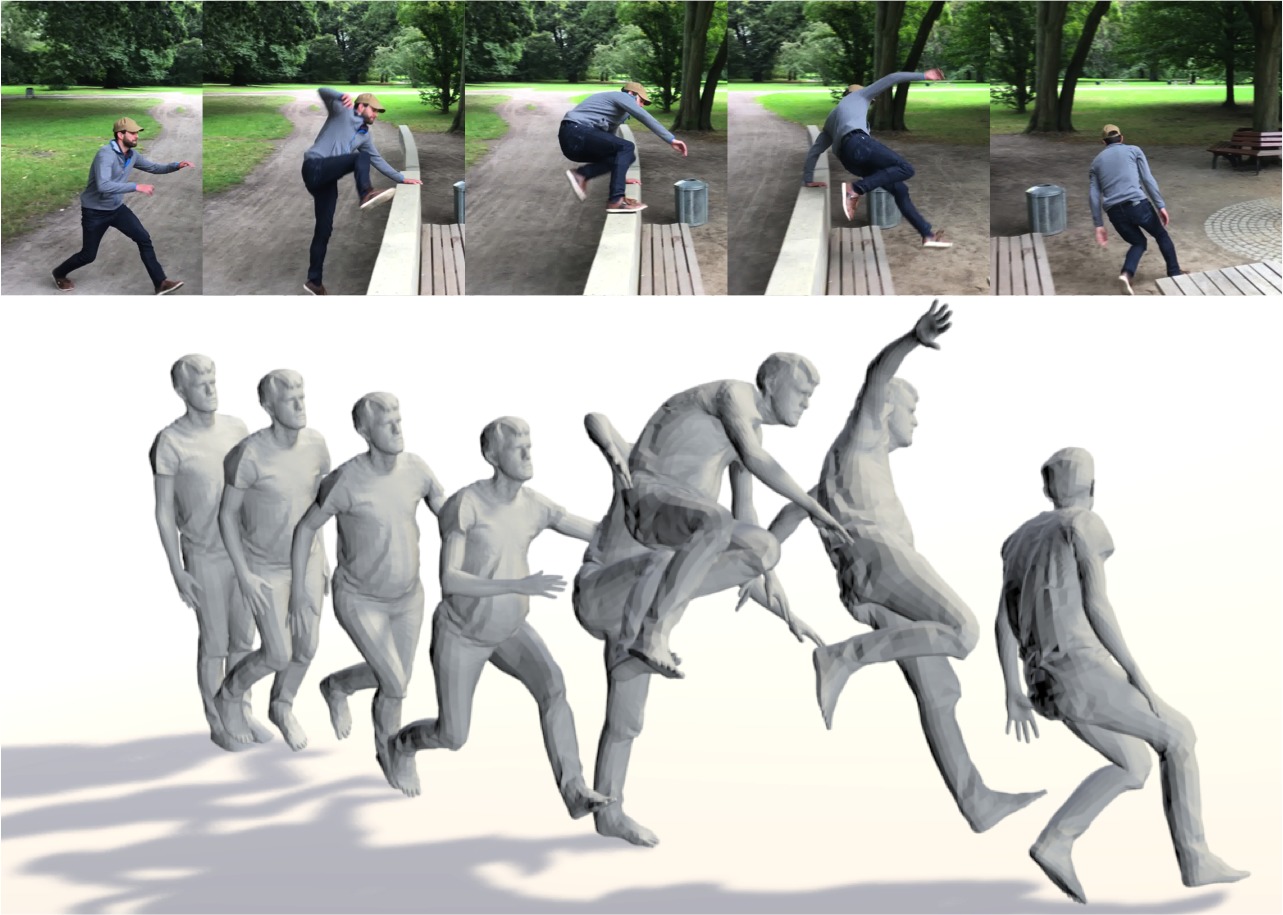}
\hspace{1 cm}
\includegraphics[height=4.5 cm]{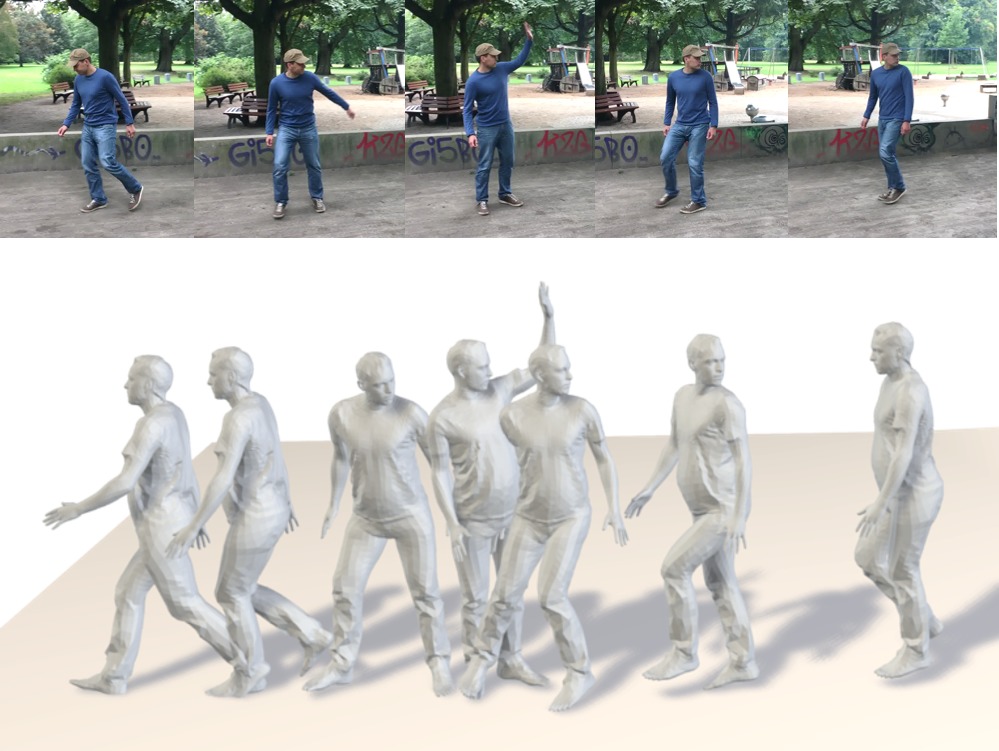}
\caption{ Unconstrained motion capture using our new Sparse Inertial Poser (SIP).
  With as few as 6 IMUs attached to the body, we recover the full
  pose of the subject. The key idea that makes this possible is to
  optimise all the poses of a statistical body model for all the
  frames in the sequence jointly to fit the orientation and
  acceleration measurements captured by the IMUs. Images are shown for
  reference but are not used during the optimisation. }
\label{fig:teaser}
}
\maketitle
\input{notation.tex}
\begin{abstract}
We address the problem of making human motion capture in the wild more
practical by using a small set of inertial sensors attached to the body. Since
the problem is heavily under-constrained, previous methods either use
a large number of sensors, which is intrusive, or they require
additional video input. 
We take a different approach and constrain the problem by: (i) making
use of a realistic statistical body model that includes anthropometric constraints and (ii) using a joint optimization framework to fit the model to
orientation and acceleration measurements over multiple frames. 
The resulting tracker Sparse Inertial Poser (SIP) enables 3D human pose estimation using only 6 sensors (attached to the wrists, lower legs, back and
head) and works for arbitrary human motions.
Experiments on the recently released TNT15 dataset show that, using the same number of sensors, SIP achieves higher accuracy than the dataset baseline without using any video data. 
We further demonstrate the effectiveness of SIP on newly recorded challenging motions in outdoor scenarios such as climbing or jumping over a wall. 
\end{abstract}
\section{Introduction}
\input{sections/introduction.tex}
\section{Related Work}
\label{sec:related_work}
\input{sections/relatedWork.tex}
\section{Background}
\label{sec:background} 
\input{sections/background.tex}
\section{Sparse Inertial Poser}
\label{sec:sparseInertialPoser} 
\input{sections/method.tex}
\section{Experiments}
\label{sec:experiments}
\input{sections/experiments.tex}
\section{Conclusions and Future Work}
\input{sections/conclusions.tex}

\input{sections/acknowledgments.tex}
\bibliographystyle{eg-alpha-doi}
\bibliography{refs}
\end{document}

%% file: notation.tex
\newcommand{\boldpara}[1]{\vspace{10pt} \noindent {\bf #1}}

\newcommand{\eg}{{\em e.g.}}
\newcommand{\Eg}{{\em E.g.}}
\newcommand{\ie}{{\em i.e.}}
\newcommand{\Ie}{{\em I.e.}}
\newcommand{\vs}{{\em vs.}}
\newcommand{\etal}{{\em et al.}}
\newcommand{\etc}{{\em etc.}}
\newcommand{\wrt}{{\em w.r.t.}}
\newcommand{\dof}{{\em d.o.f.}}
\newcommand{\cf}{{\em c.f.}}
\newcommand{\Cf}{{\em C.f.}}
\newcommand{\aka}{{\em a.k.a.}}

\newcommand{\partref}[1]{Part~\ref{#1}}
\newcommand{\chapref}[1]{Chapter~\ref{#1}}
\newcommand{\eqnref}[1]{Eq.~(\ref{#1})}
\newcommand{\sectref}[1]{Section~\ref{#1}}
\newcommand{\figref}[1]{Figure~\ref{#1}}
\newcommand{\tabref}[1]{Table~\ref{#1}}

\newcommand{\dataset}{{\cal D}}
\newcommand{\R}{\mathbb{R}}      

\newcommand{\mrf}{{\cal G}}
\newcommand{\dbn}{{\cal B}}
\newcommand{\edges}{{\cal E}}
\newcommand{\vertices}{{\cal V}}
\newcommand{\edge}{e}
\newcommand{\vertex}{v}
\newcommand{\potential}[2]{\phi_{#1}({#2})}

\newcommand{\vect}[1]{\mathbf{#1}}
\newcommand{\mat}[1]{\vect{#1}}
\newcommand{\matelem}[2]{\mat{#1}_{#2}}
\newcommand{\vecelem}[2]{\vect{#1}_{#2}}

\newcommand{\rv}[1]{{#1}}
\newcommand{\vrv}[1]{\vect{#1}}
\newcommand{\pdf}[1]{p({#1})}
\newcommand{\apppdf}[1]{q({#1})}
\newcommand{\covmat}{{\Sigma}}
\newcommand{\ppar}{\vect{x}}

\newcommand{\fkins}{\mathcal{F}}
\newcommand{\fkin}[1]{\mathcal{F}({\ppar;#1})} 
\newcommand{\jacobpose}[1]{\mat{J}_p(\ppar;#1)}

\newcommand{\pointc}[1]{\boldsymbol{#1}} 
\newcommand{\ps}{\vect{p}_s} 
\newcommand{\pb}{\vect{p}_b} 
\newcommand{\p}{\vect{p}} 
\newcommand{\ri}{\vect{r}} 
\newcommand{\homo}[1]{\bar{\vect{#1}}} 
\newcommand{\proj}{\mathcal{P}_c}
\newcommand{\E}{E}
\newcommand{\e}{\vect{e}} 
\newcommand{\fv}{\vect{r}}
\newcommand{\FV}{R}

\newcommand{\oriset}{\mathcal{O}_{\mathrm{sens}}}
\newcommand{\accset}{\mathcal{A}_{\mathrm{sens}}}

\newcommand*{\todo}[1]{\textcolor{red}{#1}}
\newcommand{\gerard}[1]{\textcolor{magenta}{\textbf{[Gerard: #1]}}}
\newcommand{\timo}[1]{\textcolor{red}{\textbf{[Timo: #1]}}}
\newcommand{\bodo}[1]{\textcolor{red}{\textbf{[Bodo: #1]}}}

\newcommand{\added}[1]{#1}

%% file: sections/introduction.tex
The recording of human motion has revolutionized the fields of biomechanics, computer animation, and computer vision. 
Human motion is typically captured using commercial marker-based systems such as \cite{Vicon} or  \cite{Simi}, and numerous recordings of human performances
are now available (\emph{e.g.,} \cite{CMU}, \cite{Mixamo}, \cite{Moves}).
The recording of human motion is also important for psychology and medicine, where biomechanical analysis can be used
to assess physical activity and diagnose pathological conditions and monitor post-operative mobility of patients.
Unfortunately, marker-based systems are intrusive and restrict motions to controlled laboratory spaces. Therefore,
activities such as skiing, biking or simple daily activities like having coffee with friends cannot be recorded with such systems. 
The vision community has seen significant progress in the estimation of 3D human pose from images, but this typically involves multi-camera calibrated systems, which again limit applicability.
Existing methods for estimating 3D human pose from single images, e.g.~ \cite{Bogo:ECCV:2016}, are still less accurate than motion capture systems.
However, to record human motion in everyday situations and in natural settings
one would need a dedicated camera to track a specific subject. Hence, it is unlikely that vision-based systems will be able to record
large amounts of continuous daily activity data. 

Systems based on Inertial Measurement Units (IMUs) do not suffer from such limitations; they can track the human pose without cameras
which make them more suitable for outdoor recordings, scenarios with occlusions, baggy clothing or where tracking with a dedicated camera is simply not possible. 
However, inertial measurement systems such as Xsens BioMech \cite{Xsens} are quite intrusive, requiring 17 sensors worn on the body or attached to a suit.
This is one of the reasons that large amounts of data have not been recorded yet. 
Hence, a less intrusive solution that can capture people through occlusions is needed.  

In this paper, we present the Sparse Inertial Poser (SIP), a method to recover the full 3D human pose from only 6 IMUs. Six sensors, measuring orientation and acceleration are attached to the wrists, lower legs, waist and head, resulting in a minimally intrusive solution to capture human activities. 
Furthermore, many consumer products already have IMUs integrated, e.g., fitness and smartwatches, smartphones, Google glasses and Oculus rift. 
Our 6-sensor system could easily be worn with a hat or glasses, two wrist bands, a belt, and shoe or ankle sensors.
However, recovering human pose from only 6 IMUs is a very difficult task. 
Orientation at the extremities and waist only provides a weak constraint on the human motion and incorporation of acceleration data is usually affected by drift. 

To solve this problem, we exploit \added{the rich statistical SMPL body model \cite{SMPL:2015}}. 
One key insight is that the body model can be fit to incomplete and ambiguous
data because it captures information about the kinematic constraints of the human body. \added{A similar observation has been made by \cite{tagliasacchi2015robust} and \cite{taylor2016efficient} who leveraged a statistical model for hand pose tracking.} Unfortunately, this alone is not sufficient to compensate for drift. 
Most previous methods (e.g. \cite{roetenberg2007moven, vlasic2007practical}) integrate acceleration frame by frame, which results in unstable estimates when using very few sensors. Optimizing frame by frame
is similar to a double explicit integration scheme, which is known to be unstable and only accurate within small time intervals.
 
We take a different approach and optimize all the poses of all the frames of a sequence at once. Hence, our objective function enforces the coherency between the body model orientation and acceleration estimates against the IMU recordings. 
Effectively, the realistic body model simplifies the estimation problem, providing sufficient constraints to solve the problem from sparse measurements, even for complex movements.
Some examples are shown in Fig.~\ref{fig:teaser}.

In several experiments we show that SIP, while simple, is very powerful and can recover all the poses of a sequence as a result of a single optimization.
We report results on the recently released \added{TNT15 dataset \cite{Marcard2016}} which features 4 subjects wearing 10 IMUs performing a variety of human actions. 
To evaluate SIP we use 6 IMUs for tracking and 4 IMUs for validation. We compare to two baselines, namely an orientation-only tracker that uses only the orientation
information and a variant of SIP that uses a different human body model. 
Qualitative and quantitative results demonstrate that SIP is significantly more accurate
than the baselines. To further demonstrate the applicability of SIP, we present additional tracking results of two subjects wearing 6 IMUs in an outdoor setting (see Fig.~\ref{fig:teaser}).

In summary, SIP makes the challenging problem of human pose estimation from sparse IMU data feasible by:
\begin{itemize}
\item Making use of a realistic body model that incorporates anthropomorphic constraints (with a skeletal rig). 
\item A joint optimization framework that fits the poses of a body model to the orientation and acceleration measurements over multiple frames. 
\end{itemize}
Altogether SIP is the first method that is able to estimate the 3D human pose from only 6 IMUs without relying on databases of MoCap or learning methods that make strong assumptions about the recorded motion. 

%% file: sections/relatedWork.tex
The literature on human pose estimation from images is vast and in this paper we focus
only on methods integrating multiple sensor modalities and methods predicting full pose from
sparse low dimensional control signals. 
\subsection{Database retrieval and learning based methods}
Some work has focused on recovering full pose from sparse incomplete sensor signals. 
In \cite{Slyper2008,tautges2011motion} they reconstruct human pose from $5$ accelerometers
by retrieving pre-recorded poses with similar accelerations from a database. 
Acceleration data is however very noisy and the space of possible accelerations is huge which makes
learning a very difficult task. 
A somewhat easier problem is addressed in \cite{chai2005performance}; they reconstruct full 3D pose from a set of
sparse markers attached at the body. They build online local PCA models using the sparse marker 
locations as input to query the database of human poses. This approach works well since the 5-10 marker
locations can constrain the pose significantly; furthermore the mapping from 3D locations to pose is much
more direct than from acceleration data. Unfortunately, this approach is restricted to a lab with cameras
capturing the reflective markers. 
Following similar ideas, in \cite{liu2011realtime} they regress to full pose using online local models but using 
$6$ IMUs to query the database. In \cite{schwarz2009discriminative} they directly regress full pose using only $4$ IMUs with 
Gaussian Process regression. Both methods report very good results when the test motions are present
in the database. In~\cite{hannink2016sensor} they extract gait parameters using deep convolutional neural networks.
Although pre-recorded human motion greatly constrains the problem, methods that heavily
rely on pre-recorded data are limited; in particular capturing arbitrary activities is difficult if it is missing in the databases. 
\subsection{Full-body IMU MoCap}
There exist commercial solutions for human motion capture from IMUs;
\cite{roetenberg2007moven} use 17 IMUs equipped with 3D accelerometers, gyroscopes and magnetometers and all 
the measurements are fused using a Kalman Filter. By achieving stable orientation measurements the 17 IMUs completely define the pose of the subject.
However it is very intrusive for a subject to wear them, and long setup times are required. 
In the seminal work of \cite{vlasic2007practical} they propose a custom made system consisting of 18 sensor boards, each equipped with an IMU and acoustic distance sensors, to compensate for typical drift in the orientation estimates. While the approach is demonstrated in challenging outdoor 
settings like ours, the system is also very intrusive and difficult to reproduce. 
Other approaches have combined sparse IMUs with video input \cite{Pons-Moll2011,Marcard2016} \added{or sparse optical markers \cite{andrews2016real}} to constrain the problem. 
Similarly \cite{helten2013real} combines sparse IMUs with a depth camera. IMUs are only used to query similar
poses in a database and depth data is used to obtain the full pose. 
While powerful, using video input does not allow human movements to be captured with occlusions or in applications that require continuous 
activity monitoring. 
Hence, instead of constraining the problem using additional sensors, we constrain the problem by using a statistical
body model and optimizing the pose over multiple frames. While 6 IMUs do not provide enough constraints to
determine the full pose for a single frame, we find that  accurate pose estimates can be obtained
when integrating all orientation and acceleration measurements into a single optimization objective.

%% file: sections/background.tex
\subsection{Exponential Map on SO(3) and SE(3)}
In this section we quickly review the concept of exponential mapping on the Special Orthogonal Group SO(3) and the Special Euclidean Group SE(3). The exponential map representation provides a geometric and elegant treatment of rigid body motion, which we use to relate pose parameters to human body motions. \added{Using the exponential map has some advantadges for optimization w.r.t. other representations such as Euler angles~\cite{pons2009balljoints}; for more details on the exponential mapping and a comparison to other parameterizations we refer the reader to
\cite{murray1994mathematical,PonsModelBased}}.

Both SO(3) and SE(3) are Lie groups with an associated Lie algebra.
Throughout this paper we will use the cross-operator $^\times$ to construct a Lie algebra element from its coordinates and the \mbox{vee-operator $^\vee$} to extract the coordinates of a Lie algebra element into a column vector.
The group of rotations about the origin in 3 dimensions SO(3) is defined as
$ \mathrm{SO}(3) = \{ {\mat{R} \in \R^{3 \times 3}:} {\mat{R}^T \mat{R} =\mat{I}}, {det(\mat{R}) = 1} \}$.
Every rotation $\mat{R}$ can be expressed in exponential form
\begin{equation}
\mat{R} = \exp(\vect{\omega}^\times),
\label{eq:expSO3}
\end{equation}
where $\vect{\omega}^\times \in so(3)$ is a skew-symmetric matrix and can be computed analytically using the Rodriguez Formula \cite{murray1994mathematical}. 
The three independent parameters $\vect{\omega} \in \R^3$ of $\vect{\omega}^\times$ are called exponential coordinates of $\mat{R}$ and define the axis of rotation and $||\vect{\omega}||$ is the angle of rotation about this axis.
The group SE(3) represents rigid body motions composed by a rotation $\mat{R} \in \mathrm{SO}(3)$ and translation $\vect{t} \in \R^3$. Any rigid motion $\mat{G} \in \R^{4 \times 4}$ can be written in exponential form
\begin{equation}
\mat{G} = \begin{bmatrix} \mat{R} & \vect{t} \\ 0 & 1 \end{bmatrix} = \exp(\vect{\xi}^\times),
\label{eq:expSE3}
\end{equation}
where $\vect{\xi}^\times \in se(3)$ is called the associated twist action and $se(3)$ refers to the corresponding Lie algebra. The six independent parameters $\vect{\xi} \in \R^6$ of $\vect{\xi}^\times$ are called exponential coordinates of $\mat{G}$. They are composed of the rotational parameters $\vect{\omega} \in \R^3$ and $\vect{v} \in \R^3$, where the latter encodes location of the axis of rotation and translation along the axis.

The inverse operation of \eqnref{eq:expSO3} and \eqnref{eq:expSE3} is the logarithm and recovers a Lie algebra element from a Lie group element.
We also introduce the Taylor expansion of the matrix exponential given by
\begin{equation}
\exp(\xi^\times) = \mat{I} + \xi^\times + \frac{(\xi^\times)^2}{2!} + \frac{(\xi^\times)^3}{3!} + \dots,
\label{eq:FirstOrderExp}
\end{equation}
and the first-order approximation for the logarithm
\begin{equation}
\log(\exp(\delta \omega^\times)\exp(\omega^\times))^\vee \approx \delta \omega + \omega,
\label{eq:FirstOrderLog}
\end{equation}
\added{for a small $\delta \omega \in \R^3$}.
\subsection{SMPL Body Model}
\label{sec:KinematicChainModel}
\begin{figure}
\centering
\subfigure[]{ \includegraphics[height=3.5cm]{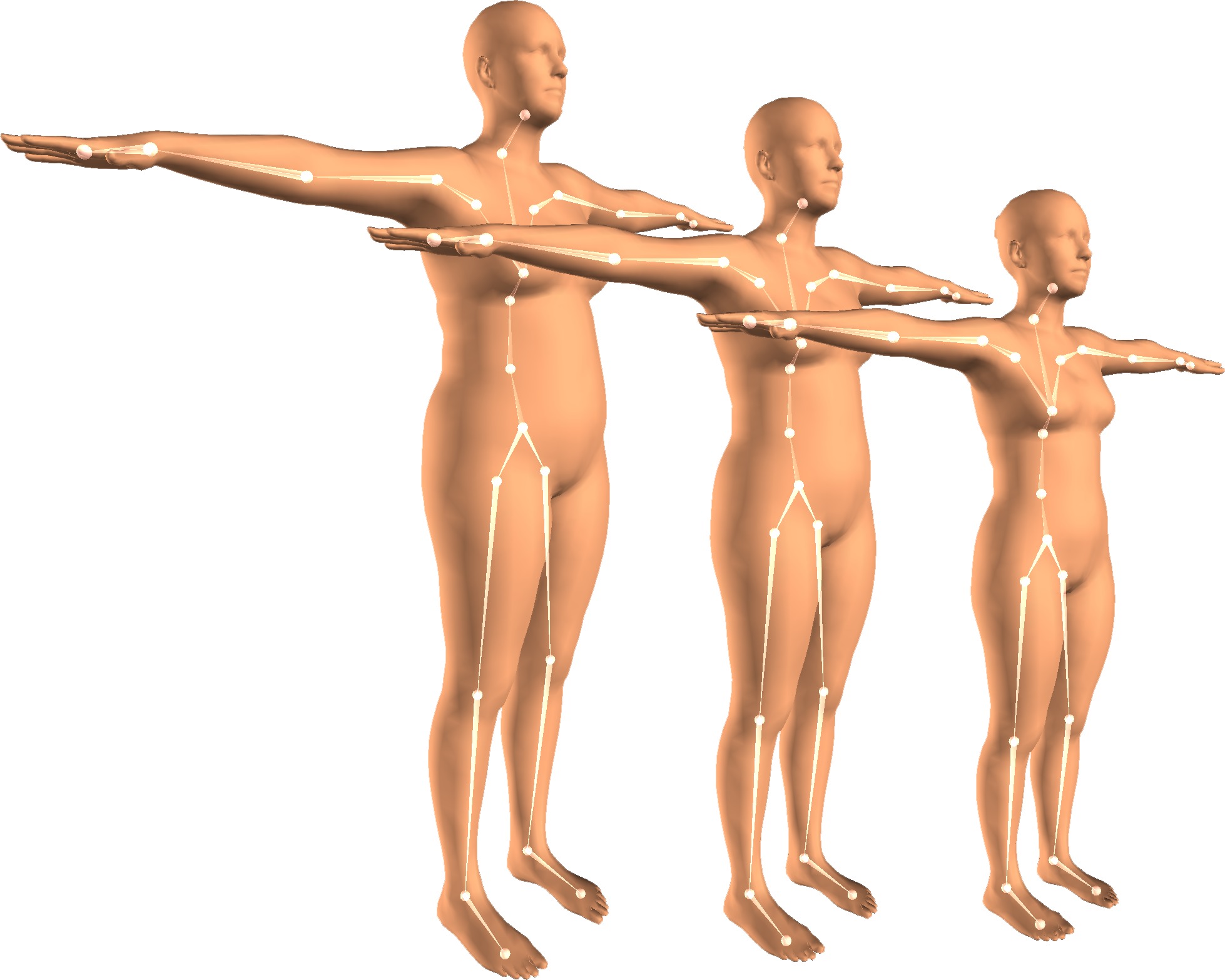}\label{fig:SMPL}}
\subfigure[]{\includegraphics[height=3.5cm]{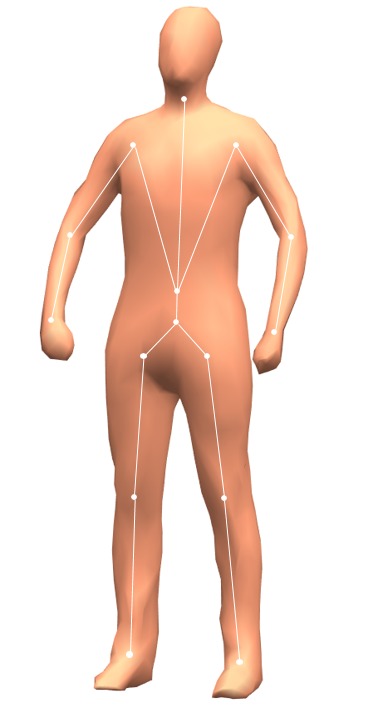}\label{fig:tntMesh}}
\caption{(a) The joints of the skeleton in SMPL are predicted as a function of the surface. 
This allows us to obtain accurate joint locations which are used to predict the acceleration measurements. (b) Manually rigged models lead to worse performance
fitting incomplete sensor measurements. }
\end{figure}
SMPL \cite{SMPL:2015} is a body model that uses a learned template with $V=6890$ vertices $\mat{T}$, and a learned rigged template skeleton. 
The actual vertex positions of SMPL are adapted according to identity-dependent shape parameters and the skeleton pose.
The skeletal structure of the human body is modeled with a kinematic chain consisting of rigid bone segments linked by $n=24$ joints. 
Each joint is modeled as a ball joint with 3 rotational Degrees of Freedom (DoF), parametrized with exponential coordinates $\vect{\omega}$.
Including translation, the pose $\ppar$ is determined by a pose vector of $d=3\times{24} + 3 = 75$ parameters.
The rigid motion $\mat{G}^{TB}(\ppar)$ of a bone depends on the states of parent joints in the kinematic chain and can be computed by 
the forward kinematic map $\mat{G}^{TB}: \R^{d} \to \mathrm{SE}(3)$:
\begin{equation}
\mat{G}^{TB}(\ppar) =  \left(  \prod_{j \in I(i)} \bigg[\begin{array}{c|c}
                                                      \exp(\vect{\omega}_j^\times) &\vect{j} \\
                                                      \hline
                                                      \vec{0} &1
                                                      \end{array}\bigg] \right) = \left(  \prod_{j \in I(i)} \exp\left( \vect{\xi}^\times_j \right) \right) ,
\label{eq:forwardKinematicMap}
\end{equation}
where $I(i) \subseteq \{ 1,\cdots,n+1 \}$ is an ordered set of parent joints, $\vect{\omega}_j \in \R^3$ are the exponential coordinates of the joint rotation, $\vect{j} \in \R^3$ is the joint location and $\vect{\xi}^\times_j \in se(3)$ is the twist action of joint $j$. The initial offset between the bone and the tracking frame is the identity. 

SMPL models body shape variation using shape blend shapes, that are linearly added to the template mesh. 
A new subject shape is typically obtained by adding a linear combination of blendshapes $\mat{S}_i \in \R^{3V}$ to the template mesh $\mat{T}^\prime = \mat{T} + \sum_i \beta_i \mat{S}_i$. 
SMPL automatically predicts the joint locations $\mat{Q} = [\mat{j}_1^T \hdots \vect{j}_n^T]^T$ as a function of the surface mesh using a sparse regression matrix $\mat{Q} = \mathcal{J} \mat{T}^\prime$. While the orientation of the limbs do not depend at all on the body joints, the linear acceleration of a particular part of the body depends on the joint locations. By using SMPL we can
track any shape without having to manually edit the skeleton, see~\figref{fig:SMPL}.
\subsection{IMUs}
\label{sec:imus}
An Inertial Measurement Unit (IMU) is a device that is commonly equipped with 3-axes accelerometers, gyroscopes and magnetometers. It measures acceleration, rate of turn and magnetic field strength with respect to the IMU-aligned sensor coordinate system $F^S$. \added{Typically, a Kalman Filter is then applied to track the sensor orientation with respect to a global inertial coordinate system $F^I$.

In order to utilize IMU data together with the body model we introduce several coordinate systems depicted in \figref{fig:coordinateSystems}. The body model is defined in the global tracking coordinate system $F^G$ and each bone segment of the body has a local coordinate system $F^B$. The map $\mat{G}^{GB}:F^B \to F^G$ defines the mapping from bone to tracking coordinate system. Equivalently, $\mat{G}^{IS}:F^S \to F^I$ defines the mapping from the local IMU sensor coordinate system $F^S$ to $F^I$.
\begin{figure}[t]
\centering
\subfigure[]{\label{fig:coordinateSystems}
\def \svgwidth{0.4\linewidth}
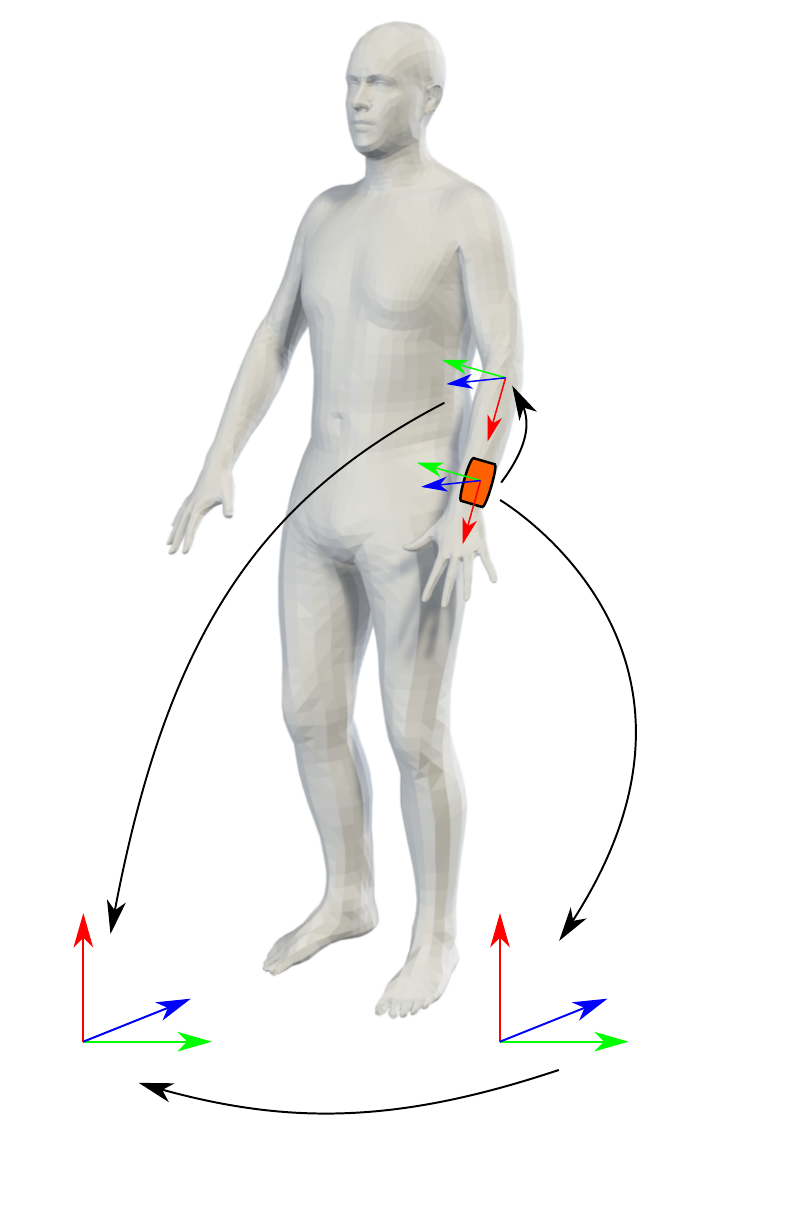}
\subfigure[]{\label{fig:sensorPlacement}
\raisebox{3mm}{\includegraphics[width=0.45\linewidth]{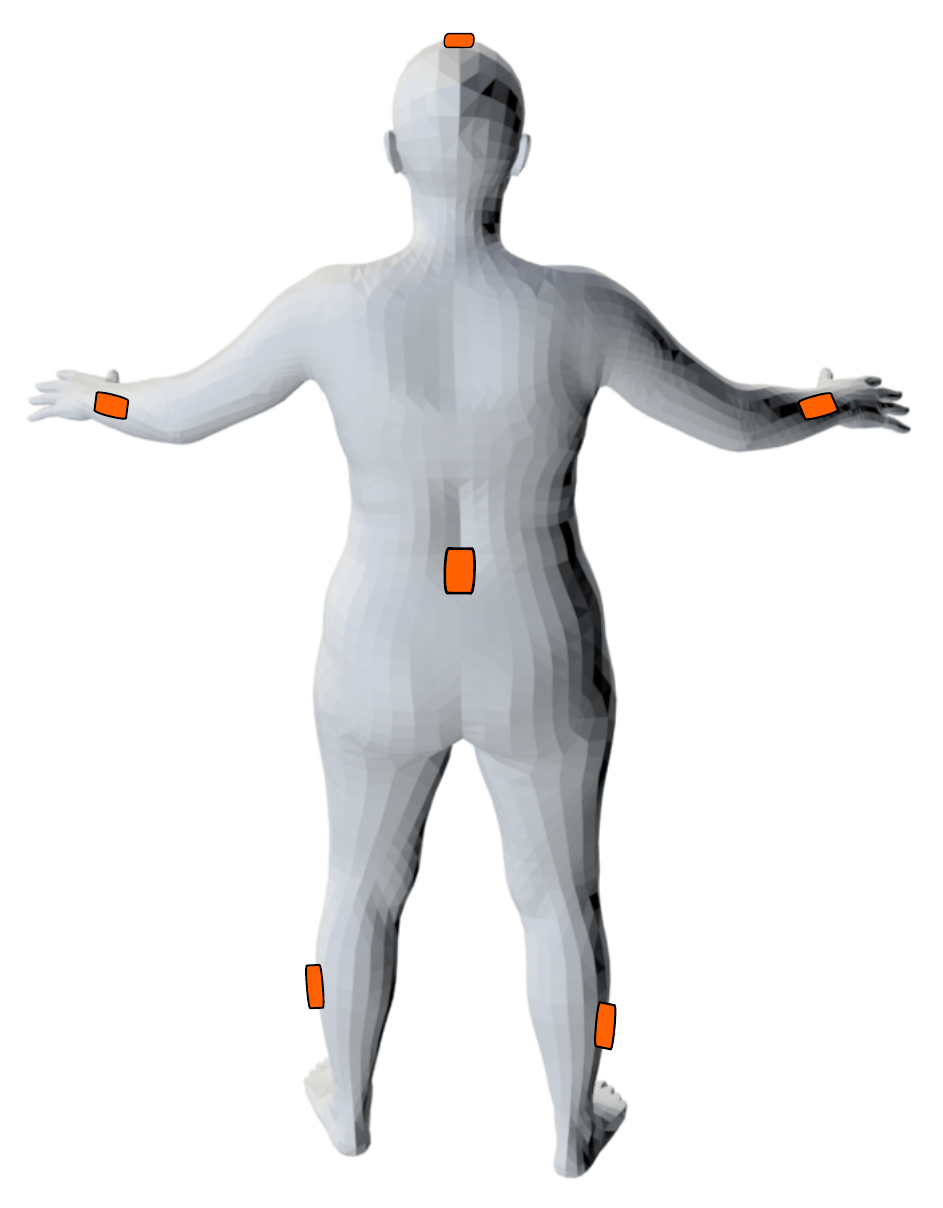}}}
\caption{ (a) Coordinate frames: Global tracking coordinate frame $F^G$, Inertial coordinate frame $F^I$, Bone coordinate frame $F^B$ and Sensor coordinate frame $F^S$. (b) Sensor placement at head, lower legs, wrists and back.}
\end{figure}
Both global coordinate systems $F^G$ and $F^I$ are related by the constant mapping $\mat{G}^{GI}:F^I \to F^G$. In the following we will assume $\mat{G}^{GI}$ is known and express all IMU readings in the global tracking frame $F^G$ using the transformation rule
\begin{equation}
\mat{G}^{GS}(t) = \mat{G}^{GI} \mat{G}^{IS}(t).
\end{equation}
For a more detailed description of relating inertial data to other sensor or model coordinate systems we refer the reader to \cite{baak2010analyzing}.
Our aim is to find a pose trajectory such that the motion of a limb is consistent with IMU acceleration and orientation attached to it. Thus we need to know the offset between IMU and its corresponding bone coordinate system $\mat{G}^{BS}(t):F^S \to F^B$. We assume that it is constant as the sensors are tightly attached to the limbs and compute it at the first frame of the tracking sequence according to
\begin{equation}
\mat{G}^{BS} = \mat{G}^{BG}(0)\mat{G}^{GS}(0).
\end{equation}
}

%% file: images/coordinateSystems2.pdf_tex
\begingroup%
  \makeatletter%
  \providecommand\color[2][]{%
    \errmessage{(Inkscape) Color is used for the text in Inkscape, but the package 'color.sty' is not loaded}%
    \renewcommand\color[2][]{}%
  }%
  \providecommand\transparent[1]{%
    \errmessage{(Inkscape) Transparency is used (non-zero) for the text in Inkscape, but the package 'transparent.sty' is not loaded}%
    \renewcommand\transparent[1]{}%
  }%
  \providecommand\rotatebox[2]{#2}%
  \ifx\svgwidth\undefined%
    \setlength{\unitlength}{232bp}%
    \ifx\svgscale\undefined%
      \relax%
    \else%
      \setlength{\unitlength}{\unitlength * \real{\svgscale}}%
    \fi%
  \else%
    \setlength{\unitlength}{\svgwidth}%
  \fi%
  \global\let\svgwidth\undefined%
  \global\let\svgscale\undefined%
  \makeatother%
  \begin{picture}(1,1.51724138)%
    \put(0,0){\includegraphics[width=\unitlength]{coordinateSystems2.pdf}}%
    \put(1.07323187,0.23603199){\color[rgb]{0,0,0}\makebox(0,0)[lb]{\smash{
}}}%
    \put(-0.03448276,0.13793103){\color[rgb]{0,0,0}\makebox(0,0)[lb]{\smash{$F^T$}}}%
    \put(0.62068966,1.06896551){\color[rgb]{0,0,0}\makebox(0,0)[lb]{\smash{$F^B$}}}%
    \put(0.45724138,0.75310344){\color[rgb]{0,0,0}\makebox(0,0)[lb]{\smash{$F^S$}}}%
    \put(0.81034483,0.62068965){\color[rgb]{0,0,0}\makebox(0,0)[lb]{\smash{$G^{IS}$}}}%
    \put(0.37931034,0.02448275){\color[rgb]{0,0,0}\makebox(0,0)[lb]{\smash{$G^{TI}$}}}%
    \put(0.04896552,0.62068965){\color[rgb]{0,0,0}\makebox(0,0)[lb]{\smash{$G^{TB}$}}}%
    \put(0.65517241,0.93103447){\color[rgb]{0,0,0}\makebox(0,0)[lb]{\smash{$G^{BS}$}}}%
    \put(0.75862069,0.13793103){\color[rgb]{0,0,0}\makebox(0,0)[lb]{\smash{$F^I$}}}%
  \end{picture}%
\endgroup%

%% file: sections/method.tex
\begin{figure}
\centering
\includegraphics[width=\linewidth]{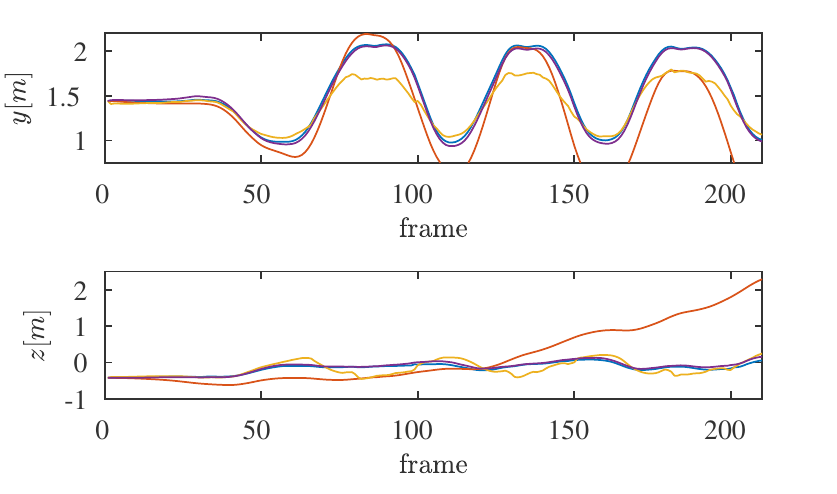}
\caption{Y- and Z-coordinates of the left wrist sensor position (Y pointing upwards) for a jumping jack sequence, which is also shown in \figref{fig:optimizationIterations}. Ground truth positions obtained by tracking with 10 IMUs, are shown in purple and are almost indistinguishable from the estimated sensor positions obtained with SIP (blue). Using only orientation (yellow) of 6 IMUs provides accurate estimates for some portions of the sequence, but cannot correctly reconstruct the extended, raised arm. Double integrating acceleration values (red) provides only reasonable estimates at the beginning of the sequences and the error accumulates over time.}
\label{fig:accelerationDrift}
\end{figure}
\begin{figure}[t]
\centering
\def \svgwidth{0.45\linewidth}
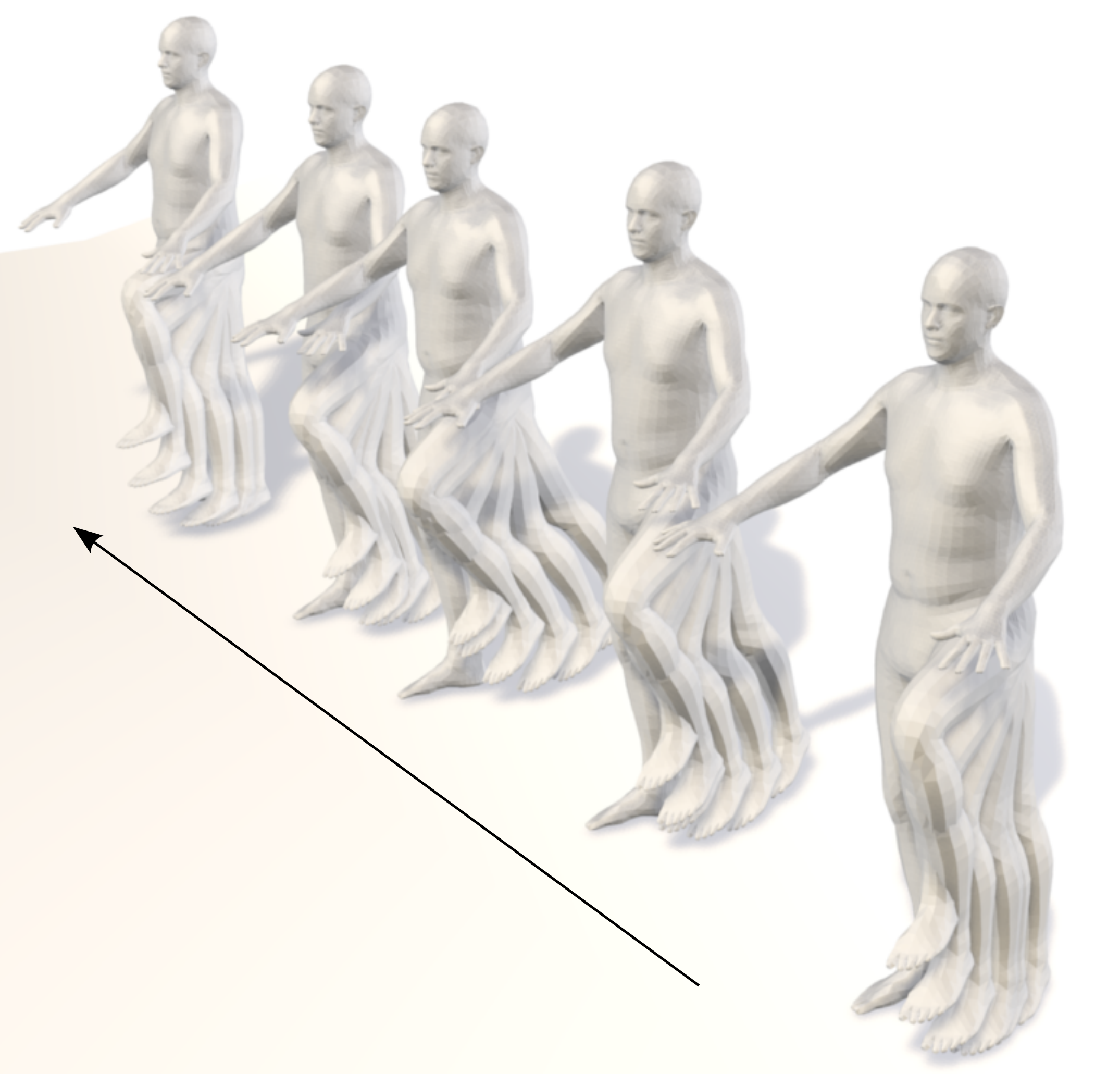
\def \svgwidth{0.45\linewidth}
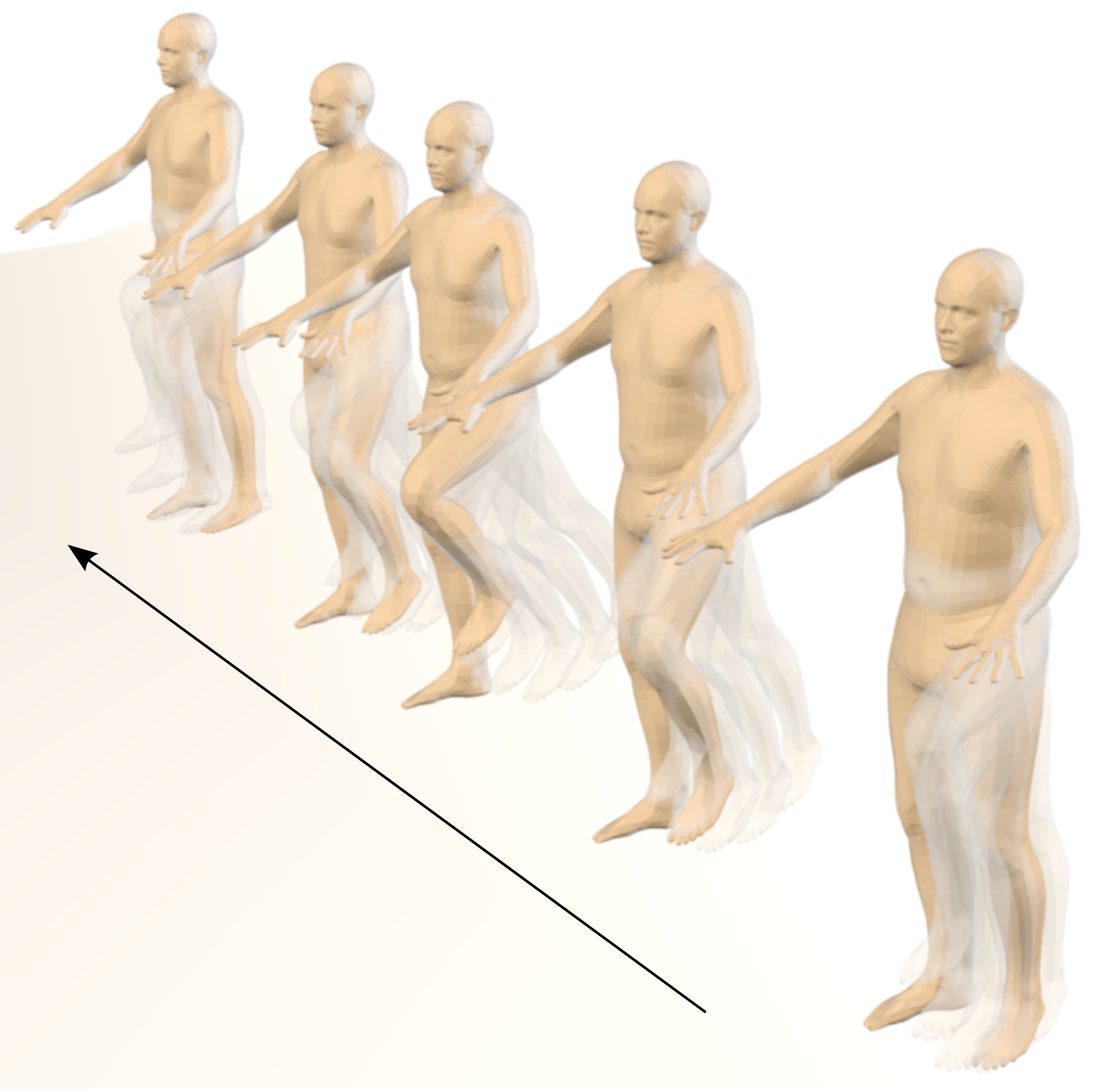
\caption{SIP joint optimization: sparse IMUs give only weak constraints on the full pose. As illustrated on the left figure, multiple poses fit well the IMU orientation of the lower left leg. By optimizing all poses over the sequence we can successfully find the pose trajectory (shown in orange) that is also consistent with the acceleration data as can be seen on the right figure. The joint optimization allows the use of acceleration readings, which would produce severe drift otherwise.}
\label{fig:SIP_illustration}
\end{figure}
Recovering full pose from only $N=6$ IMUs (strapped at lower arms, lower legs, head and waist) is highly ambiguous. 
Assuming no sensor noise, orientation data only constrains the full pose to lie on a lower dimensional manifold. 
Acceleration measurements are noisy and naive double integration to obtain position leads to unbounded exponential drift, see \figref{fig:accelerationDrift}. 
Looking at a single frame the problem is ill-posed. However, looking at the full sequence, and using anthropometric constraints
from a body model, makes the problem much more constrained, see \figref{fig:SIP_illustration}. This motivates us to formulate the following multi-frame objective function:
\begin{equation}
\ppar_{1:T}^{*} = \underset{\ppar_{1:T}}{\text{arg min }} E_\mathrm{motion}(\ppar_{1:T},\mat{R}_{1:T},\vect{a}_{1:T}),
\label{eq:Objective}
\end{equation}
where $\ppar_{1:T} \in \R^{75T}$ is a vector consisting of stacked model poses for each time step $t = 1 \dots T$. \added{$\mat{R}_{1:T}$ are the sensor orientations $\mat{R}_t \in SO(3)$ and $\vect{a}_{1:T}$ are the sensor acceleration measurements respectively. We} define $E_\mathrm{motion}: \R^{d \times T} \times \R^{3N \times T} \times \R^{3N \times T} \to \R$ as
\begin{equation} \label{eq:Energy}
\begin{split}
E_{\mathrm{motion}}(\ppar_{1:T},\omega_{1:T},\vect{a}_{1:T}) = &  w_{\mathrm{ori}} \cdot E_{ori}(\ppar_{1:T},\mat{R}_{1:T}) \\
 &+ w_\mathrm{acc} \cdot E_\mathrm{acc}(\ppar_{1:T},\vect{a}_{1:T}) \\ 
 &+ w_\mathrm{anthro} \cdot E_\mathrm{anthro}(\ppar_{1:T}) , 
\end{split}
\end{equation}
where $E_\mathrm{ori}$, $E_\mathrm{acc}$ and $E_\mathrm{anthro}$ are energies related to orientation, acceleration and anthropometric consistency. The weights of \eqnref{eq:Energy} are fixed during all experiments, see experimental section. In the following, we detail each of the objective terms. 

\subsection{The Orientation Term}
The sensor orientations, $\mat{R}^{GS}(t): F^S\to F^G$ are related to the bone orientations by a constant rotational offset $\mat{R}^{BS}$. Hence, we define the estimated sensor orientation $\hat{\mat{R}}^{GS}(\ppar_t)$ at the current pose $\ppar_t$ as
\begin{equation}
\hat{\mat{R}}^{GS}(\ppar_t) = \mat{R}^{GB}(\ppar_t) \mat{R}^{BS},
\end{equation}
where $\mat{R}^{GB}(\ppar_t)$ is the rotational part of the forward kinematics map defined in \eqnref{eq:forwardKinematicMap} and $\mat{R}^{BS}$. The \emph{orientation error} $\vect{e}_{ori} \in \R^3$ \added{are the exponential coordinates of the rotational offset between estimated and measured sensor orientation:}
\begin{equation}
\vect{e}_{ori}(\ppar_t) = \log \left( \hat{\mat{R}}^{GS}(\ppar_t) \left( \mat{R}^{GS}(t) \right)^{-1} \right)^\vee,
\label{eq:OriResidual}
\end{equation}
\added{where} the $\vee$-operator is used to extract the coordinates of the skew-symmetric matrix obtained from the log-operation.
We define the orientation consistency $E_\mathrm{ori}$ across the sequence as 
\begin{equation}
E_{ori} =\added{\frac{1}{TN}} \sum_{t = 1}^T  \sum_{n = 1}^N || \vect{e}_{ori,n}(t) ||^2,
\label{eq:OriEnergy}
\end{equation}
which is the sum of squared L2-norm of orientation errors over all frames $t$ and all sensors $n$. Actually, the squared L2-norm of $\vect{e}_{ori}$ corresponds to the geodesic distance between $\hat{\mat{R}}^{GS}(\ppar_t)$ and $\mat{R}^{GS}(t)$ \cite{Hartley2013, Marcard2016}.
\subsection{The Acceleration Term}
IMU acceleration measurements $\vect{a}^S$ are provided in the sensor coordinate system $F^S$ shown in \figref{fig:coordinateSystems}. To obtain the corresponding sensor acceleration $\vect{a}^G$ in global tracking frame coordinates $F^G$ we have to transform $\vect{a}^S$ by the current sensor orientation $\mat{R}^{GS}(t)$ and subtract gravity $\vect{g}^G$
\begin{equation}
\vect{a}^G_t = \mat{R}^{GS}_t \vect{a}^S_t - \vect{g}^G.
\end{equation}
We aim to recover a sequence of poses such that the actual sensor acceleration matches the corresponding vertex acceleration of the body model. The corresponding vertex is manually selected; since the model has the same topology across subjects this operation is done only once. The vertex acceleration $\hat{\vect{a}}^G(t)$ is approximated by numerical differentiation
\begin{equation}
\hat{\vect{a}}_t^G = \frac{\vect{p}_{t-1}^G - 2 \vect{p}_{t}^G + \vect{p}_{t+1}^G}{dt^2},
\end{equation}
where $\vect{p}_t^G$ is the vertex position at time instance $t$ and $dt$ is the sampling time. The vertex position is related to the model pose $\ppar$ by the forward kinematic map defined in \eqnref{eq:forwardKinematicMap} and equates to
\begin{equation}
\bar{\vect{p}}^G(\ppar) = \mat{G}^{GB}(\ppar) \bar{\vect{p}}^B(0),
\label{eq:SensorPosition}
\end{equation}
where $\bar{\vect{p}}$ indicates homogeneous coordinates. Hence, we define the acceleration error as the difference of estimated and measured acceleration
\begin{equation}
\vect{e}_\mathrm{acc}(t) = \hat{\vect{a}}^G(\ppar_{t-1}, \ppar_{t},\ppar_{t+1}) - \vect{a}_t^G.
\label{eq:AccResidual}
\end{equation}
Adding up the acceleration error for all $T$ frames and $N$ sensors defines the motion acceleration consistency $E_{acc}$:
\begin{equation}
E_\mathrm{acc} = \added{\frac{1}{TN}} \sum_{t = 1}^T  \sum_{n = 1}^N || \vect{e}_{\mathrm{acc},n}(t) ||^2.
\label{eq:AccEnergy}
\end{equation}
\subsection{The Anthropometric Term}
In order to constrain the skeletal joint states to human-like poses we use a multivariate Gaussian distribution of model poses with a mean pose $\vect{\mu}_{\ppar}$ and covariance matrix $\Sigma_{\ppar}$ learned from the scan registrations of SMPL. While this encodes
anthropometric constraints it is not motion specific as it is learned from a variety of static poses. Note that this is much less restrictive
than learning based or database retrieval based approaches. 
We use the Mahalanobis distance that measures the likelihood of a pose $\ppar$ given the distribution $\mathcal{N}(\vect{\mu}_{\ppar},\Sigma_{\ppar})$:
\begin{equation}
d_\mathrm{mahal} = \sqrt{\left(\ppar - \vect{\mu}_{\ppar} \right)^T \Sigma_{\ppar}^{-1} \left( \ppar - \vect{\mu}_{\ppar} \right)}.
\label{eq:Mahalanobis}
\end{equation}
Additionally, we explicitly model joint limits by an error term which produces repulsive forces if a joint limit is violated. We define the joint limit error $\vect{e}_\mathrm{limit}$ as 
\begin{equation}
\vect{e}_\mathrm{limit} = \min(\ppar- \vect{l}_\mathrm{lower},\vect{0}) + \max(\ppar- \vect{l}_\mathrm{upper},\vect{0}) \,
\label{eq:JointLimits}
\end{equation}
where $\vect{l}_{lower}$ and $\vect{l}_{upper}$ are lower and upper joint limit parameters.
Altogether, the anthropometric energy term $E_\mathrm{antro}$ is a weighted combination of terms
\begin{equation}
E_\mathrm{anthro} = w_\mathrm{mahal}\added{\frac{1}{T}} \sum_{t = 1}^T  d_\mathrm{mahal}(t) ^2 +w_\mathrm{limit}\added{\frac{1}{T}}  \sum_{t = 1}^T  ||\vect{e}_\mathrm{limit}(t)|| ^2 \,
\label{eq:PriorEnergy}
\end{equation}
where the weighting factors $w_\mathrm{mahal}$ and $w_\mathrm{limit}$ balance the influence of the pose prior term and the joint limits term.
%
\subsection{Energy Minimization}
\begin{figure}
\centering
\subfigure[]{ \includegraphics[width=0.23\linewidth]{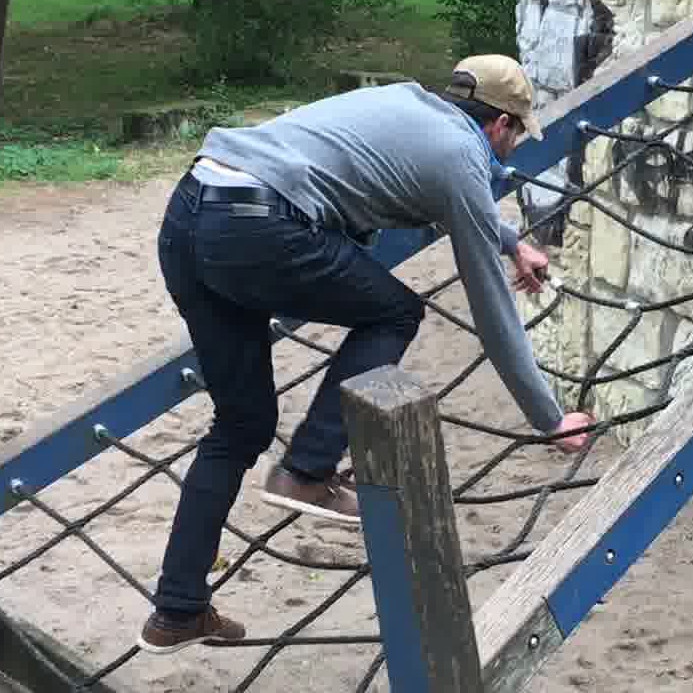}}
\subfigure[]{ \includegraphics[width=0.23\linewidth]{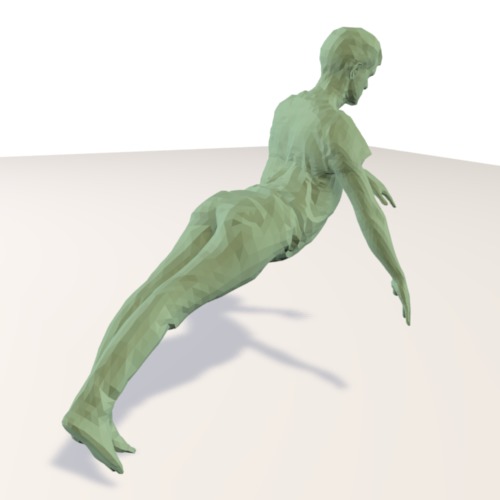}}
\subfigure[]{ \includegraphics[width=0.23\linewidth]{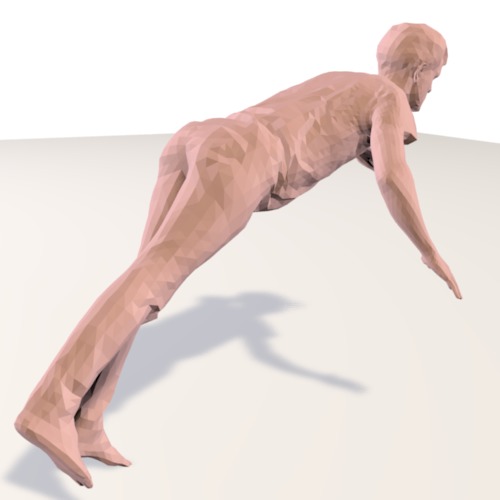}}
\subfigure[]{ \includegraphics[width=0.23\linewidth]{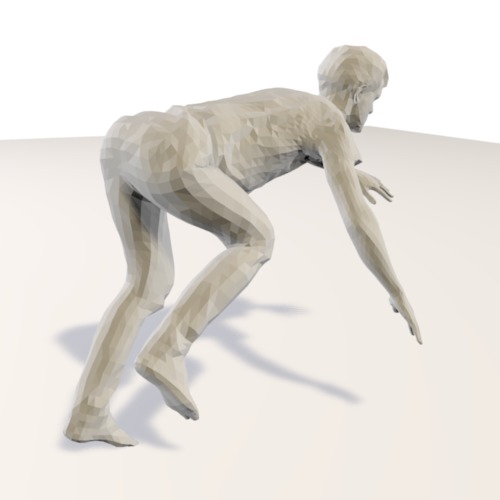}}
\caption{Influence of the anthropometric, orientation and acceleration consistency terms. (a) image of a climbing scene (b) using only orientation without anthropometric consistency term, (c) using orientation with anthropometric consistency term , (d) our proposed SIP using anthropometric, orientation and acceleration consistency terms.}
\end{figure}
$E_\mathrm{motion}$ is a highly non-linear function and generally difficult to optimize. However, the exponential map formulation enables us to analytically compute gradients and since $E_\mathrm{motion}$ is composed of a sum of squared residual terms we can use the \added{Levenberg-Marquardt} algorithm. In order to compute an update-step for the \added{Levenberg-Marquardt} algorithm, we have to linearize the residual terms around the current solution with the Jacobian matrix $\mat{J}$. The Jacobian maps a pose increment $\delta \ppar$ to an increment of the residual according to
\begin{equation}
\vect{e}(\ppar,\delta \ppar) \approx \vect{e}(\ppar) + \mat{J}\delta \ppar.
\end{equation}
In the following we show how to linearize the respective residual terms associated to orientation, acceleration and anthropometric consistency.

The orientation residual defined in \eqnref{eq:OriResidual} can be rewritten in terms of an incremental change of the pose $\delta \ppar$ such that 
\begin{equation}
\vect{e}_{ori}(\ppar,\delta \ppar) = \log \left( \mat{R}(\delta \ppar) \hat{\mat{R}}^{GS}(\ppar) \left( \mat{R}^{GS} \right)^{-1} \right)^\vee ,
\end{equation}
where $\mat{R}(\delta \ppar)$ is the rotational part of the forward kinematic map computed at the current pose $\ppar$. Using the first-order approximation for the logarithm of \eqnref{eq:FirstOrderLog} we get a linearized expression of the orientation residual according to \added{
\begin{equation}
\vect{e}_{ori}(\ppar,\delta \ppar) \approx \vect{e}_{ori}(\ppar) + \delta \vect{e}_{ori}(\delta \ppar).
\end{equation}
The first term corresponds to the actual orientation residual defined in \eqnref{eq:OriResidual} and the latter term is given by
\begin{equation}
\delta \vect{e}_{ori}(\delta \ppar) = J_{ori} \delta \ppar,
\end{equation}
where $\mat{J}_{ori}: \R^{d} \to \R^3$}
is the articulated Jacobian, mapping an incremental variation of the pose vector to rotational increments in the tangent space of SO(3), see \cite{PonsModelBased}.

In order to linearize the acceleration residual of \eqnref{eq:AccResidual}, we rewrite the estimated sensor position (\eqnref{eq:SensorPosition}) at a single time instance in terms of an incremental change in the pose vector $\delta \ppar$ according to
\begin{equation}
\bar{\vect{p}}(\ppar,\delta \ppar) = \mat{G}(\delta \ppar)\mat{G}^{GB}(\ppar) \bar{\vect{p}}(0) = \mat{G}(\delta \ppar)\bar{\vect{p}}(\ppar),
\end{equation}
where $\mat{G}(\delta \ppar)$ is the forward kinematic map computed at the current pose $\ppar$. Using the Taylor expansion (Eq.~\eqref{eq:FirstOrderExp}) of the exponential map of SE(3) up to the first order we get 
\begin{equation}
\bar{\vect{p}}(\ppar,\delta \ppar) \approx \bar{\vect{p}}(\ppar) + \xi^\times_{\delta \ppar}\bar{\vect{p}}(\ppar).
\end{equation}
The second term of the previous equation can be rewritten as
\begin{equation}
\xi^\times_{\delta \ppar}\mat{G}^{GB}(\ppar) \bar{\vect{p}}(0) = \mat{J}_{p(\ppar)} \delta \ppar
\end{equation}
where $\mat{J}_{p(\ppar)}: \R^{d} \to \R^3$ is the positional Jacobian at point $p(\ppar)$, mapping an incremental variation of the pose vector to positional increments in Cartesian coordinates, see also \cite{PonsModelBased}. By combining the position estimates of three successive time steps we get the linearized acceleration error according to
\begin{equation}
\vect{e}_{acc}(t,\delta \ppar) \approx \\ 
\vect{e}_{acc}(t) + 
 \begin{bmatrix} \mat{J}_{p(\ppar_{t-1})} & -2 \mat{J}_{p(\ppar_{t})} & \mat{J}_{p(\ppar_{t+1})}\end{bmatrix} \left[\begin{array}{l}\delta \ppar_{t-1} \\ \delta \ppar_{t} \\ \delta \ppar_{t+1}\end{array}\right].
\end{equation}
The residual terms related to anthropomorphic consistency defined in \eqnref{eq:Mahalanobis} and \eqnref{eq:JointLimits} are already linear in the pose $\ppar$. For the Mahalanobis prior we compute 
the Cholesky factorization of the inverse covariance matrix
\begin{equation}
\Sigma^{-1}_{\ppar} = \mat{L}^T \mat{L} 
\end{equation}
and rewrite the squared Mahalanobis distance as
\begin{equation}
d_\mathrm{mahal}^2 = \left(\ppar - \vect{\mu}_{\ppar} \right)^T  \mat{L}^T \mat{L}  \left( \ppar - \vect{\mu}_{\ppar} \right) = \vect{e}_\mathrm{mahal}^T \vect{e}_\mathrm{mahal}.
\end{equation}
Then it becomes obvious that $\vect{e}_\mathrm{mahal}: \ppar \to L  \left( \ppar - \vect{\mu}_{\ppar} \right)$ is a linear mapping with $\mat{J}_\mathrm{mahal} = \mat{L}$.

\begin{figure}
\centering
\includegraphics[height=2.6cm]{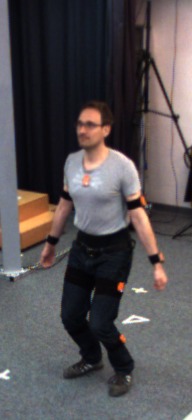}
\includegraphics[height=2.6cm]{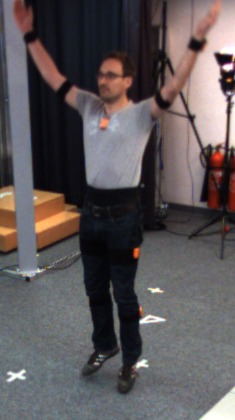}
\includegraphics[height=2.6cm]{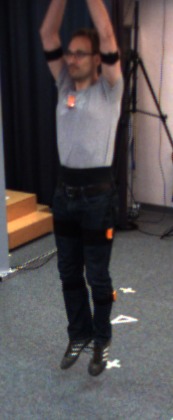}
\includegraphics[height=2.6cm]{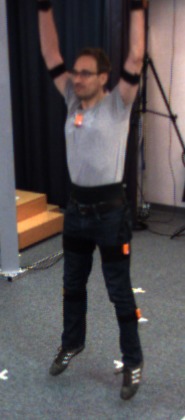}
\includegraphics[height=2.6cm]{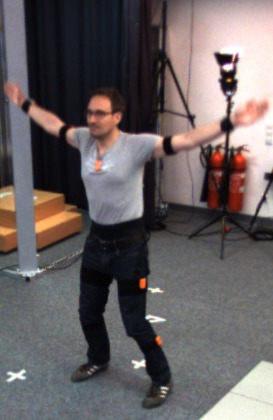}

\includegraphics[width=0.9\linewidth]{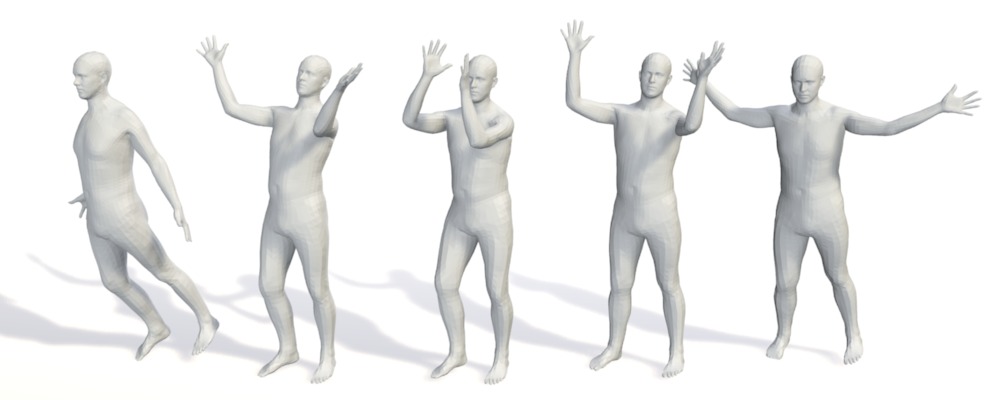}
\includegraphics[width=0.9\linewidth]{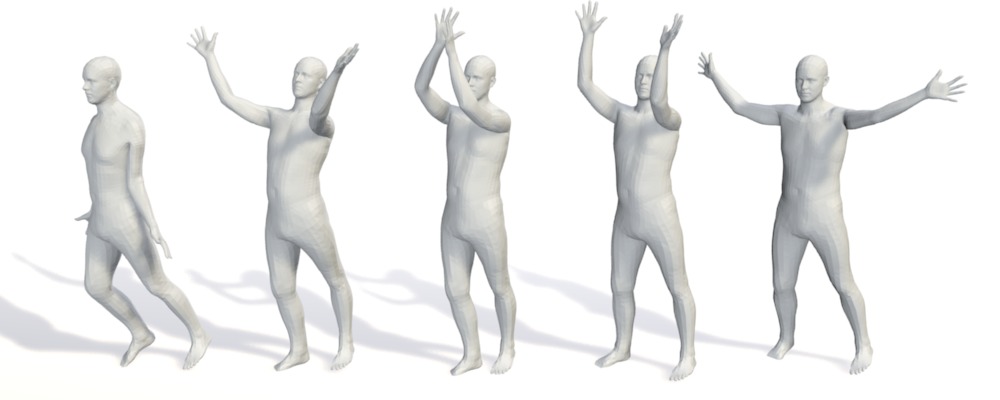}
\includegraphics[width=0.9\linewidth]{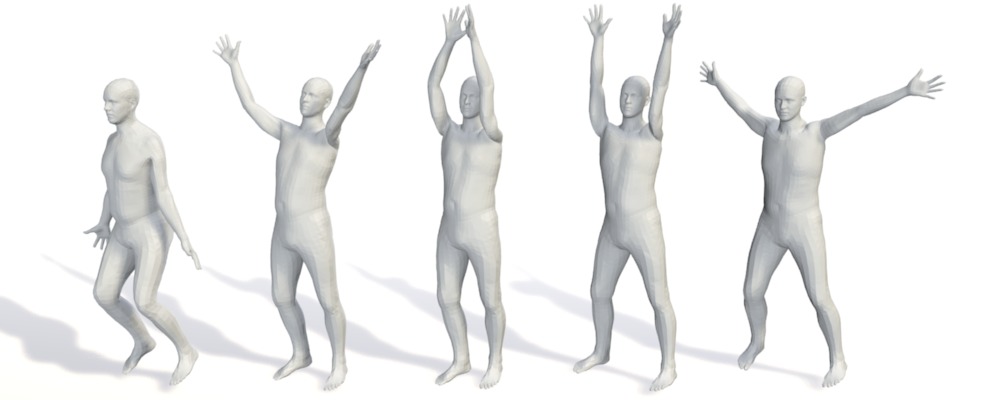}
\caption{We show three iterations of the optimization of $E_{motion}$ for a jumping jack sequence. First row: images of the scene, second row: pose initialization obtained by minimizing orientation and anthropometric consistency, third row: intermediate iteration, forth row: result of SIP, i.e. final pose estimates after convergence.}
\label{fig:optimizationIterations}
\end{figure}

In order to compute a descent update step to minimize $E_\mathrm{motion}$, we can now simply stack the linearized residual terms for all frames. For orientation and anthropometric terms this leads to sparse equations with the following block-diagonal structure
\begin{equation}
\begin{bmatrix}
\ddots  &  &   &   &  \\
 & \mat{J}_{t-1} &  &  &  \\
 &  & \mat{J}_{t} &  &  \\
   & &  & \mat{J}_{t+1} &  \\
 &       &        &     &  \ddots
\end{bmatrix}
\begin{bmatrix}
\vdots  \\
 \delta \mat{x}_{t-1}  \\
\delta \mat{x}_{t}  \\
\delta \mat{x}_{t+1}  \\
\vdots
\end{bmatrix} =
\begin{bmatrix}
\vdots  \\
 \mat{e}({t-1})  \\
\mat{e}({t})  \\
\mat{e}({t+1})  \\
\vdots
\end{bmatrix},
\end{equation}
where $\mat{J}_t$ denotes the respective Jacobian of the residual term $\mat{e}(t)$ at time step $t$. Similarly, the linearized residual terms of the acceleration residuals can be combined to obtain
\begin{equation}
\begin{bmatrix}
\ddots  &  \ddots &   &   &  \\
 \ddots & -2 \mat{J}_{t-1} & \mat{J}_{t} &  &  \\
 & \mat{J}_{t-1} & -2\mat{J}_{t} & \mat{J}_{t+1} &  \\
   & & \mat{J}_{t} & -2\mat{J}_{t+1} &  \ddots \\
 &       &        &    \ddots  &  \ddots
\end{bmatrix}
\begin{bmatrix}
\vdots  \\
 \delta \vect{x}_{t-1}  \\
\delta \vect{x}_{t}  \\
\delta \vect{x}_{t+1}  \\
\vdots
\end{bmatrix} =
\begin{bmatrix}
\vdots  \\
 \mat{e}_\mathrm{acc}(t-1)  \\
\mat{e}_\mathrm{acc}(t)  \\
\mat{e}_\mathrm{acc}(t+1)  \\
\vdots
\end{bmatrix}.
\end{equation}
By stacking the respective linearized multi-frame residual terms, we can now simply solve for the parameter updates and iterate until convergence. Iteration results for a jumping jack sequence are illustrated in \figref{fig:optimizationIterations}.
\subsection{IMU placement}
\label{sec:imuPlacement}
\added{Our proposed Sparse Inertial Poser is capable of recovering human motion from only 6 IMUs strapped to the lower legs, the lower arms, waist and head, see \figref{fig:sensorPlacement}. We found that this sensor configuration constrains a large number of pose parameters and produces good quantitative and qualitative results (see the supplemental video). An alternative sensor configuration would be to move the lower-leg and lower-arm IMUs to the end-effectors, i.e. feet and hands. Theoretically, this would constraint all joint parameters of the human body. However, we found that this adds too much uncertainty along the kinematic chain structure and results in worse performance than the proposed sensor placement.}

%% file: images/ori_posemanifold.pdf_tex
\begingroup%
  \makeatletter%
  \providecommand\color[2][]{%
    \errmessage{(Inkscape) Color is used for the text in Inkscape, but the package 'color.sty' is not loaded}%
    \renewcommand\color[2][]{}%
  }%
  \providecommand\transparent[1]{%
    \errmessage{(Inkscape) Transparency is used (non-zero) for the text in Inkscape, but the package 'transparent.sty' is not loaded}%
    \renewcommand\transparent[1]{}%
  }%
  \providecommand\rotatebox[2]{#2}%
  \ifx\svgwidth\undefined%
    \setlength{\unitlength}{619.12890625bp}%
    \ifx\svgscale\undefined%
      \relax%
    \else%
      \setlength{\unitlength}{\unitlength * \real{\svgscale}}%
    \fi%
  \else%
    \setlength{\unitlength}{\svgwidth}%
  \fi%
  \global\let\svgwidth\undefined%
  \global\let\svgscale\undefined%
  \makeatother%
  \begin{picture}(1,0.95833336)%
    \put(0,0){\includegraphics[width=\unitlength]{ori_posemanifold.pdf}}%
    \put(0.18361384,0.35917905){\color[rgb]{0,0,0}\rotatebox{-36.0135674}{\makebox(0,0)[lt]{\begin{minipage}{0.31024367\unitlength}\raggedright $frames$\end{minipage}}}}%
  \end{picture}%
\endgroup%

%% file: images/acc_posemanifold.pdf_tex
\begingroup%
  \makeatletter%
  \providecommand\color[2][]{%
    \errmessage{(Inkscape) Color is used for the text in Inkscape, but the package 'color.sty' is not loaded}%
    \renewcommand\color[2][]{}%
  }%
  \providecommand\transparent[1]{%
    \errmessage{(Inkscape) Transparency is used (non-zero) for the text in Inkscape, but the package 'transparent.sty' is not loaded}%
    \renewcommand\transparent[1]{}%
  }%
  \providecommand\rotatebox[2]{#2}%
  \ifx\svgwidth\undefined%
    \setlength{\unitlength}{607.1734375bp}%
    \ifx\svgscale\undefined%
      \relax%
    \else%
      \setlength{\unitlength}{\unitlength * \real{\svgscale}}%
    \fi%
  \else%
    \setlength{\unitlength}{\svgwidth}%
  \fi%
  \global\let\svgwidth\undefined%
  \global\let\svgscale\undefined%
  \makeatother%
  \begin{picture}(1,0.9721029)%
    \put(0,0){\includegraphics[width=\unitlength]{acc_posemanifold.pdf}}%
    \put(0.18538858,0.35008019){\color[rgb]{0,0,0}\rotatebox{-36.0135674}{\makebox(0,0)[lt]{\begin{minipage}{0.31635248\unitlength}\raggedright $frames$\end{minipage}}}}%
  \end{picture}%
\endgroup%

%% file: sections/experiments.tex
We evaluate here the performance of SIP. In \sectref{sec:trackerSetup} we present details on the general tracking procedure and computation times. \sectref{sec:baselineTrackers} introduces two baseline trackers which we use to compare and evaluate the tracking performance. We provide a quantitative assessment on a publicly available data set in \sectref{sec:quantitativeResults} and present qualitative results on additional recordings in \sectref{sec:qualitativeResults}. We refer to the video for more results.
\subsection{Tracker Setup}
\label{sec:trackerSetup}
In order to reconstruct the full-body motion with our proposed SIP we require 
\begin{itemize}
\item A SMPL body model of the actor,
\item The initial pose at the beginning of the sequence
\item IMU sensor locations on the body.
\end{itemize}
Initial pose and sensor locations are required to determine the sensor to bone offsets $\mat{G}^{BS}$, see \sectref{sec:imus}. Since IMUs are attached to different locations on the body, we manually selected the SMPL vertices once, and use them as sensor locations for all actors and experiments. Initial poses for the quantitative assessment were provided by the TNT15 data set. For the outdoor recordings we simply asked the actor to pose upright with straight arms and legs at the beginning of each sequence. We obtained SMPL body models by fitting the SMPL template to laser scans. If laser scans are not available we can also run SIP with approximate body models estimated with the method of "bodies from words"~\cite{Streuber:SIGGRAPH:2016}. In this case shape is estimated from only height, weight and \added{15} user ratings of the actor body shape. 

The general tracking procedure then works as follows. Starting with the initial pose we optimize pose for every frame sequentially using the orientation
and anthropometric terms. \added{We call this method Sparse Orientation Poser (SOP) and we use it as a baseline later. The resultant pose trajectory from SOP serves as initialization for optimizing the full cost function defined in \eqnref{eq:Energy}. 
 As can be seen in \figref{fig:optimizationIterations}, optimizing orientation and anthropometric consistency terms already recovers the pose reasonably well. This step is important, since \eqnref{eq:Energy} is highly non-linear and we apply a local, gradient-based optimization approach. After initialization, we use a  standard Levenberg-Marquardt algorithm to optimize the full cost cost function and iterate until convergence.}

For all experiments, we use the same energy weighting parameters listed in \tabref{tb:weightingParams}, which have been determined empirically. The overall processing time for a 1000 frame sequence and 20 \added{cost function evaluations} on a quad-core Intel Core i7 3.5GHz CPU is 7.5 minutes using single-core, non-optimized MATLAB code. For each iteration the majority of time is spent on \added{updating the body model (14.4s)} and setting up the Jacobians \added{(3.3s)}, while solving the sparse equations for a \added{Levenberg-Marquardt} update step takes approximately 1.5s. 
\added{Parallelization of model updates and Jacobian entries on the GPU would drastically reduce computation time and we leave
it as future work.}
\begin{table}[h]
\centering
\begin{tabular}{|c|c|c|c|c|}
\hline 
$w_{ori}$ & $w_{acc}$ & $w_{anthro}$ & $w_{mahal}$ & $w_{limits}$ \\ 
\hline 
1 & 0.05 & 1 & 0.003 & 0.1 \\ 
\hline 
\end{tabular}
\caption{Weighting parameters of $E_{motion}$, which have been used for all experiments.}
\label{tb:weightingParams}
\end{table}
\subsection{Baseline Trackers}
\label{sec:baselineTrackers}
We compare our tracking results to two baseline methods:
\begin{itemize}
\item \emph{Sparse Orientation Poser} (SOP): Minimizes orientation and anthropomorphic consistency terms but disregards acceleration.
\item \emph{SIP using an alternative body model} (SIP-M): Identical to SIP, but uses a manually rigged body model.
\end{itemize} 
The estimated pose trajectory obtained by SOP is used as the initialization of our proposed SIP. The second baseline, the SIP-M, uses a body model provided along the TNT15 data set as depicted in \figref{fig:tntMesh}. It is a body model with manually placed joints and fewer pose parameters. Anatomical constraints are imposed by using hinge joints, e.g. for the knee. In total, the body model has 31 pose parameters and the manual rigging procedure is representative for models that have been used for tracking so far (e.g.\cite{vlasic2008articulated,Pons-Moll2011,Marcard2016,gall2009motion}). In contrast, the SMPL model of SIP uses a statistical model to estimate joint positions. Every joint has 3 DoFs and anatomical constraints are imposed with the covariance of joint parameters. By comparing SIP and SIP-M we 
\added{want to asses the significance of using a statistically learned body model in contrast to a typical hand-rigged one.}

We also experimented with a single-frame acceleration tracker, which combines the SOP approach with acceleration data using a Kalman filter (similarly as in \cite{vlasic2007practical,roetenberg2007moven} but with only 6 sensors). Unfortunately, only 6 IMUs do not provide sufficient constraints on the poses to prevent drift caused by acceleration. 
In all cases, the tracker got unstable and failed after a few frames.
\subsection{Quantitative Results}
\label{sec:quantitativeResults}
\begin{figure}
\centering
\includegraphics[width=\linewidth]{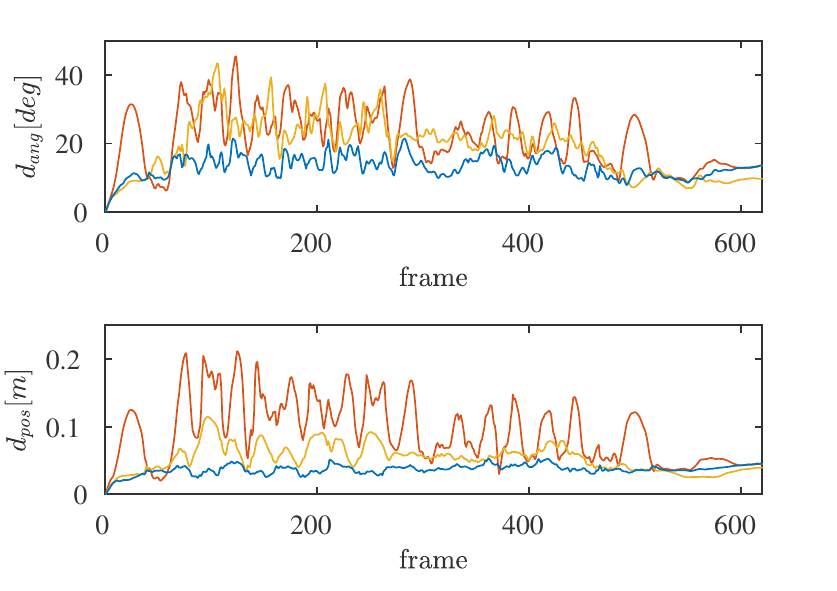}
\caption{Mean orientation and position error of a jumping jack sequence of the TNT15 data set. Our proposed SIP (blue) clearly outperforms both baseline trackers SOP (red) and SIP-M (yellow). }
\label{fig:sequenceErrors}
\end{figure}
For a quantitative analysis we evaluate the tracking performance of our proposed Sparse Inertial Poser (SIP) against the baseline trackers on the publicly available TNT15 data set published along \cite{Marcard2016}. This data set contains recordings of four subjects performing five activities each and provides inertial sensor data of 10 IMUs attached to lower legs, thighs, lower arms, upper arms, waist and chest. Additionally, multi-view video is provided which we only use for visualization purposes. Similar to \cite{Marcard2016} we split the 10 IMUs into tracking and validation sets. IMUs attached to lower legs, lower arms, waist and chest are used for tracking and the other IMUs serve as validation sensors. 

In order to evaluate the tracking performance we define two error metrics. On the one hand we use the mean orientation error $d_{ori}$ of the $N_v=4$ validation IMUs
\begin{equation}
d_{ori} = \frac{1}{T N_v} \sum_{t = 1}^T  \sum_{n = 1}^{N_v} || \vect{e}_{ori,n}(t) ||^2,
\end{equation}
where $\vect{e}_{ori,n}$ is defined in \eqnref{eq:OriResidual} and $T$ is the number of frames of the respective sequence. Second we compare the mean position error $d_{pos}$ of $N_m=13$ virtual markers on the body model
\begin{equation}
d_{pos} = \frac{1}{T N_m} \sum_{t = 1}^T  \sum_{n = 1}^{N_m} || \vect{p}_{n}(t)-\hat{\vect{p}}_{n}(t) ||^2
\end{equation}
where $\vect{p}$ is considered as ground-truth marker position obtained by tracking with all 10 IMUs and $\hat{\vect{p}}$ is the estimated marker position based on the estimated poses. The virtual marker positions comprise the SMPL-model joint locations of hips, knees, ankles, shoulders, elbows, wrists and neck. Since we cannot obtain \added{stable} ground-truth global translation from 10 IMUs alone, we set it to zero for calculating $d_{pos}$.

The mean position error is a common metric in video-based human motion tracking benchmarks (e.g. HumanEva \cite{sigal2010humaneva}, Human3.6M \cite{h36m_pami}) and is partially complementary to the mean orientation error. While the joint locations might be perfect, a rotation about a bone's axis does not alter the position error. This is only visible in the orientation error. On the other hand, a vanishing orientation error of the 4 validation IMUs does not necessarily imply correct joint positions as the spine or end-effectors might be incorrectly oriented. Hence, tracking performance is considered good if both error metrics are small.

\begin{figure*}
\begin{center}
$
\begin{array}{cccccccc}
\includegraphics[width = 0.13\linewidth]{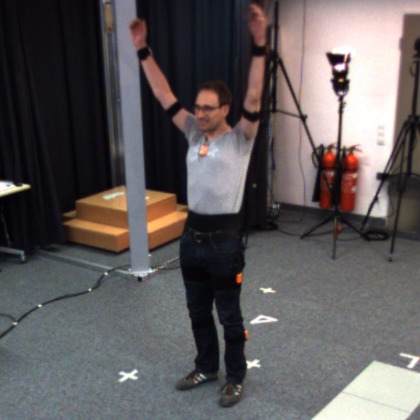} &
\includegraphics[width = 0.13\linewidth]{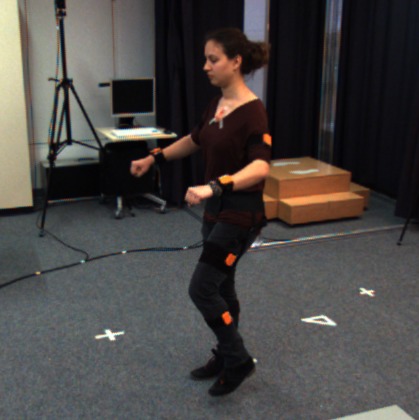} &
\includegraphics[width = 0.13\linewidth]{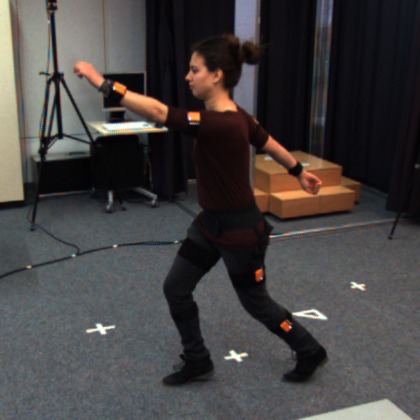} &
\includegraphics[width = 0.13\linewidth]{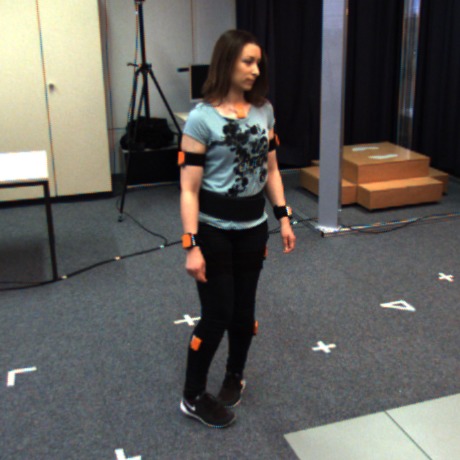} &
\includegraphics[width = 0.13\linewidth]{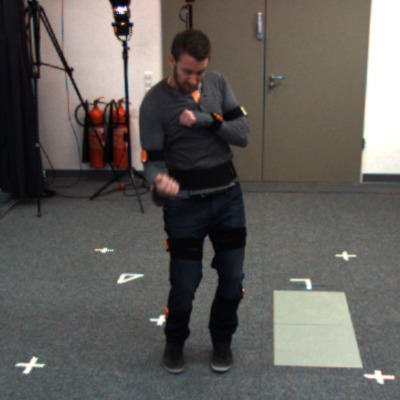} &
\includegraphics[width = 0.13\linewidth]{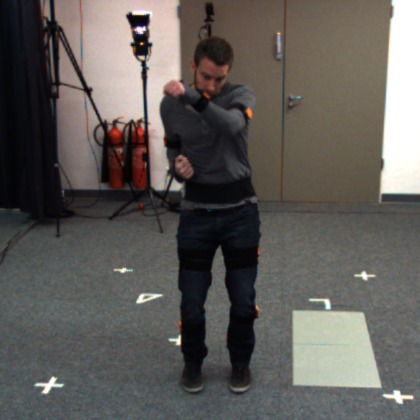} \\
\includegraphics[width = 0.13\linewidth]{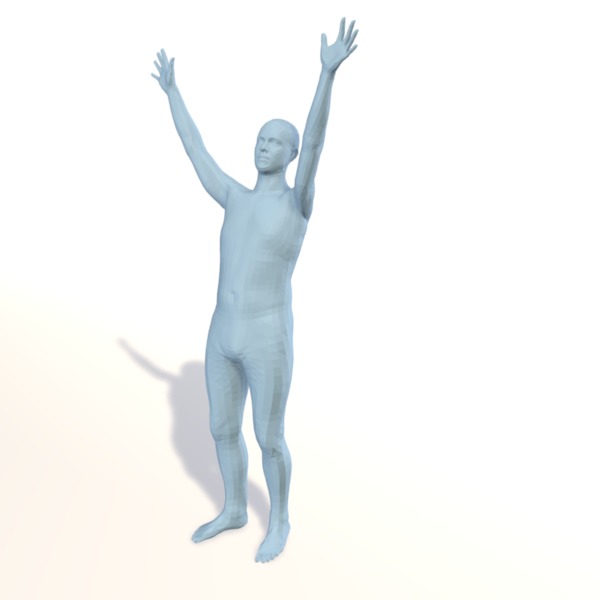} &
\includegraphics[width = 0.13\linewidth]{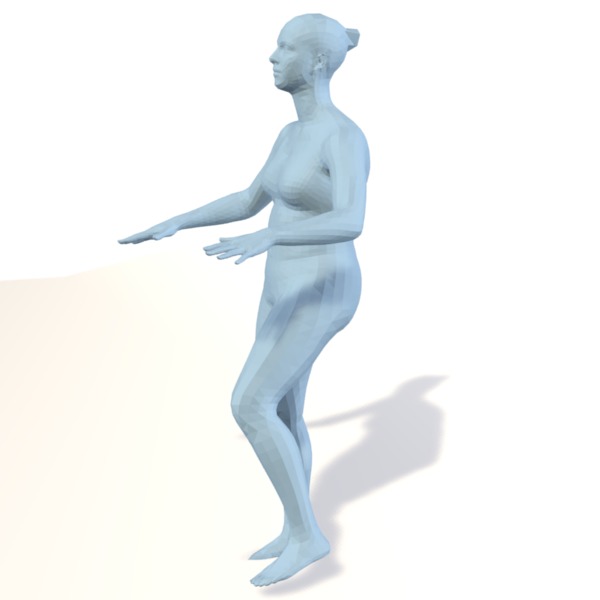} &
\includegraphics[width = 0.13\linewidth]{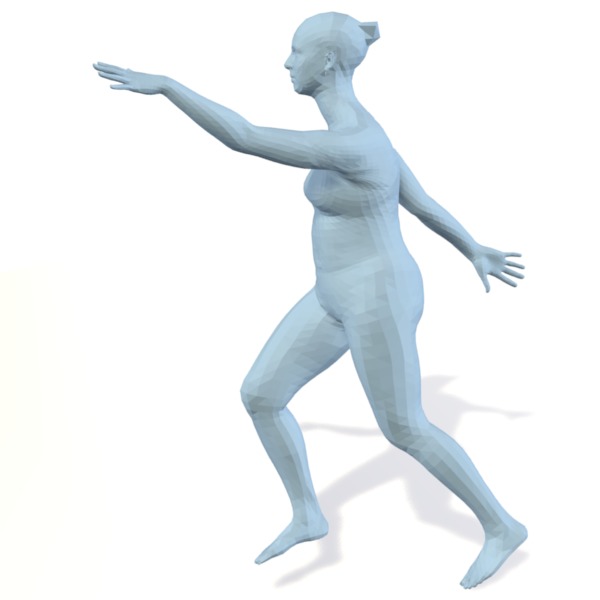} &
\includegraphics[width = 0.13\linewidth]{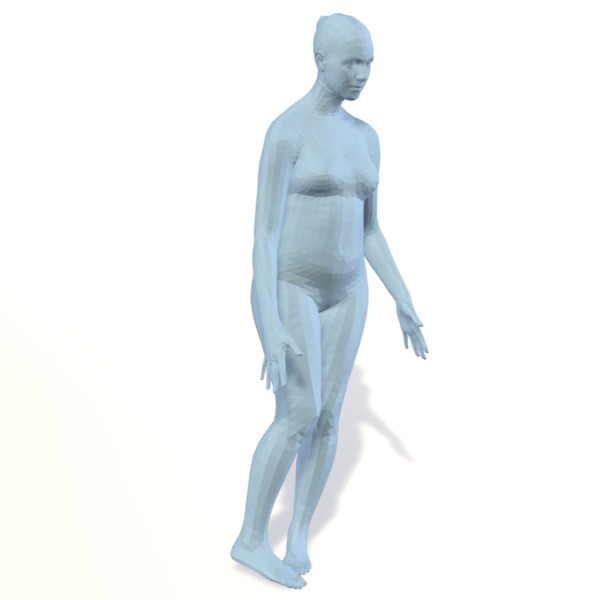} &
\includegraphics[width = 0.13\linewidth]{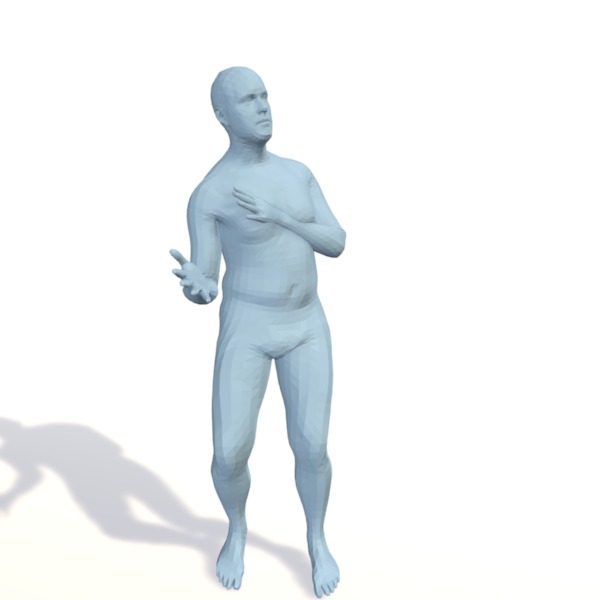} &
\includegraphics[width = 0.13\linewidth]{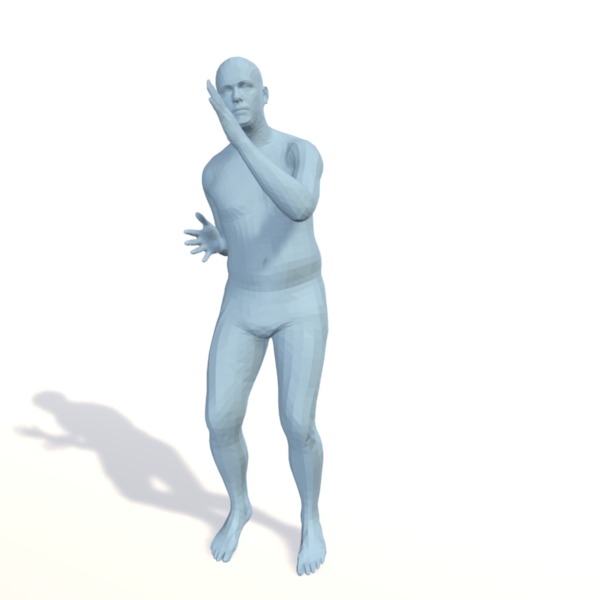}\\
\includegraphics[width = 0.13\linewidth]{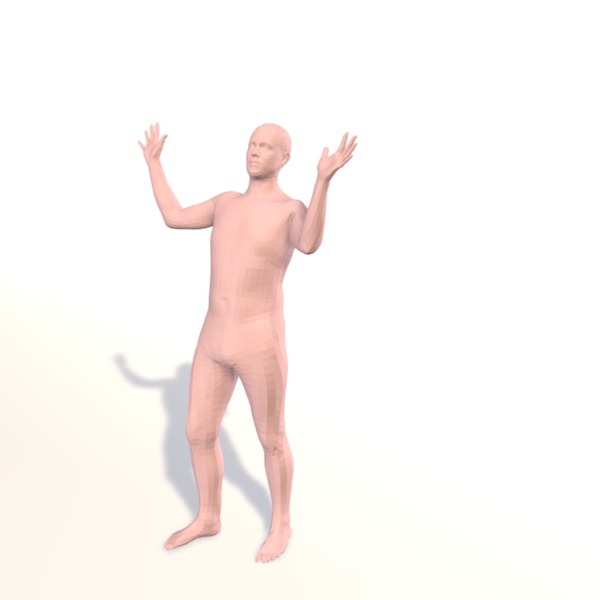} &
\includegraphics[width = 0.13\linewidth]{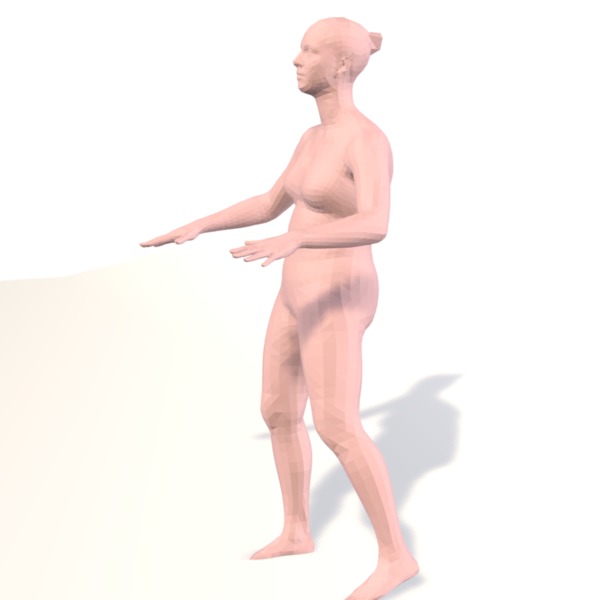} &
\includegraphics[width = 0.13\linewidth]{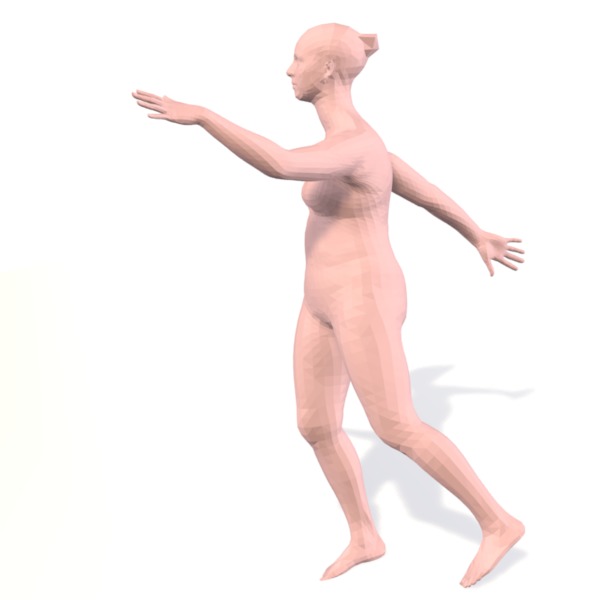} &
\includegraphics[width = 0.13\linewidth]{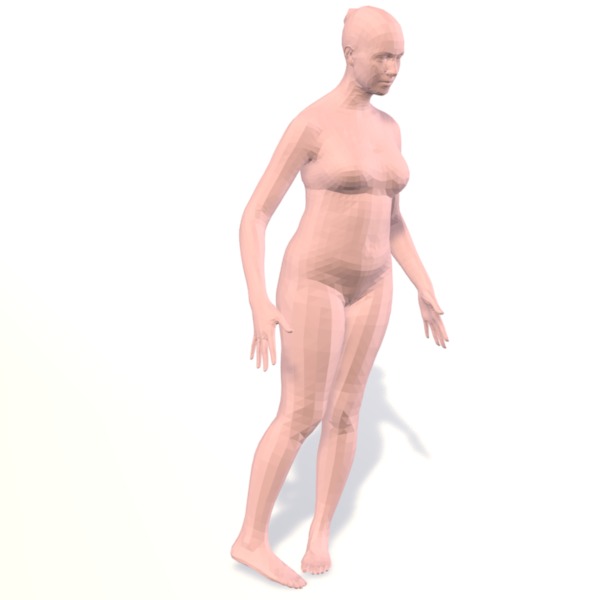} &
\includegraphics[width = 0.13\linewidth]{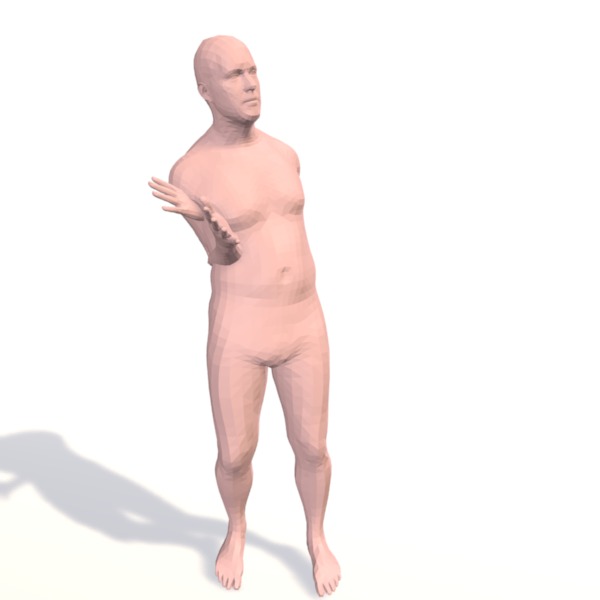} &
\includegraphics[width = 0.13\linewidth]{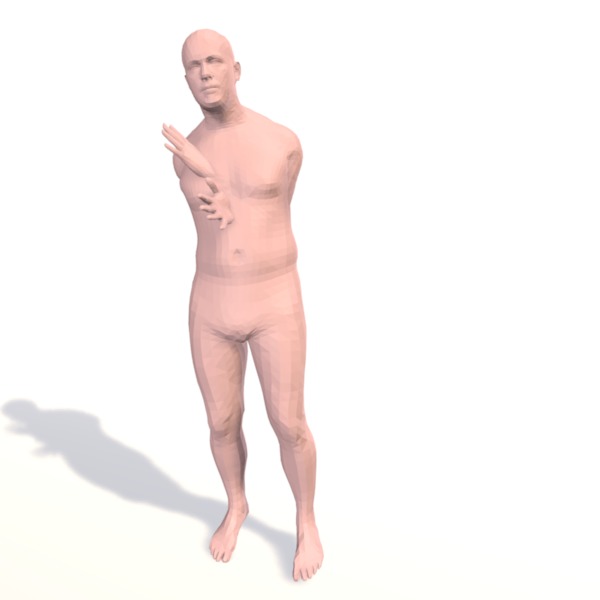}\\
\includegraphics[width = 0.13\linewidth]{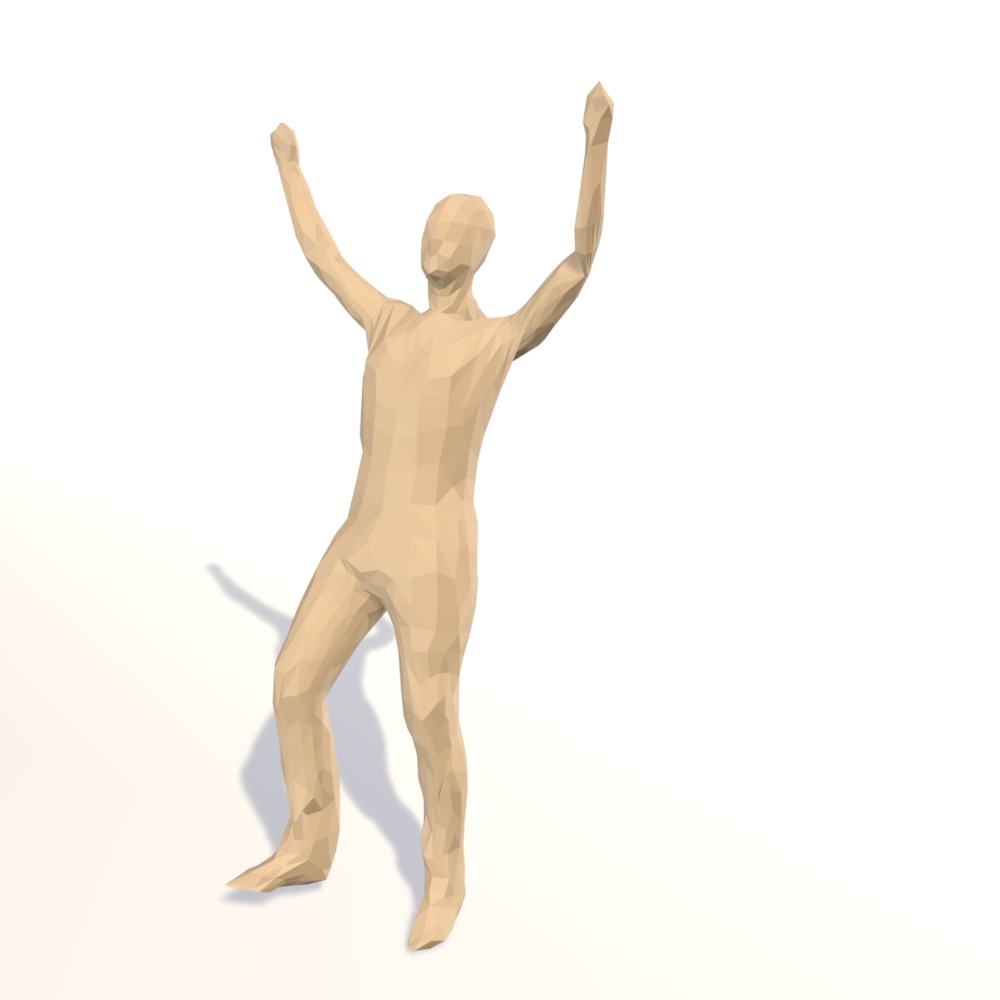} &
\includegraphics[width = 0.13\linewidth]{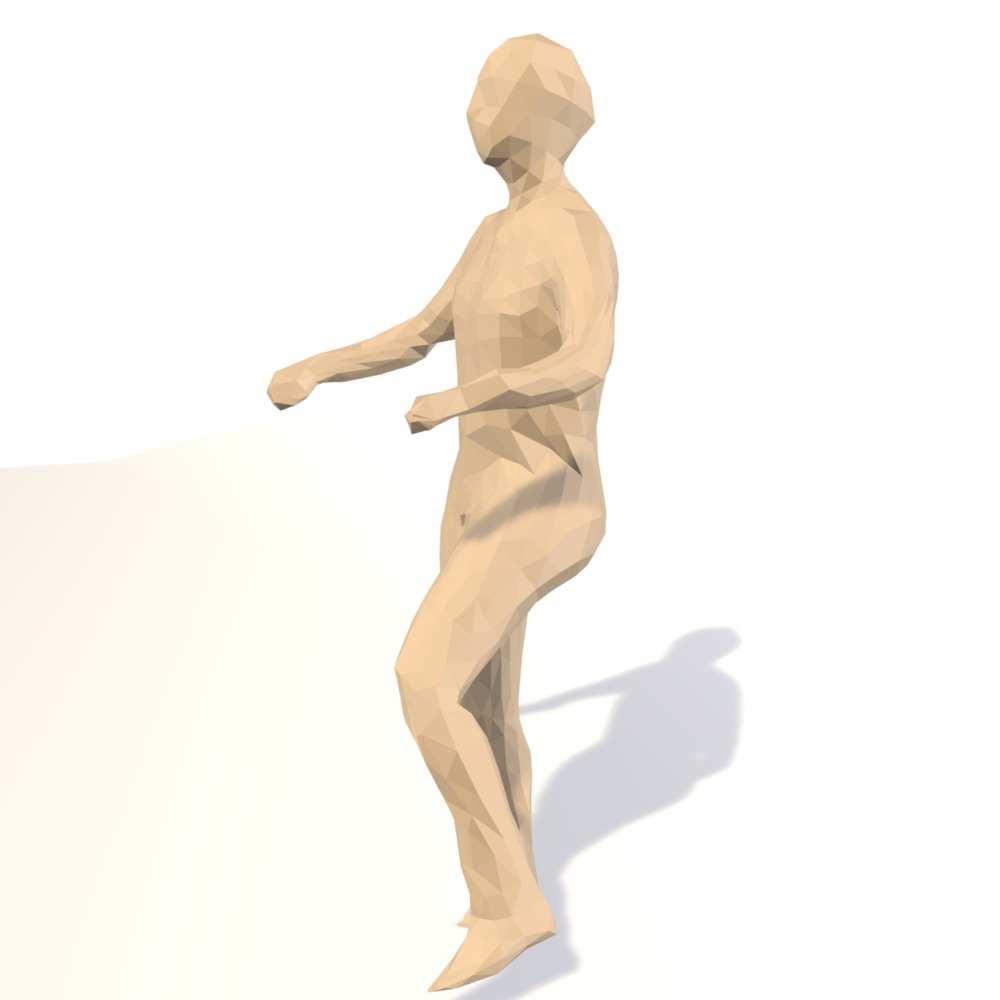} &
\includegraphics[width = 0.13\linewidth]{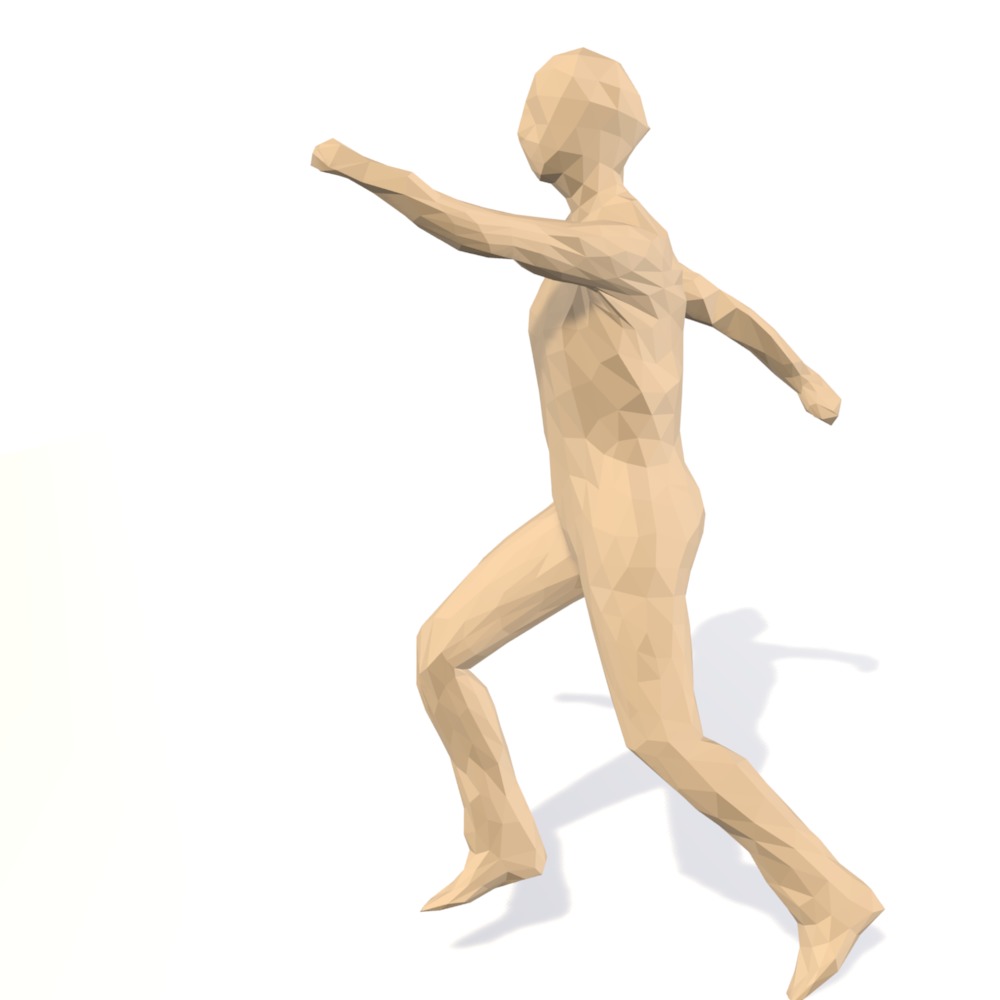} &
\includegraphics[width = 0.13\linewidth]{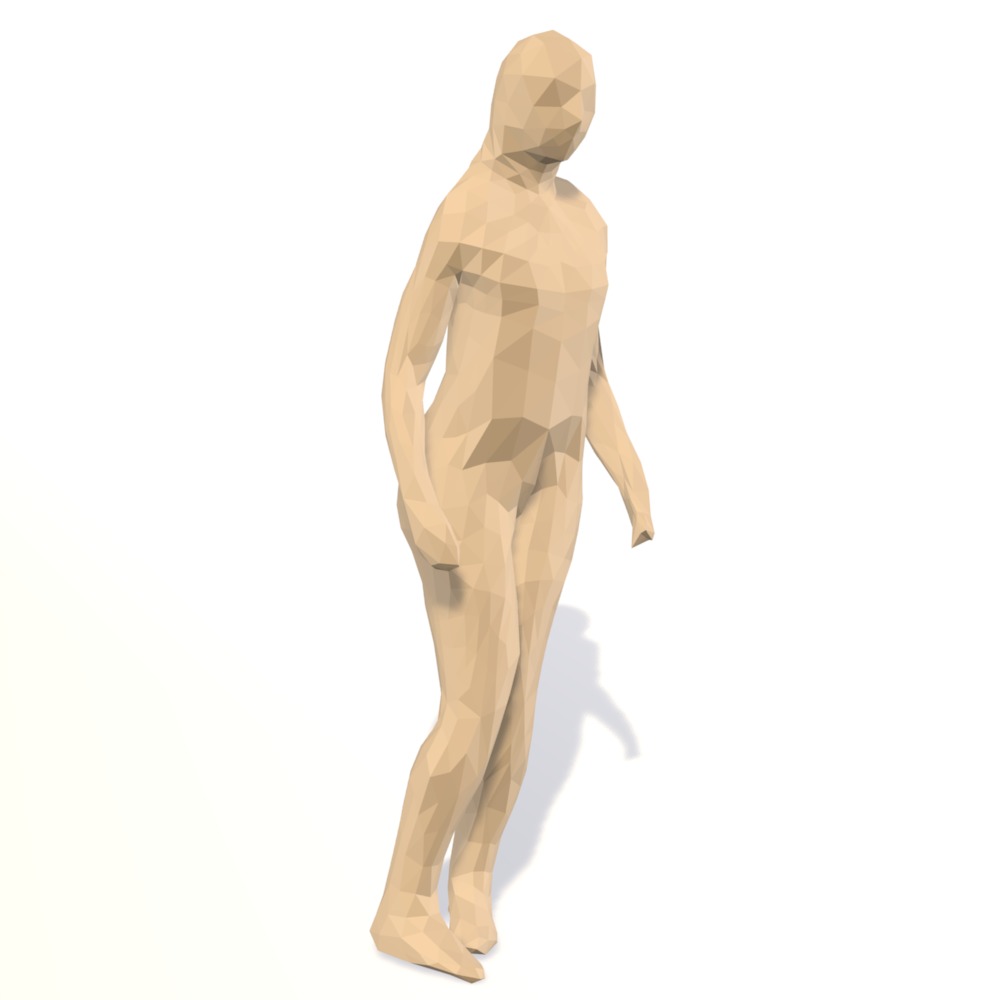} &
\includegraphics[width = 0.13\linewidth]{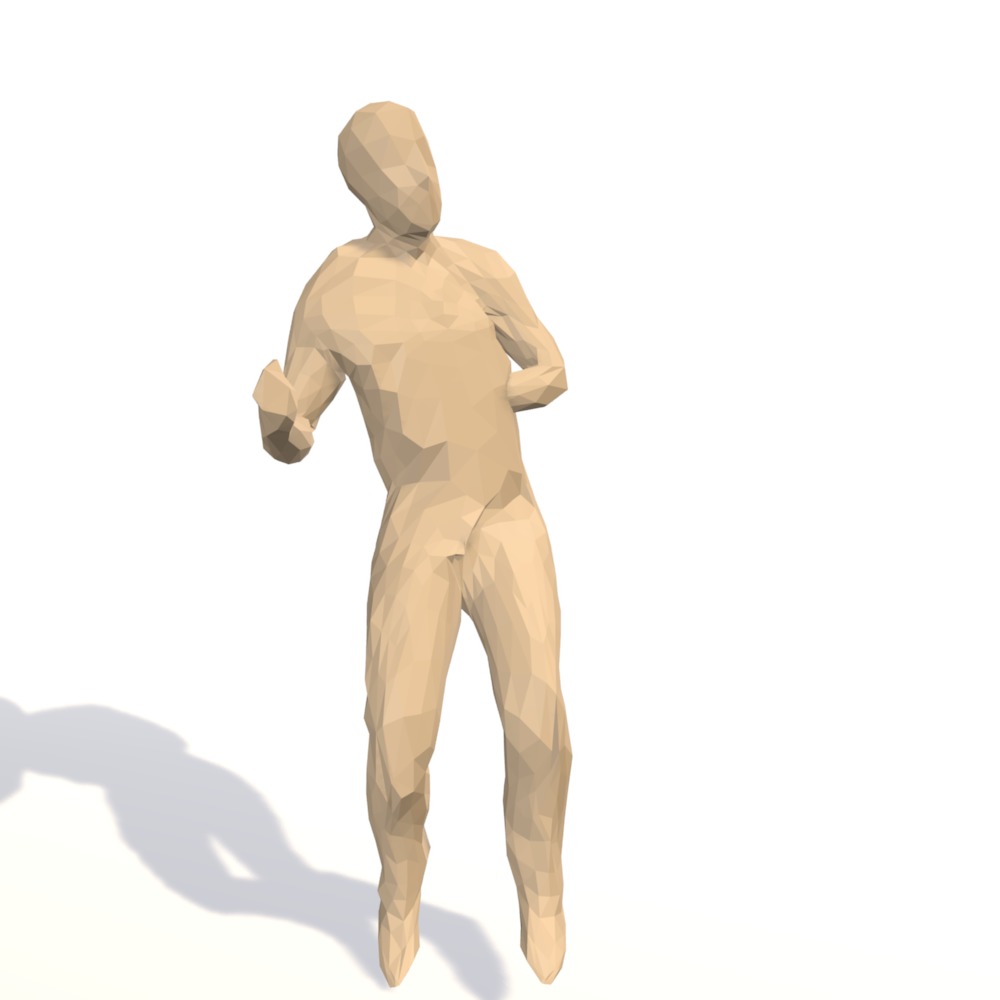} &
\includegraphics[width = 0.13\linewidth]{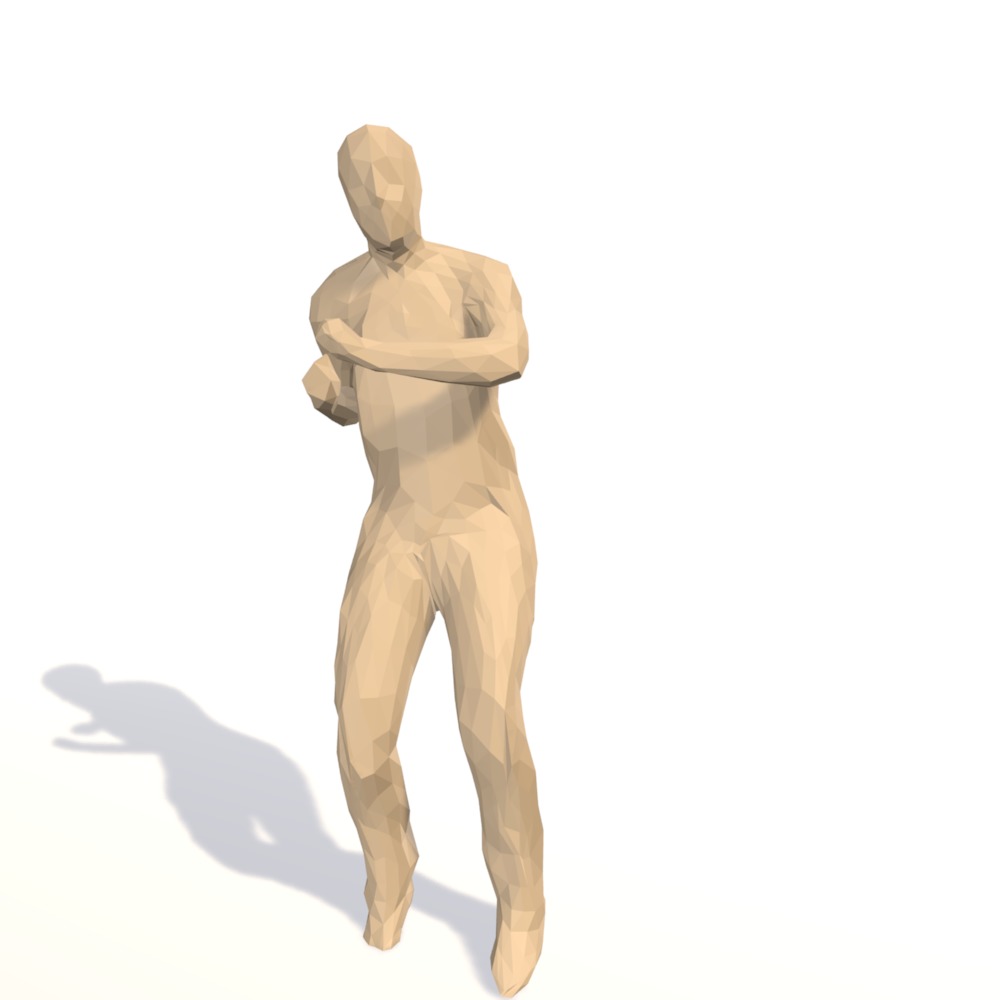} \\
\includegraphics[width = 0.13\linewidth]{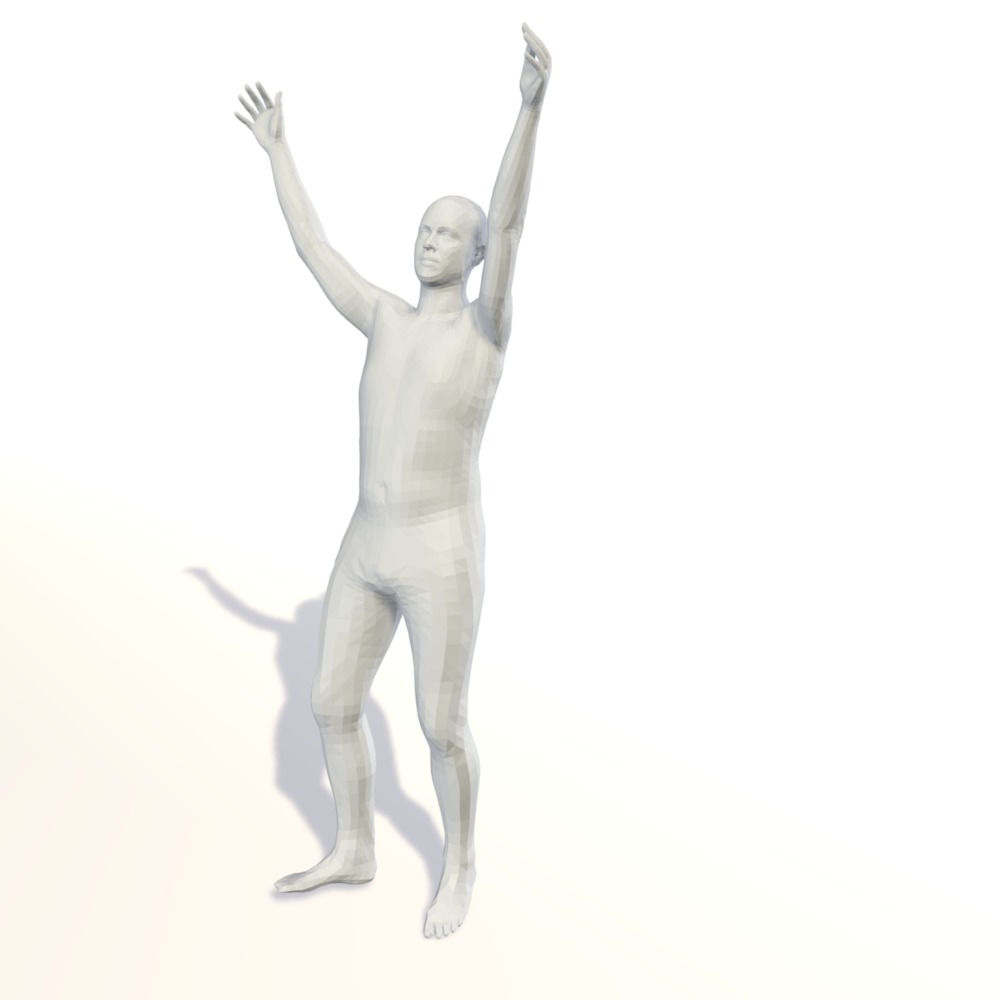} &
\includegraphics[width = 0.13\linewidth]{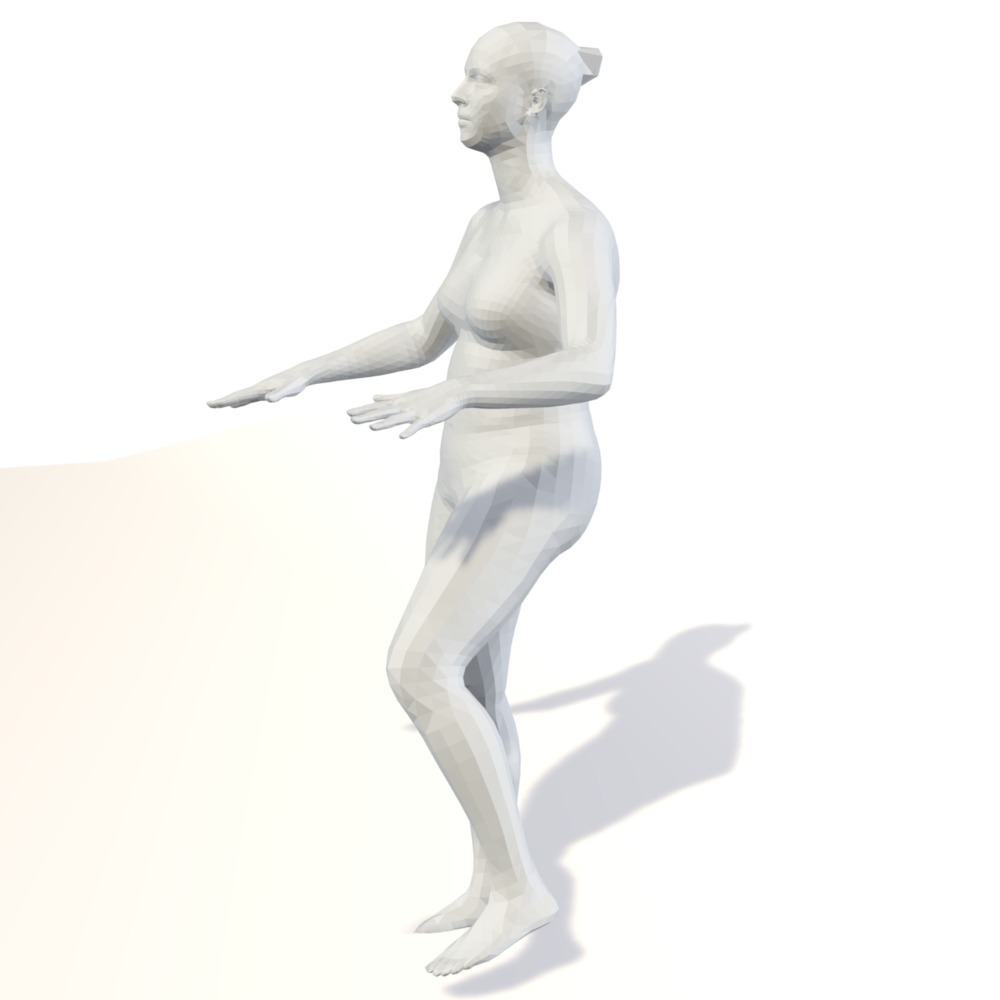} &
\includegraphics[width = 0.13\linewidth]{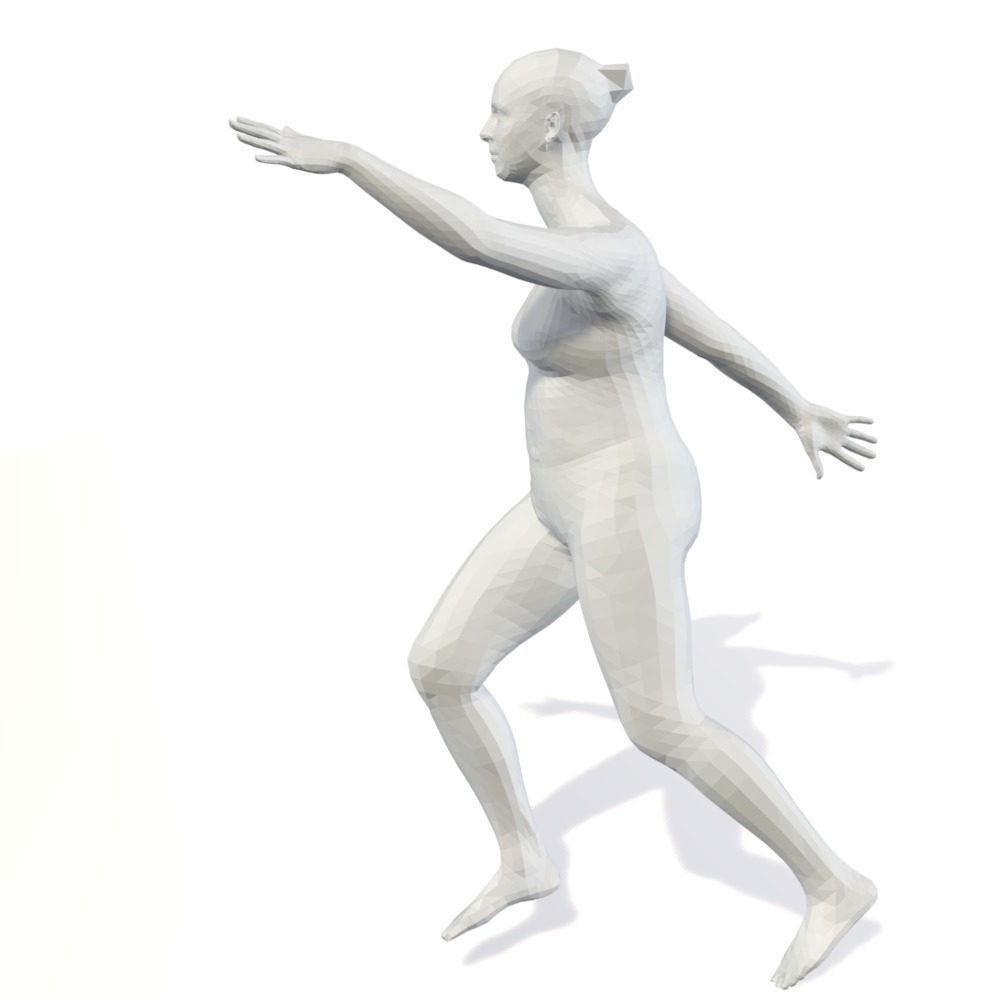} &
\includegraphics[width = 0.13\linewidth]{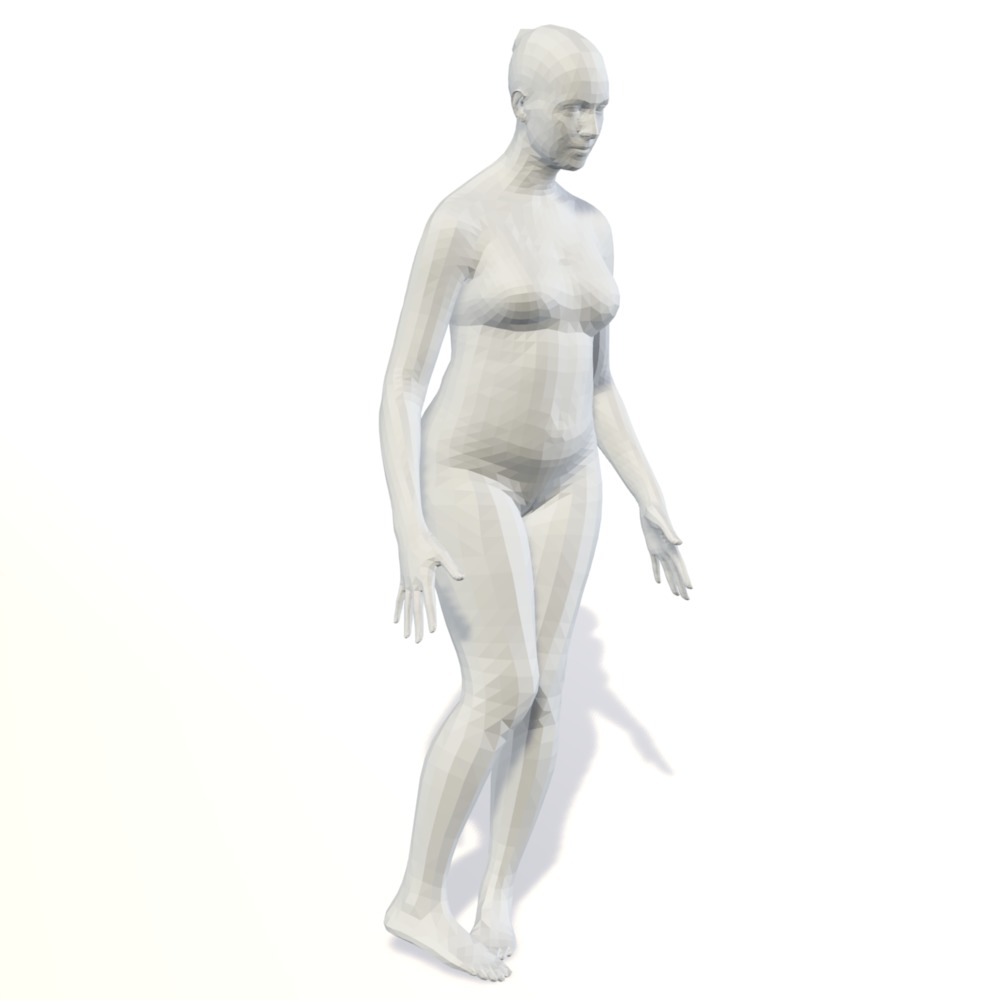} &
\includegraphics[width = 0.13\linewidth]{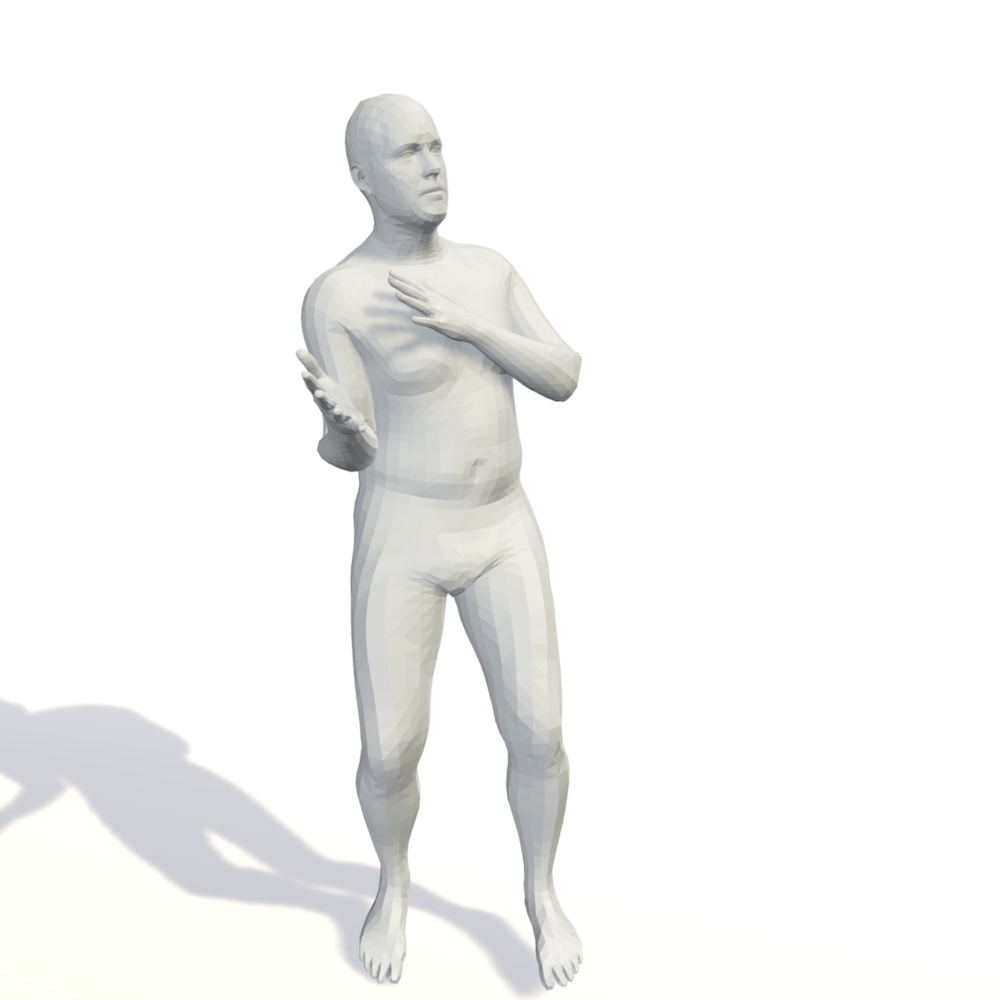} &
\includegraphics[width = 0.13\linewidth]{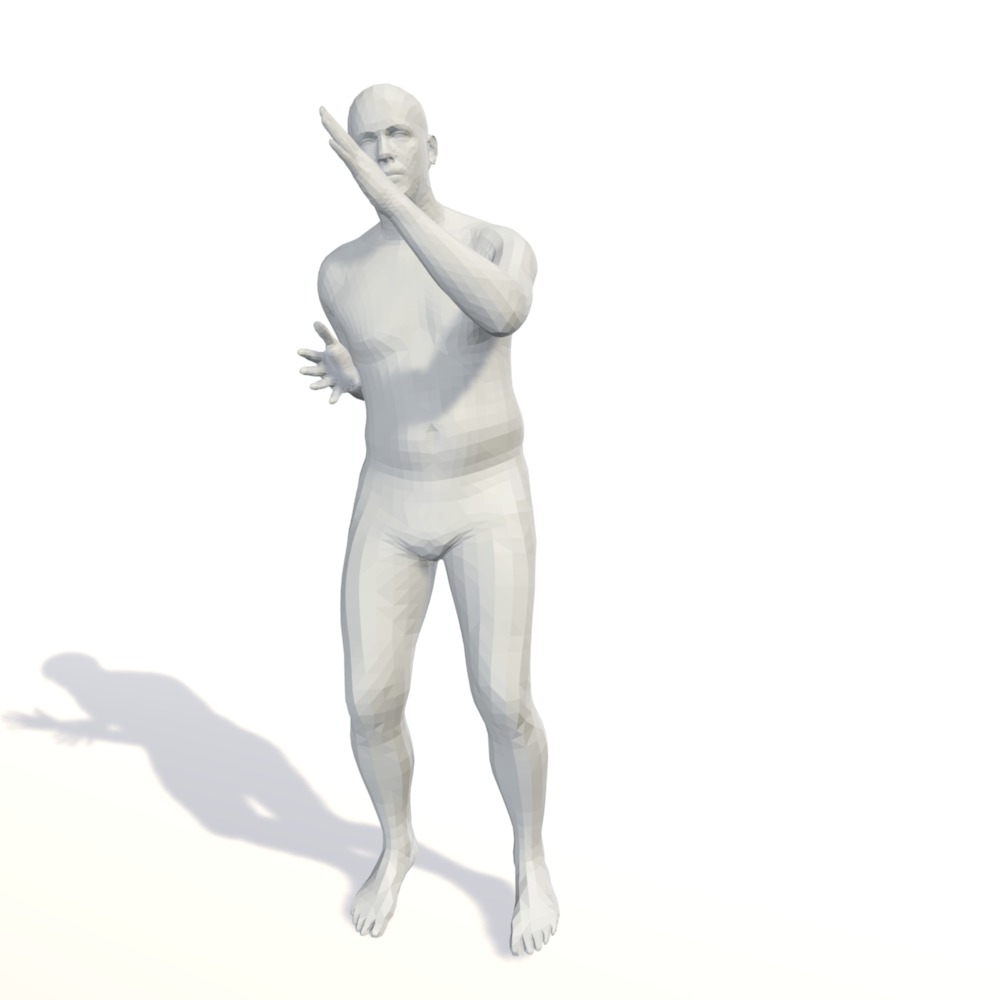}
\end{array}$
\end{center}
\caption{We compare our proposed SIP to ground truth and two baselines, the Sparse Orientation Poser (SOP), and our SIP with a manually rigged body model (SIP-M). Top row: images from the TNT dataset sequences,
second row: ground truth poses obtained by tracking with 10 IMUs (for reference), third row: results obtained with SOP, fourth row: results obtained with SIP-M and fifth row: results obtained with SIP. Best results are obtained with SIP. Without acceleration the pose remains ambiguous for the orientation poser (SOP) and leads to incorrect estimates, the SIP-M can disambiguate the poses by incorporating acceleration data but suffers from a limited skeletal model, which prevents the pose from appropriately fitting to the sensor data.
Differences are best seen in the supplemental video.}
\label{fig:tnt15Examples}
\end{figure*}

\figref{fig:sequenceErrors} shows the tracking errors for a jumping jack sequence of the TNT15 data set. This sequence contains extended arm and leg motions, also visible in \figref{fig:optimizationIterations}, as well as two foot stamps around frames 25 and 500. The SOP fails to accurately reconstruct these motions as orientation measurements of 6 IMUs are too ambiguous. This is easily illustrated for the case of a foot stamp, which can be seen in the second column of \figref{fig:tnt15Examples}. During this motion the lower leg is tilted, but without acceleration data it is impossible to infer whether the thigh was lifted at the same time. The SIP-M can resolve this ambiguity but the limited body model is not sufficiently expressive to accurately reconstruct the jumping jacks and skiing exercises. In contrast our proposed SIP shows low orientation and position errors for the whole sequence and clearly outperforms both baseline trackers.
 
\begin{figure}[ht]
\centering
\includegraphics[width=\linewidth]{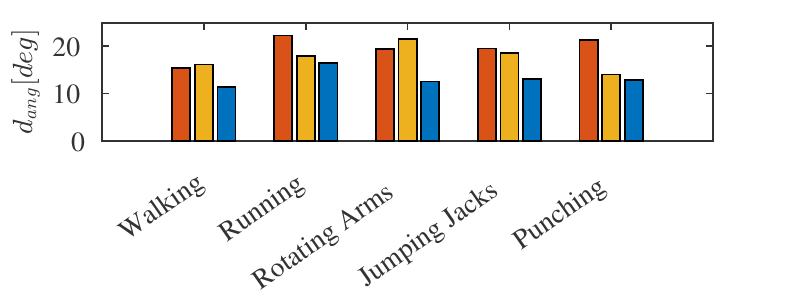}
\caption{Mean orientation error on the TNT15 data set: comparison of SOP(red), SIP-M(yellow) against our proposed SIP (blue).}
\label{fig:tntOrientationErrors}
\end{figure}
\begin{figure}[ht]
\centering
\includegraphics[width=\linewidth]{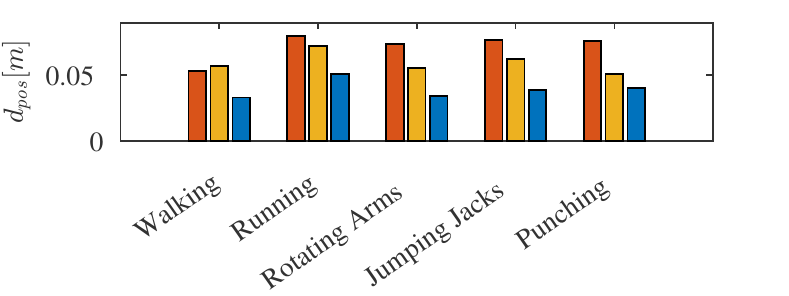}
\caption{Mean position error on the TNT15 data set: comparison of SOP(red), SIP-M(yellow) against our proposed SIP (blue). }
\label{fig:tntPositionErrors}
\end{figure}
\added{
\begin{table}[ht]
\centering
\begin{tabular}{ l c c c c}
Approach & $\mu_{ang}[deg]$ & $\sigma_{ang}[deg]$ & $\mu_{pos}[m]$ & $\sigma_{pos}[m]$ \\ 
\hline 
SOP & 19.64 & 17.35 & 0.072 & 0.089 \\
SIP-M & 18.24 & 15.82 & 0.06 & 0.053 \\
\textbf{SIP} & \textbf{13.32} & \textbf{10.13} & \textbf{0.039} & \textbf{0.04}  \\
SIP-BW & 13.45 & 9.94 & 0.042 & 0.04 \\
SIP-110 & 13.67 & 10.38 & 0.046 & 0.045 \\
SIP-120 & 14.27 & 10.6 & 0.056 & 0.053 
\end{tabular}
\caption{Tracking errors on TNT15.}
\label{tb:tntResults}
\end{table}}
The tracking result of the jumping jack sequence is exemplary for the overall tracking performances on the TNT15 data set. In \figref{fig:tntOrientationErrors} we show the average orientation error for all actors, separated by activities. Similarly, \figref{fig:tntPositionErrors} shows the mean position error.
Additionally, \tabref{tb:tntResults} shows the overall tracking errors on the TNT15 data set. We have added additional rows for SIP-BW\added{, SIP-110 and SIP-120. SIP-BW} is identical to SIP but uses a SMPL model estimated with the "bodies from words" approach. The tracking error difference is insignificant, which further improves applicability of SIP. Thus, we do not need the accuracy of a laser scan, making the proposed solution very easy to use. \added{SIP-110 and SIP-120 use a scaled version of the SIP body model, where body size was increased by 10\% and 20\% respectively. Again, the the tracking error remains comparably small and it further demonstrates that SIP is very robust to moderate variations in body shape.}
 
It is remarkable, that SIP-M and SIP achieve a mean orientation error of \added{$18.24^\circ$ and $13.32^\circ$}, respectively. \cite{Marcard2016} reported an average orientation error of $15.71^\circ$, using 5 IMUs and 8 cameras minimizing single-frame orientation and silhouette consistency terms. SIP-M uses the same body model and is just slightly worse. Using the SMPL body model in SIP results in an even smaller orientation error. Thus, without relying on visual cues of 8 cameras we achieve competitive orientation errors by simply taking IMU accelerations into account and optimizing over all frames simultaneously.

Quantitative results demonstrate that accurate full-body motion tracking with sparse IMU data becomes feasible by incorporating acceleration data. In comparison to the SOP which uses only orientation data, our proposed SIP reduces the mean orientation error on the TNT15 data set from \added{$19.64^\circ$ to $13.32^\circ$} and the mean position error decreases from \added{$7.2cm$ to $3.9cm$}. We have also shown that for our tracking approach, the statistically learned body model SMPL leads to more accurate tracking results than using a representative manually rigged body model. Further, the SMPL model can be even created using only linguistic ratings, which obviates the need for a laser scan of the person. In \figref{fig:tnt15Examples} we show several example frames of the tracking results obtained on the TNT15 data set.
\begin{figure}[t!]
\begin{center}
\includegraphics[height = 1.7cm]{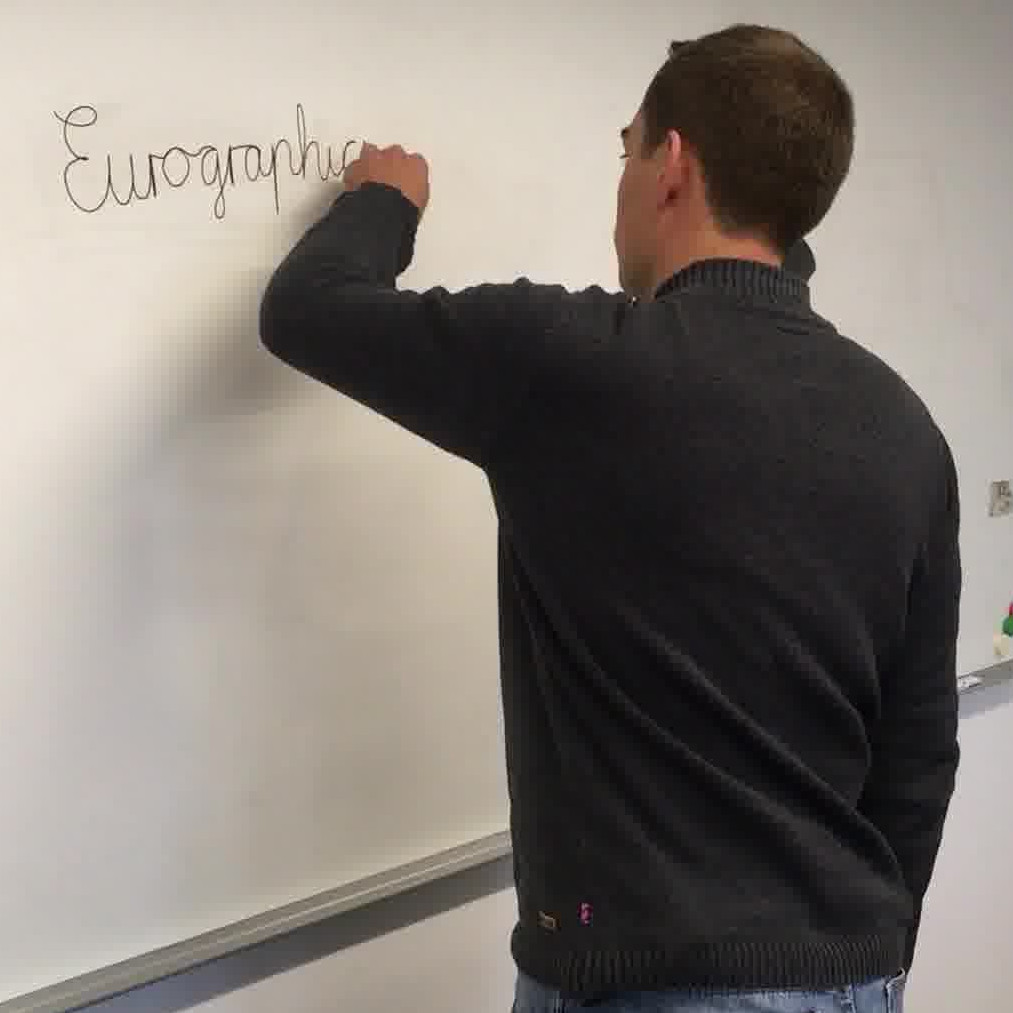} 
\includegraphics[height = 1.7cm]{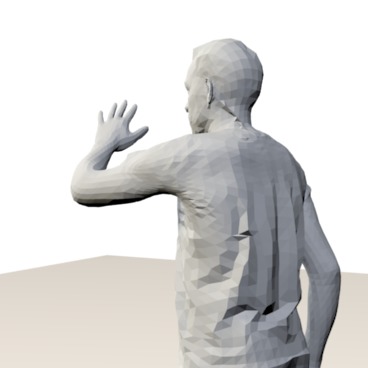}
\includegraphics[height = 1.7cm]{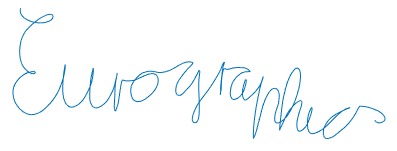}
\end{center}
\caption{SIP is capable of recovering the handwriting on a whiteboard. Left figure: image of the writing scene, middle figure: recovered pose at the end of the handwriting, right figure: recovered wrist motion projected on the whiteboard plane.}
\label{fig:whiteboardDrawing}
\end{figure}
\subsection{Qualitative Results}
\label{sec:qualitativeResults}
In order to further demonstrate the capabilities of our proposed SIP we recorded additional motions. For all recordings we have used 6 Xsens MTw IMUs \cite{Xsens} attached to the lower legs, wrists, head and back. The sensor placement is illustrated in \figref{fig:tntMesh}. Orientation and acceleration data were recorded at 60Hz and transmitted wirelessly to a laptop. Additionally, we have captured the motions with a smartphone camera to qualitatively assess the tracking accuracy.

\begin{figure*}
\begin{center}$
\begin{array}{ccccc}
\includegraphics[width=0.14\linewidth]{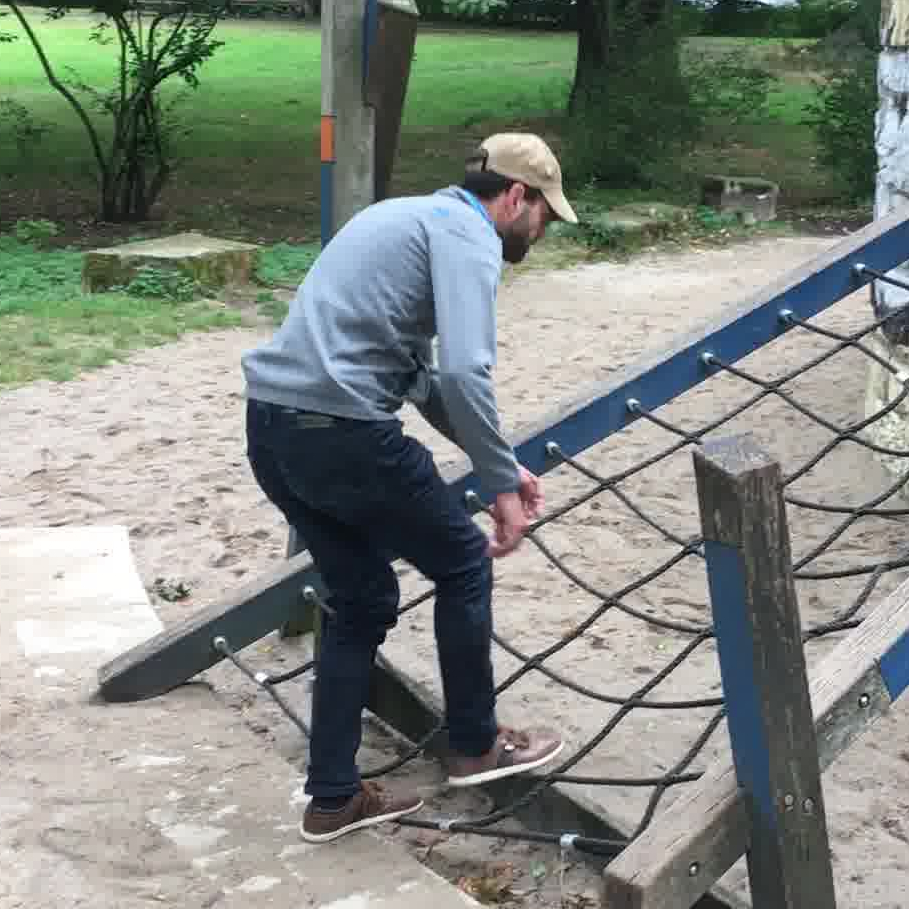} &
\includegraphics[width=0.14\linewidth]{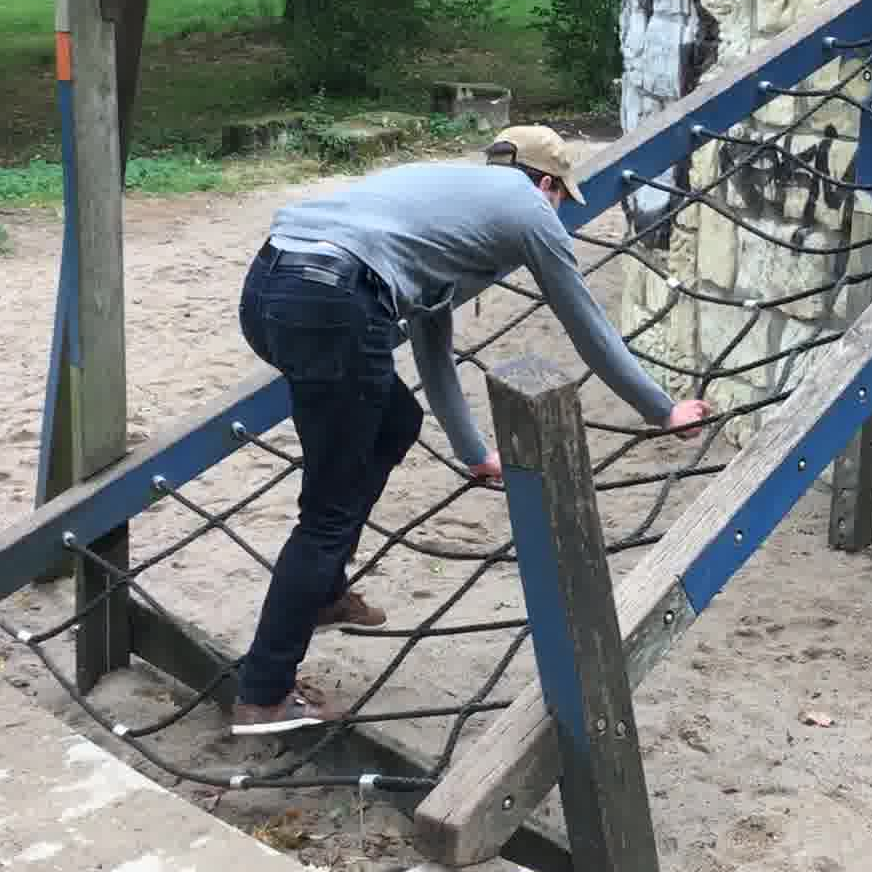} &
\includegraphics[width=0.14\linewidth]{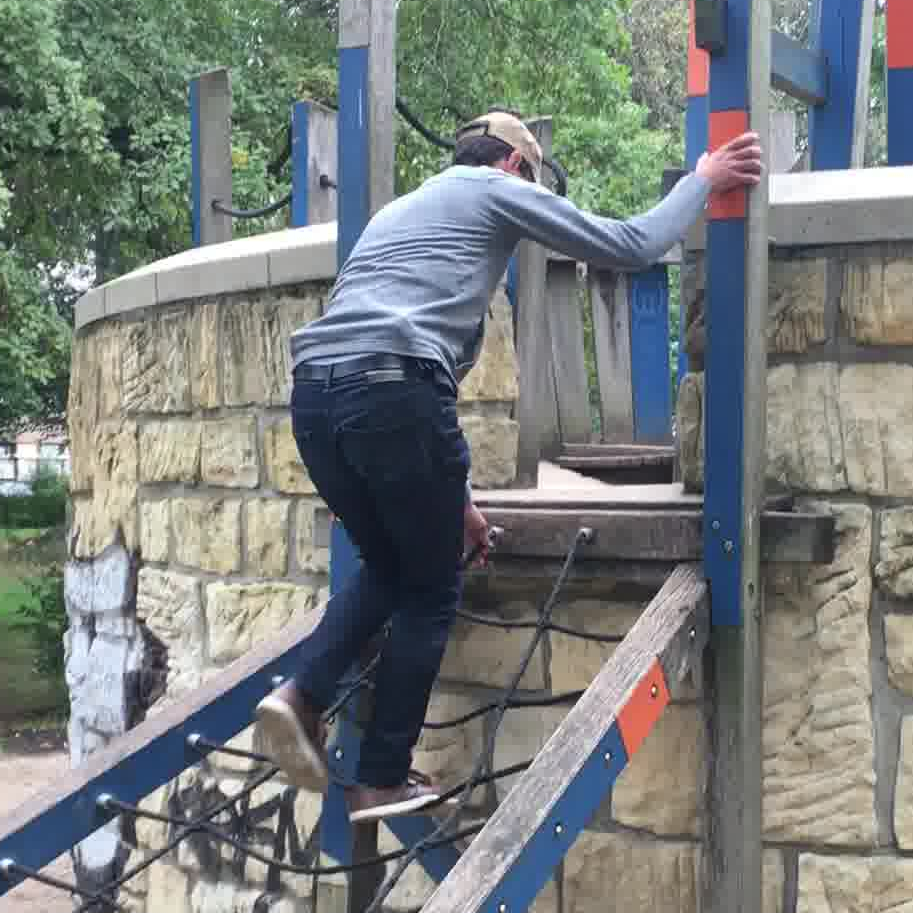} &
\includegraphics[width=0.14\linewidth]{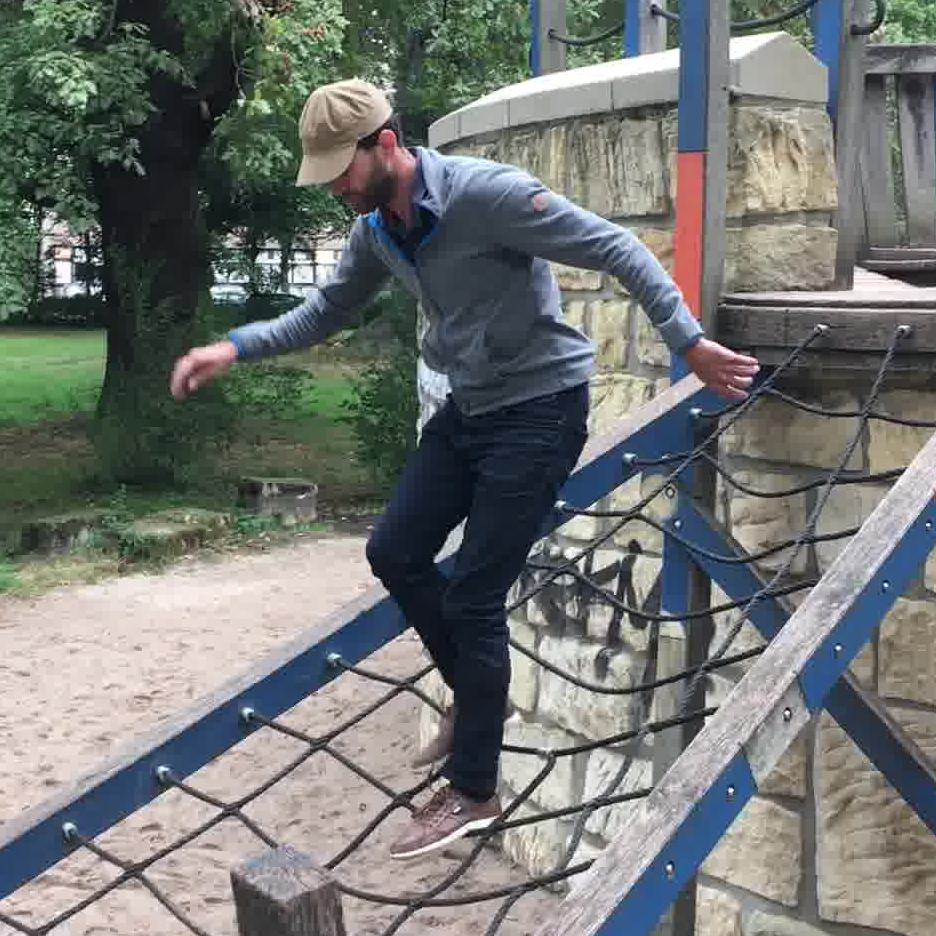} &
\includegraphics[width=0.14\linewidth]{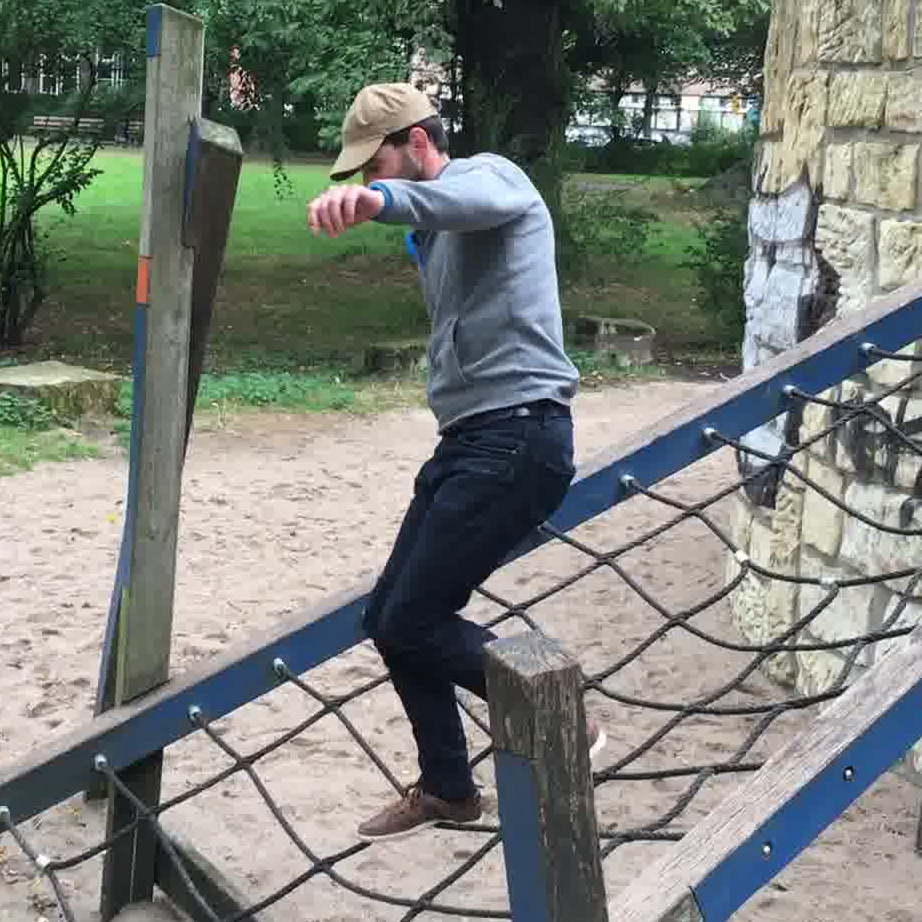} \\
\includegraphics[width=0.14\linewidth]{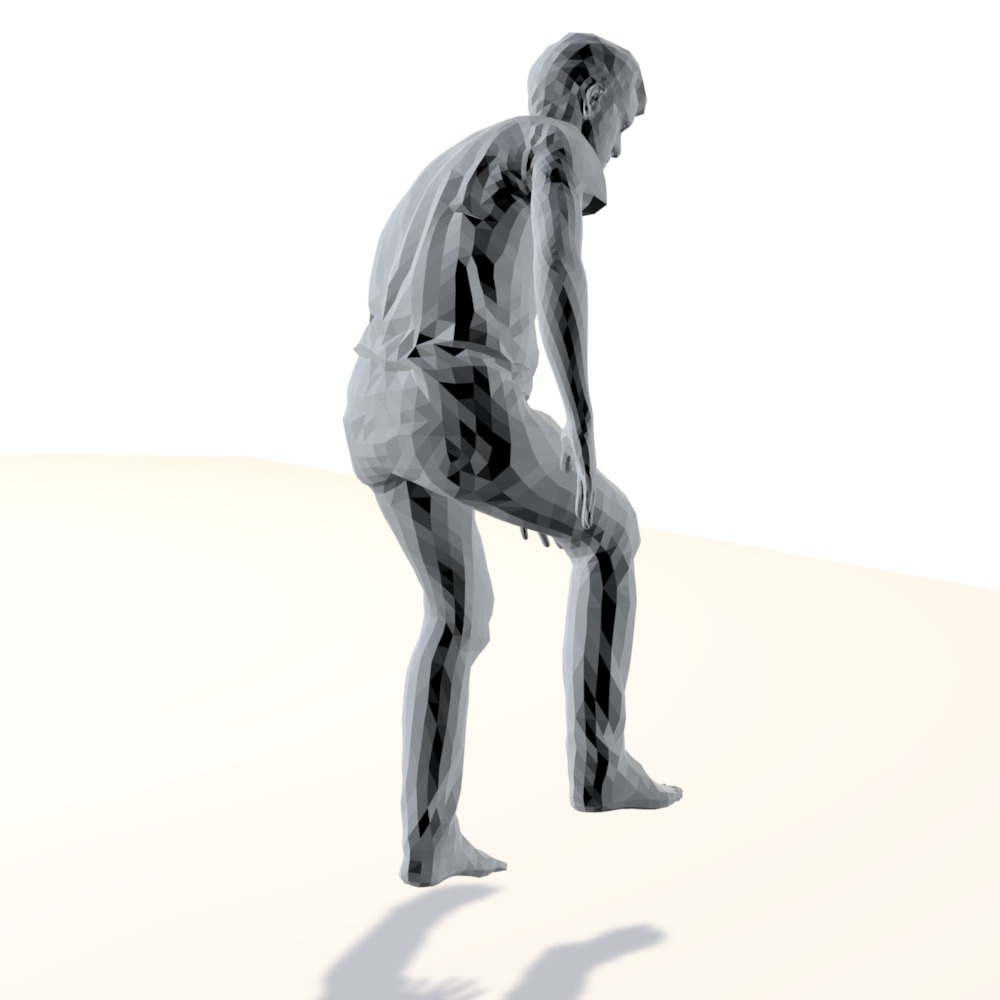} &
\includegraphics[width=0.14\linewidth]{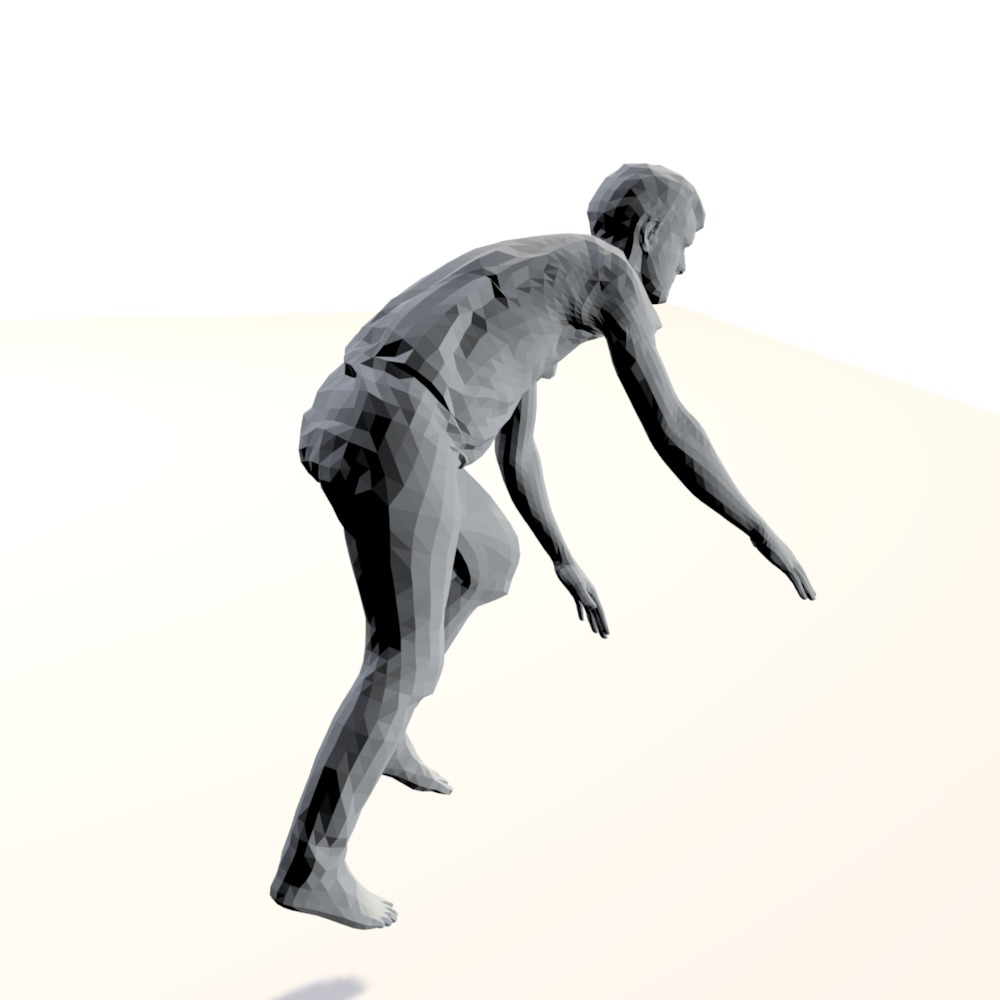} &
\includegraphics[width=0.14\linewidth]{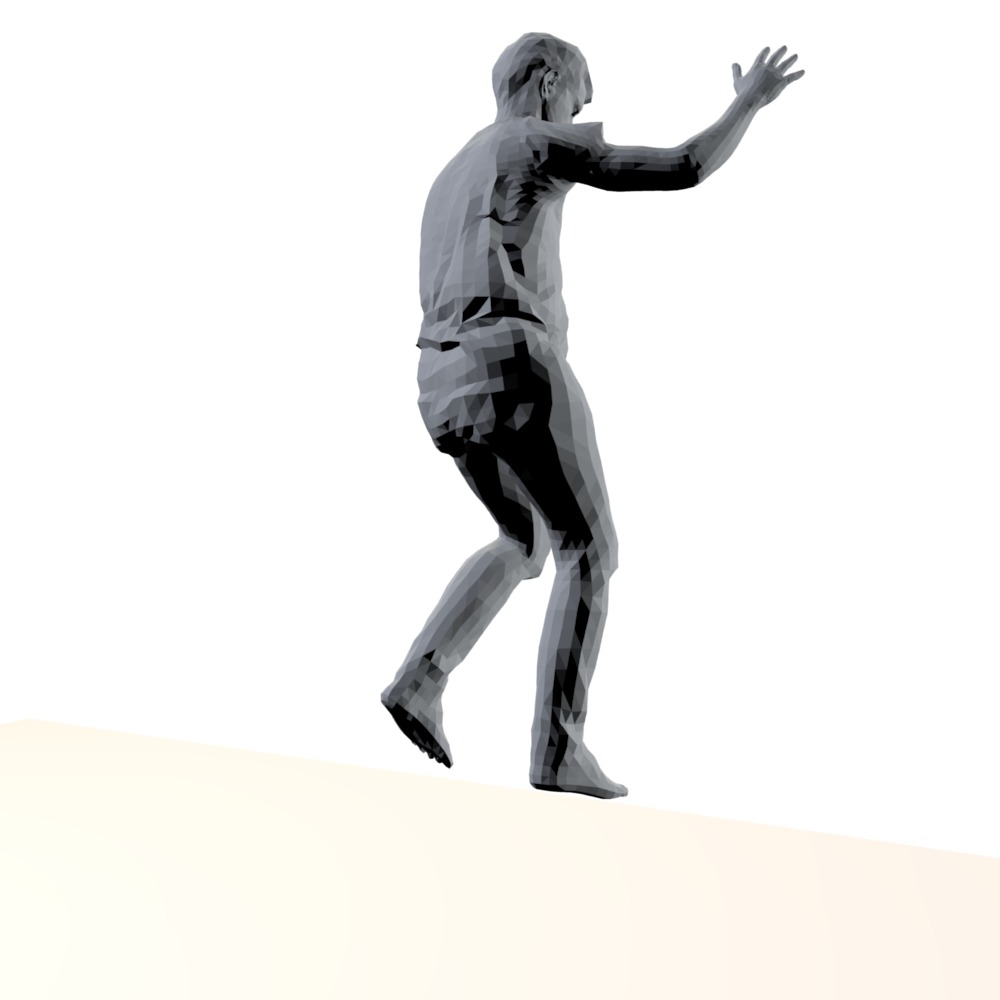} &
\includegraphics[width=0.14\linewidth]{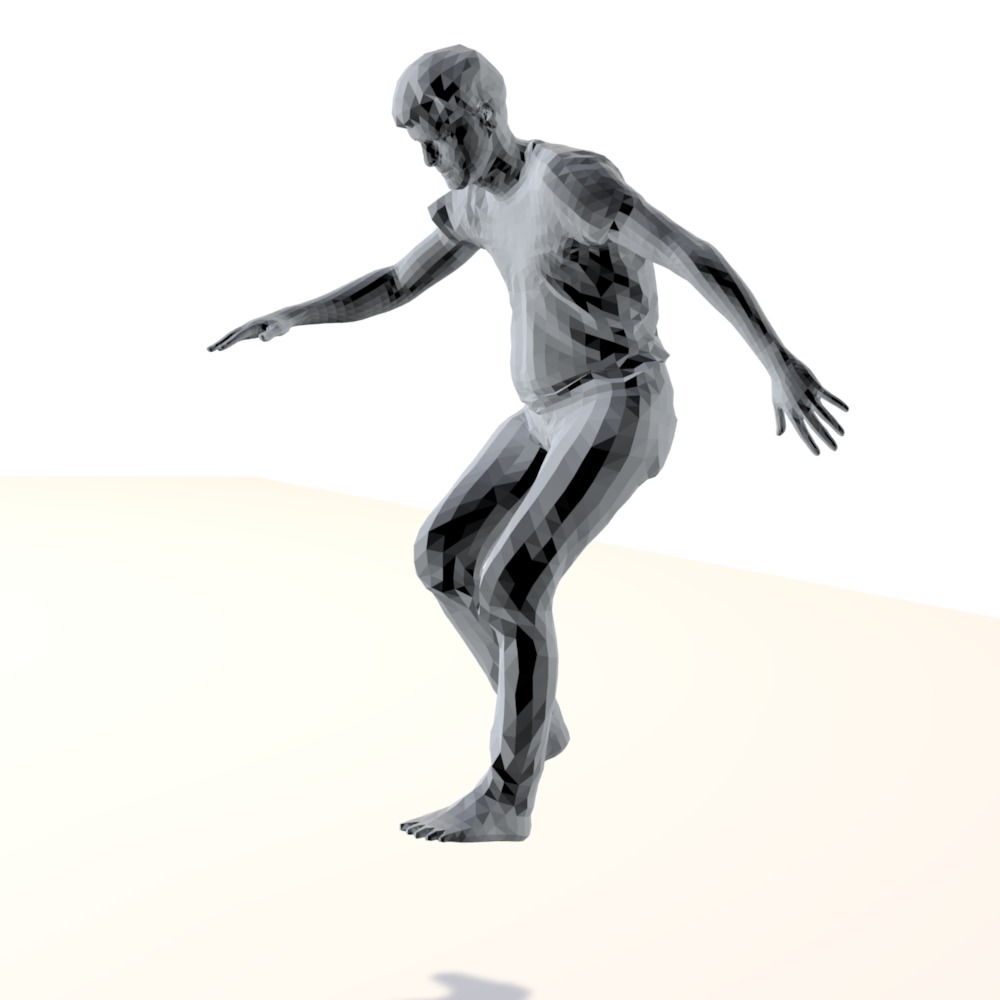} &
\includegraphics[width=0.14\linewidth]{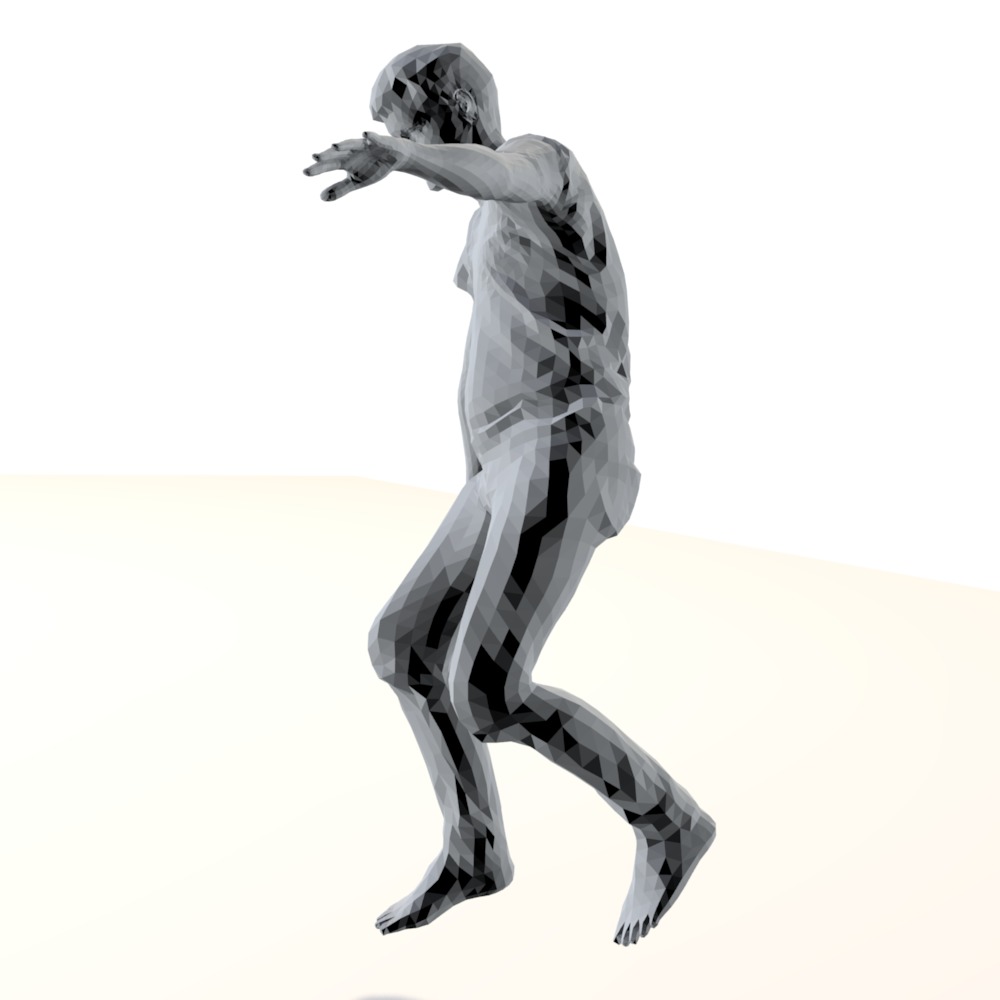}\\
\includegraphics[width=0.14\linewidth]{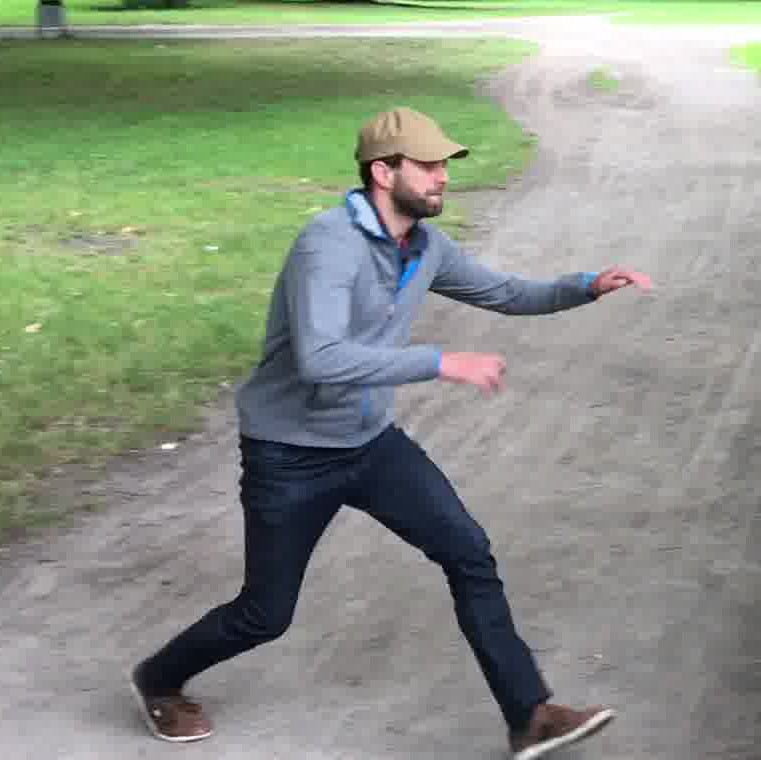} &
\includegraphics[width=0.14\linewidth]{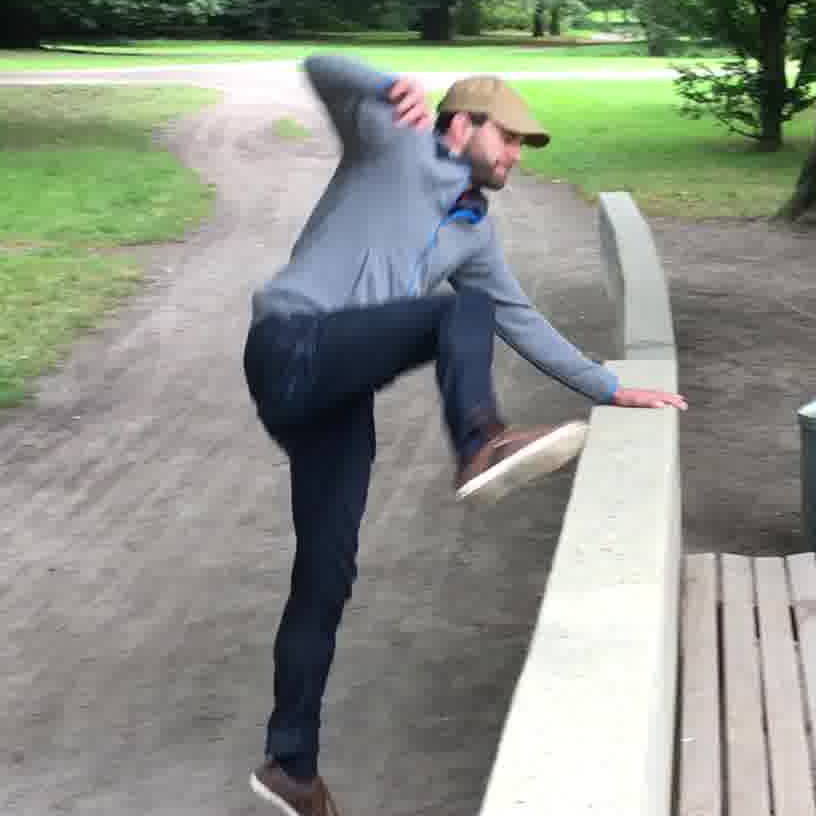} &
\includegraphics[width=0.14\linewidth]{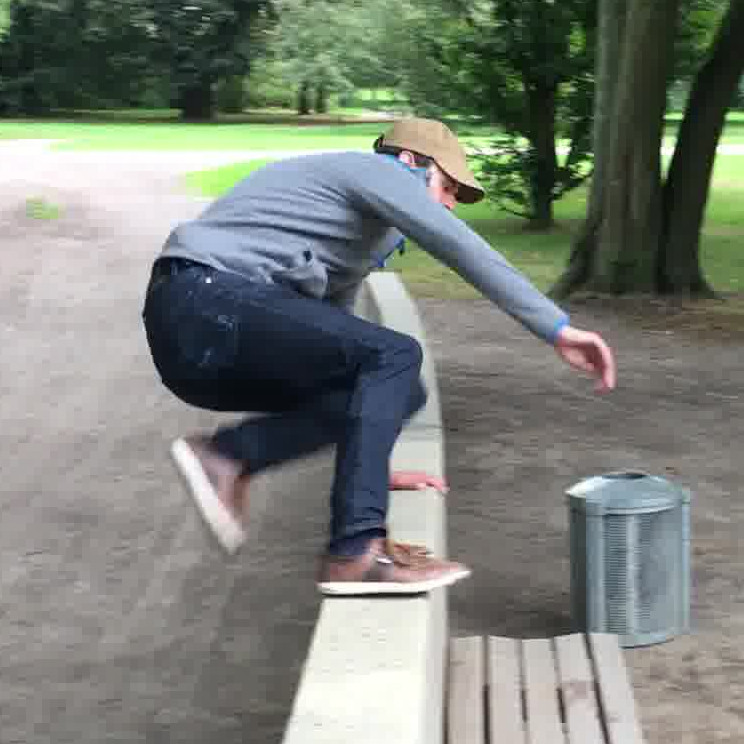} &
\includegraphics[width=0.14\linewidth]{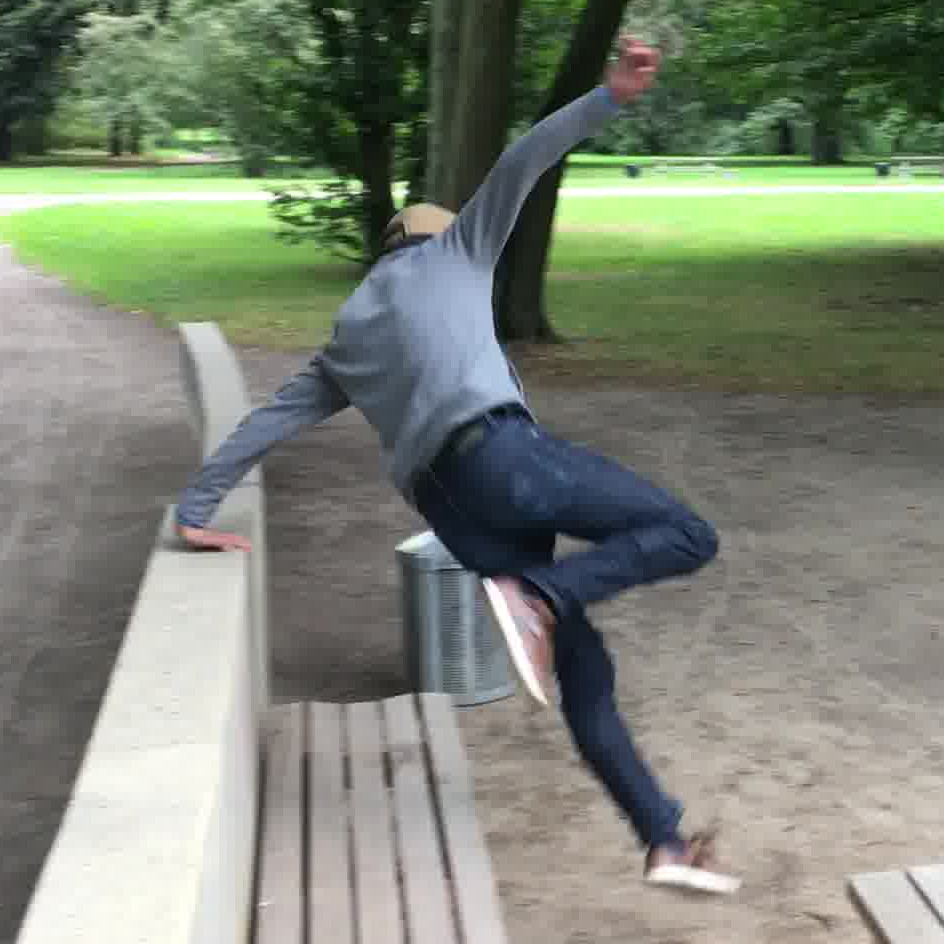} &
\includegraphics[width=0.14\linewidth]{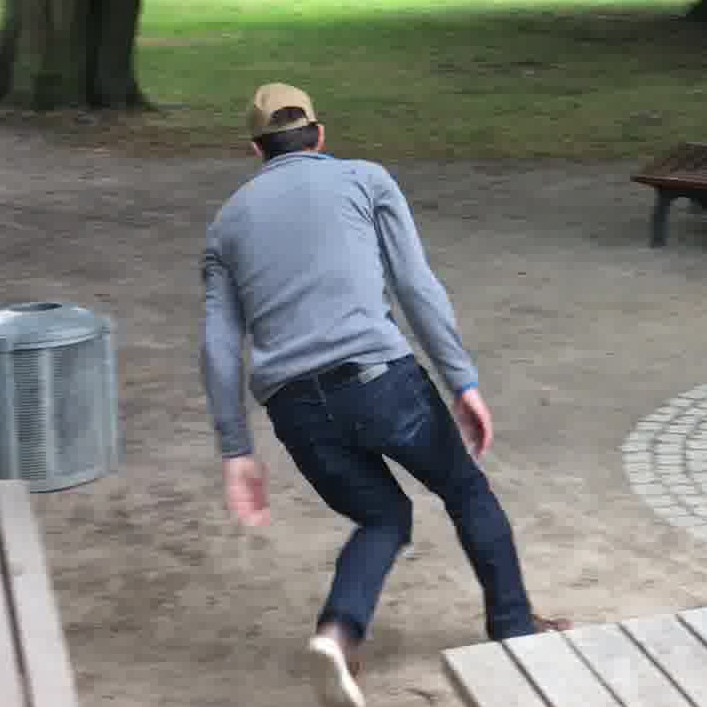} \\
\includegraphics[width=0.14\linewidth]{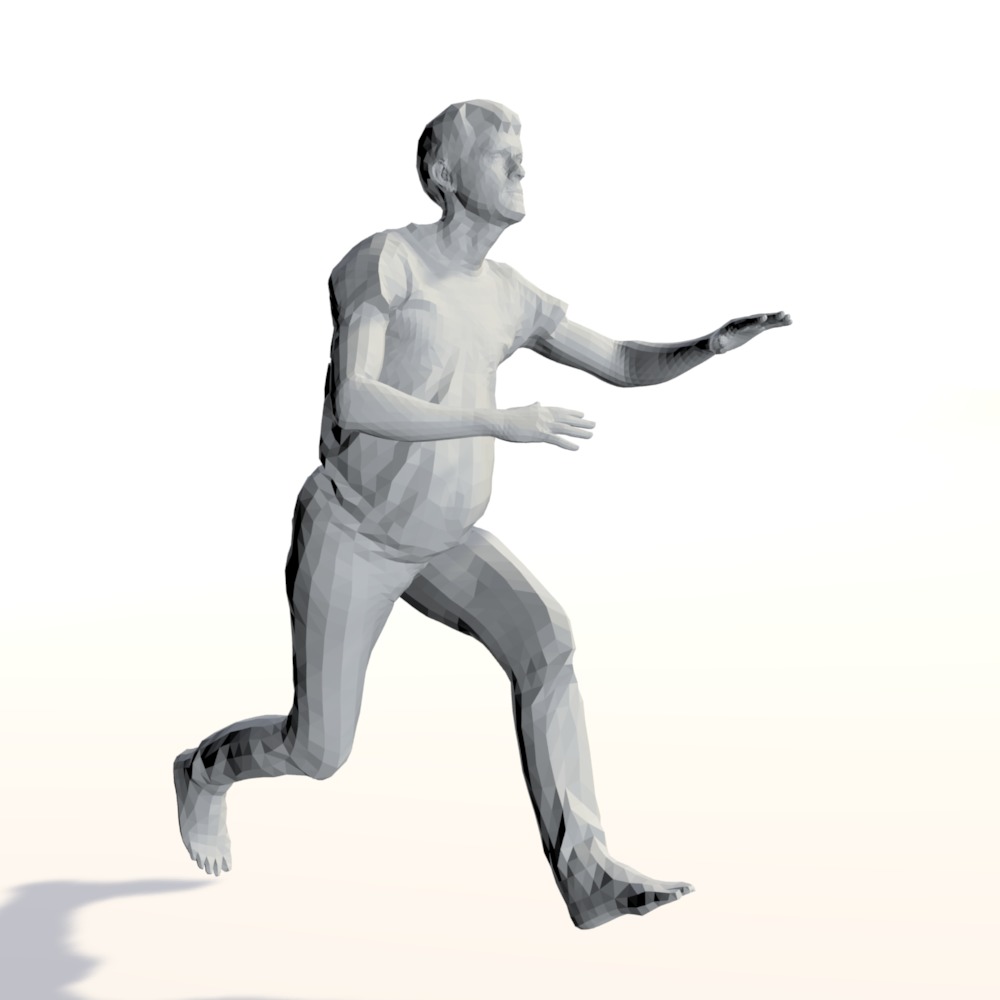} &
\includegraphics[width=0.14\linewidth]{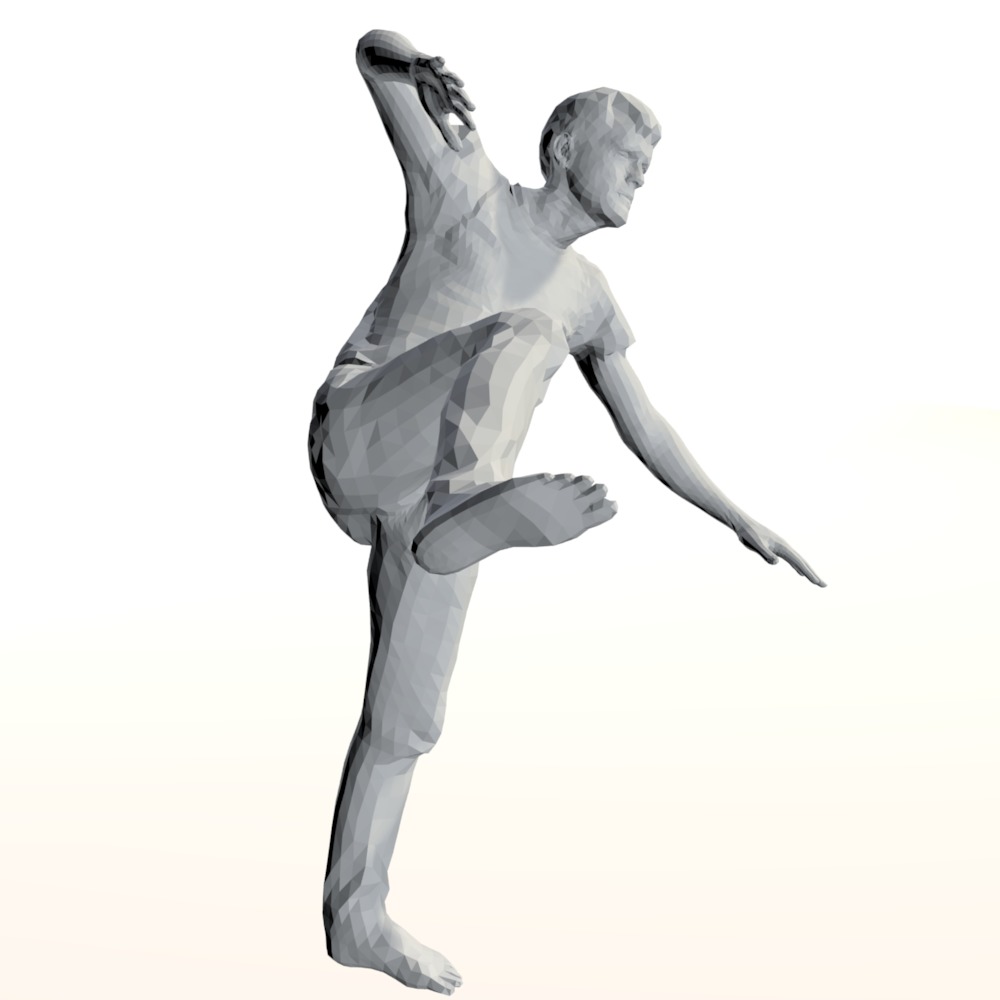} &
\includegraphics[width=0.14\linewidth]{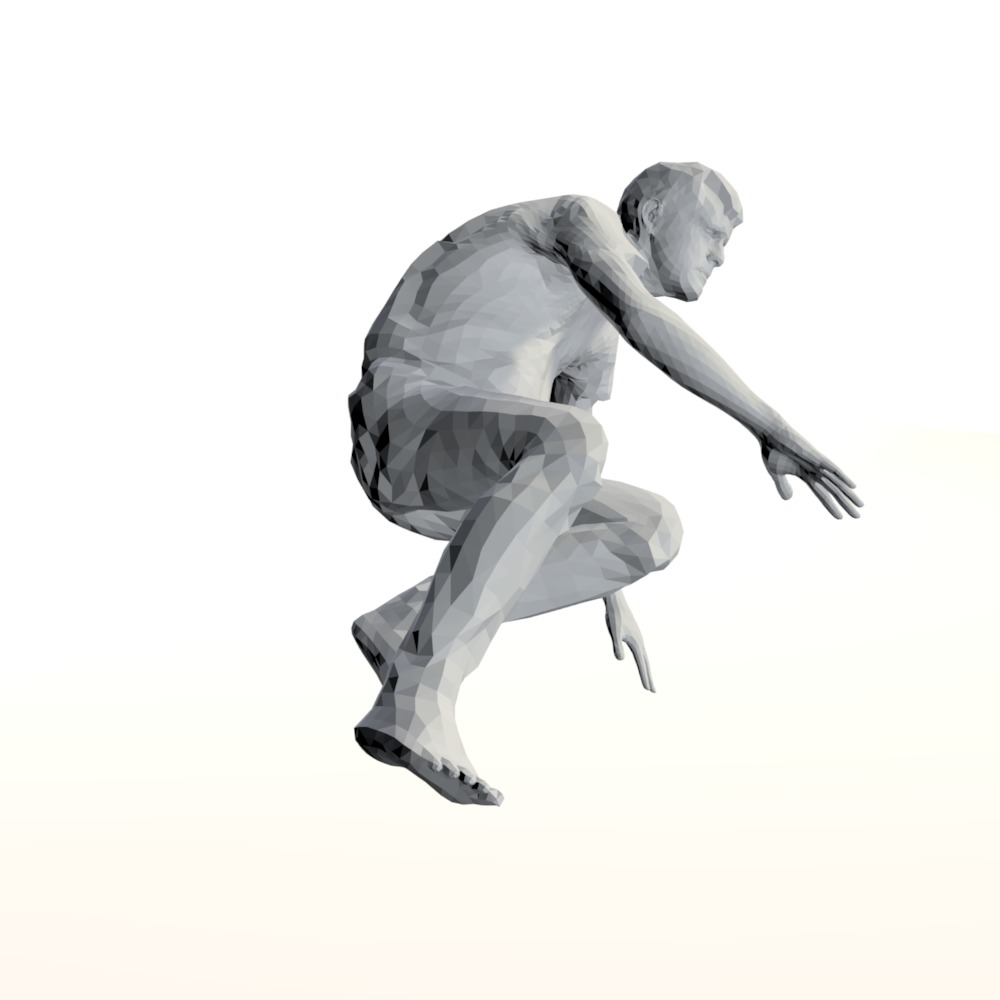} &
\includegraphics[width=0.14\linewidth]{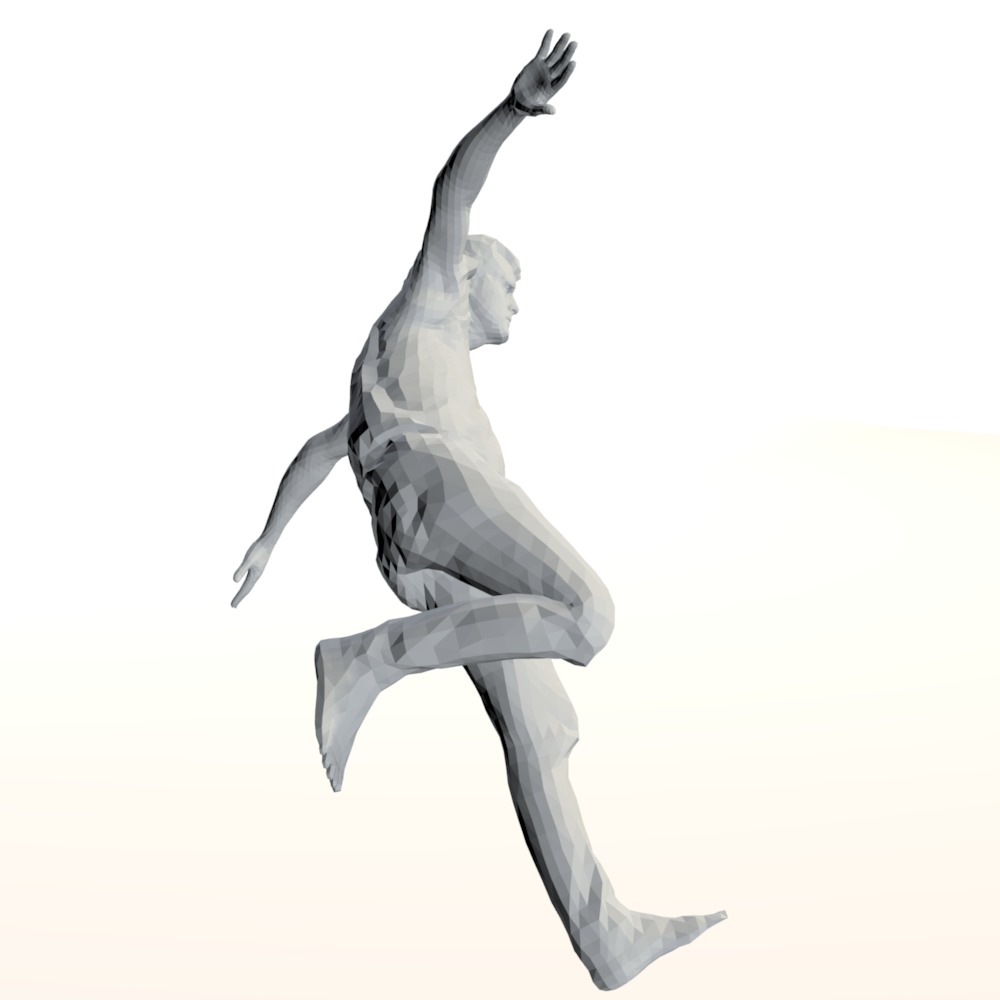} &
\includegraphics[width=0.14\linewidth]{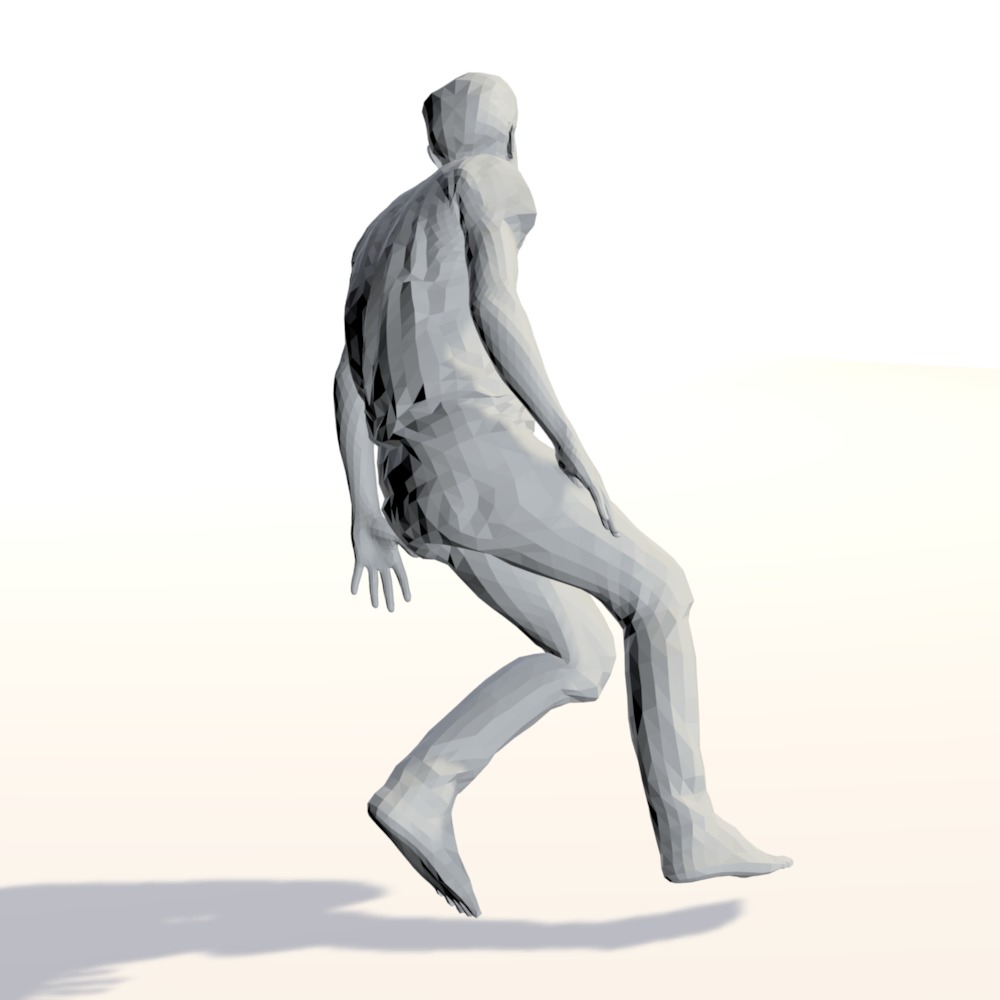} \\
\includegraphics[width=0.14\linewidth]{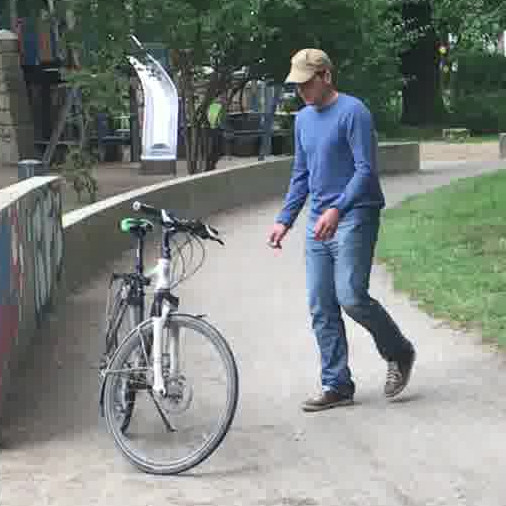} &
\includegraphics[width=0.14\linewidth]{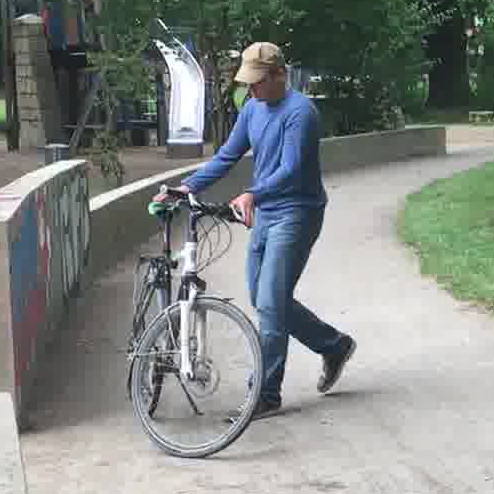} &
\includegraphics[width=0.14\linewidth]{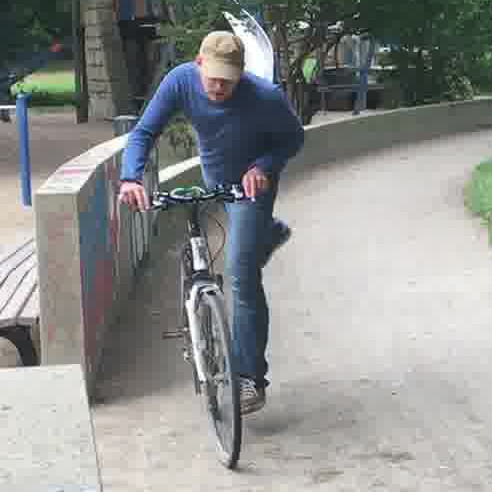} &
\includegraphics[width=0.14\linewidth]{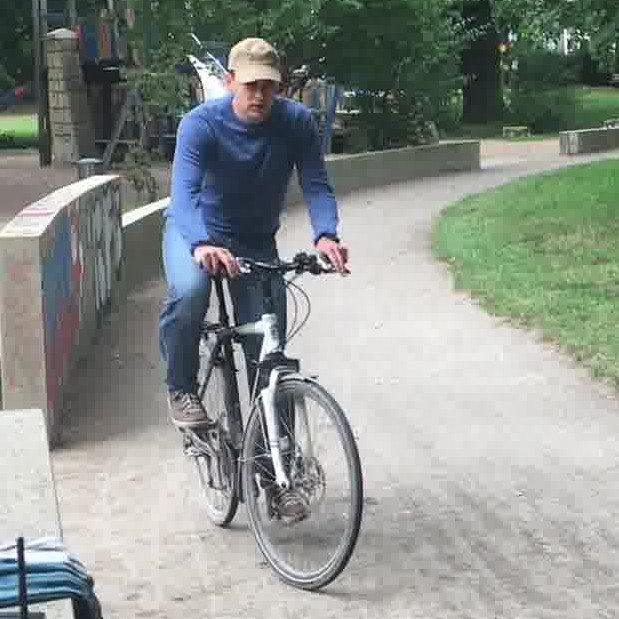} &
\includegraphics[width=0.14\linewidth]{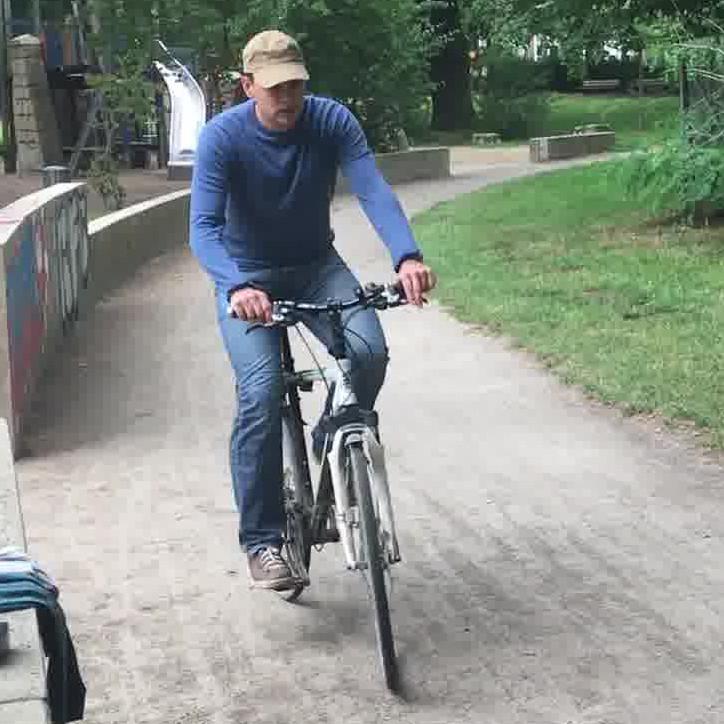} \\
\includegraphics[width=0.14\linewidth]{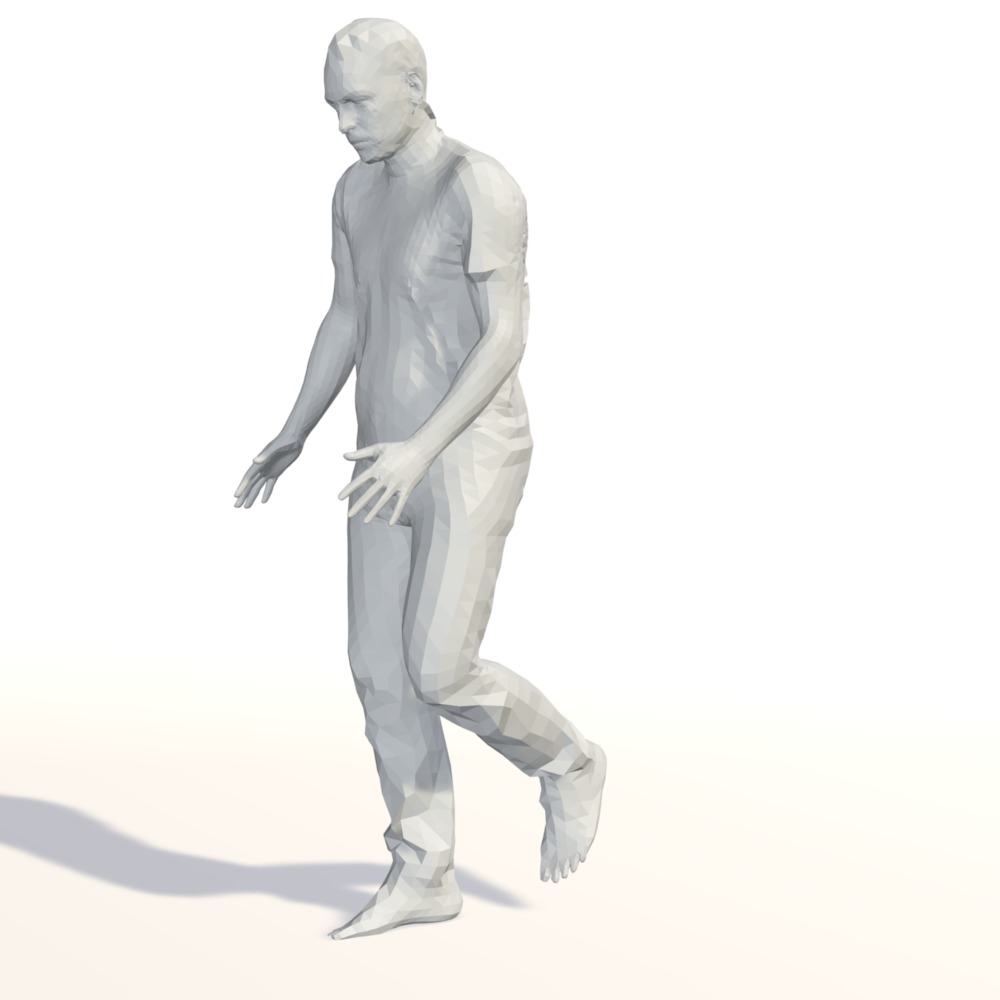} &
\includegraphics[width=0.14\linewidth]{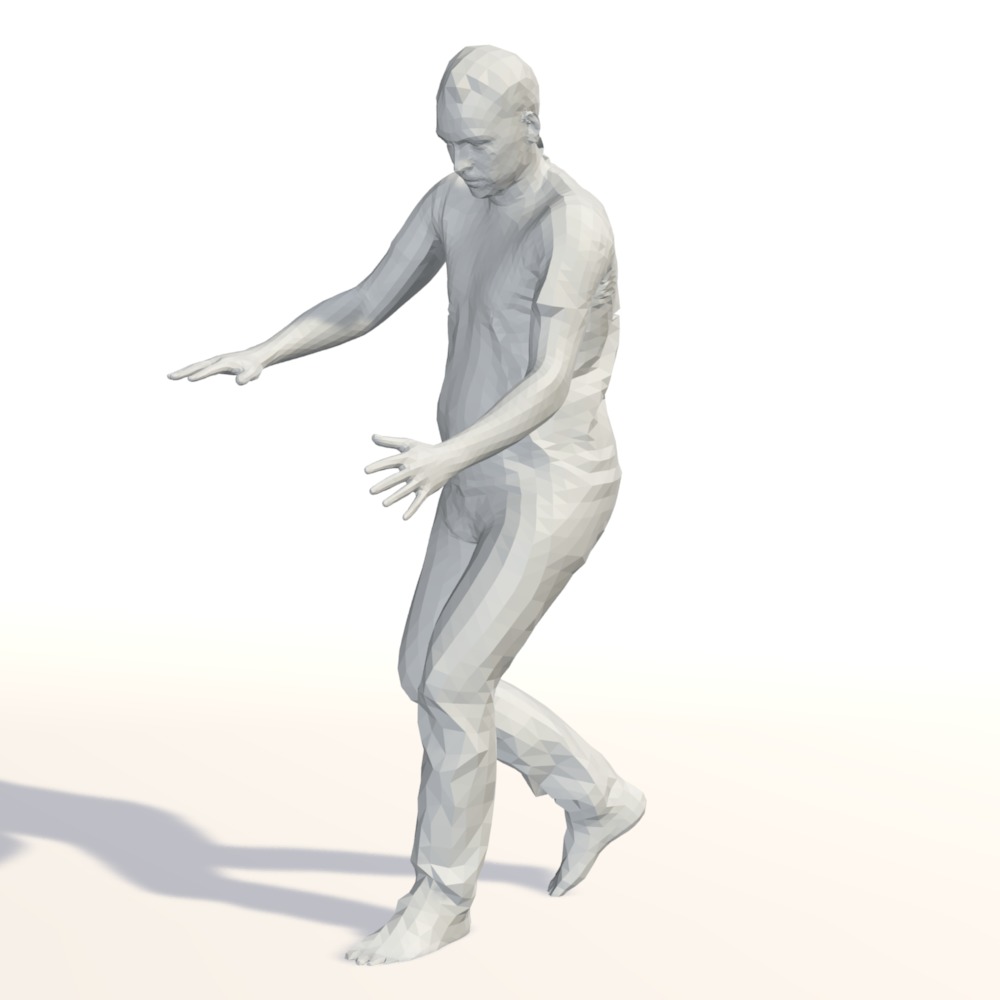} &
\includegraphics[width=0.14\linewidth]{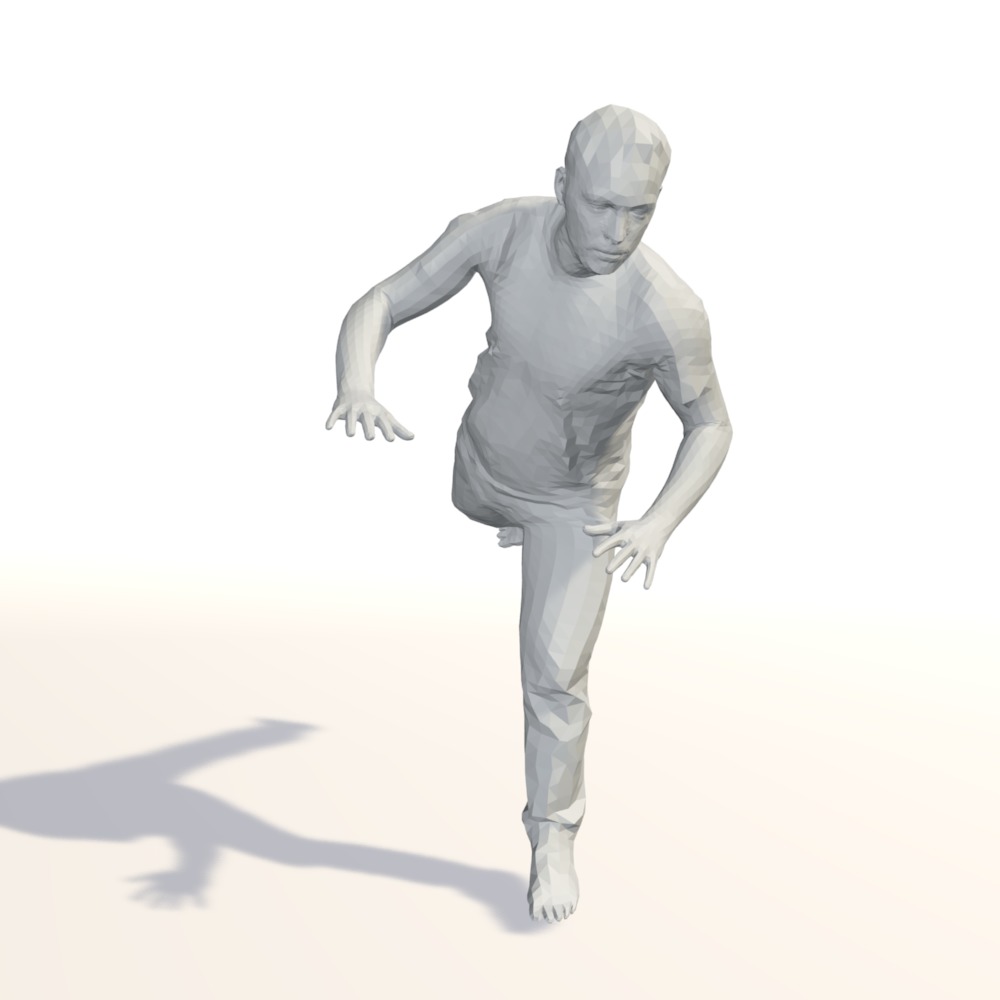} &
\includegraphics[width=0.14\linewidth]{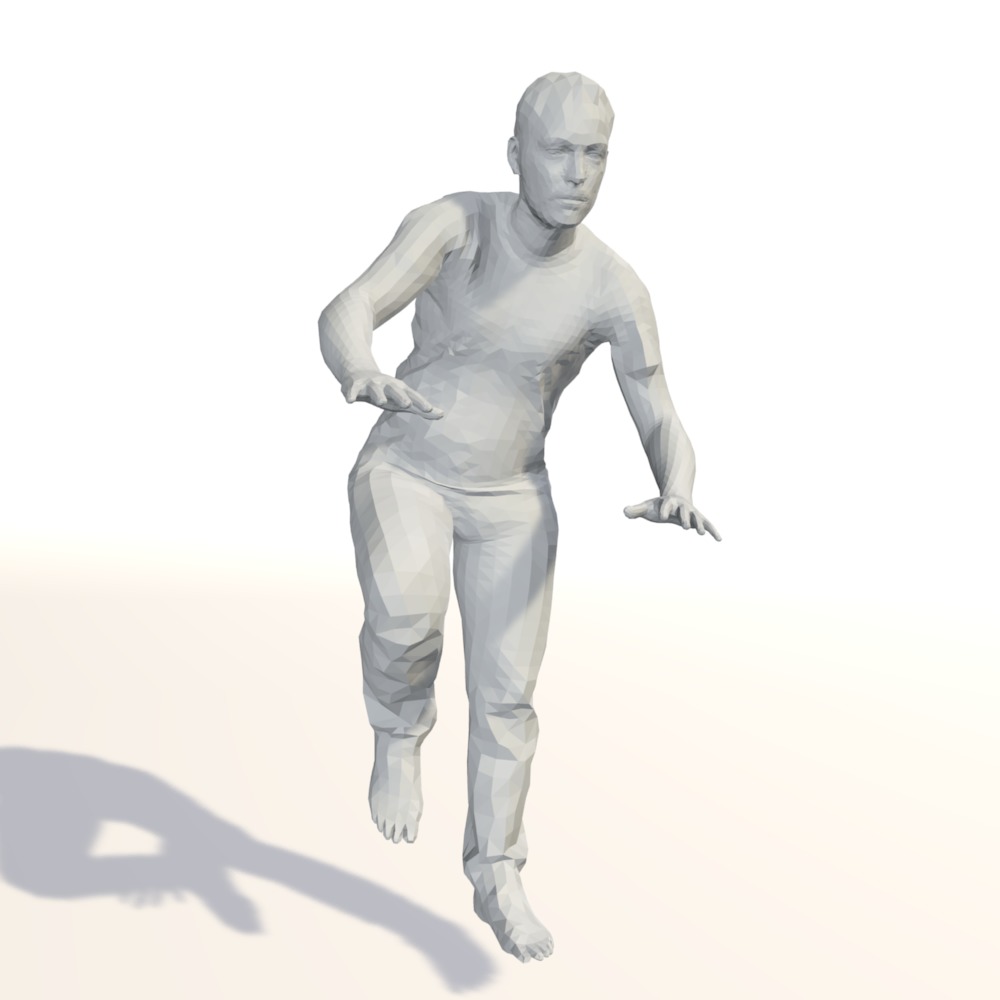} &
\includegraphics[width=0.14\linewidth]{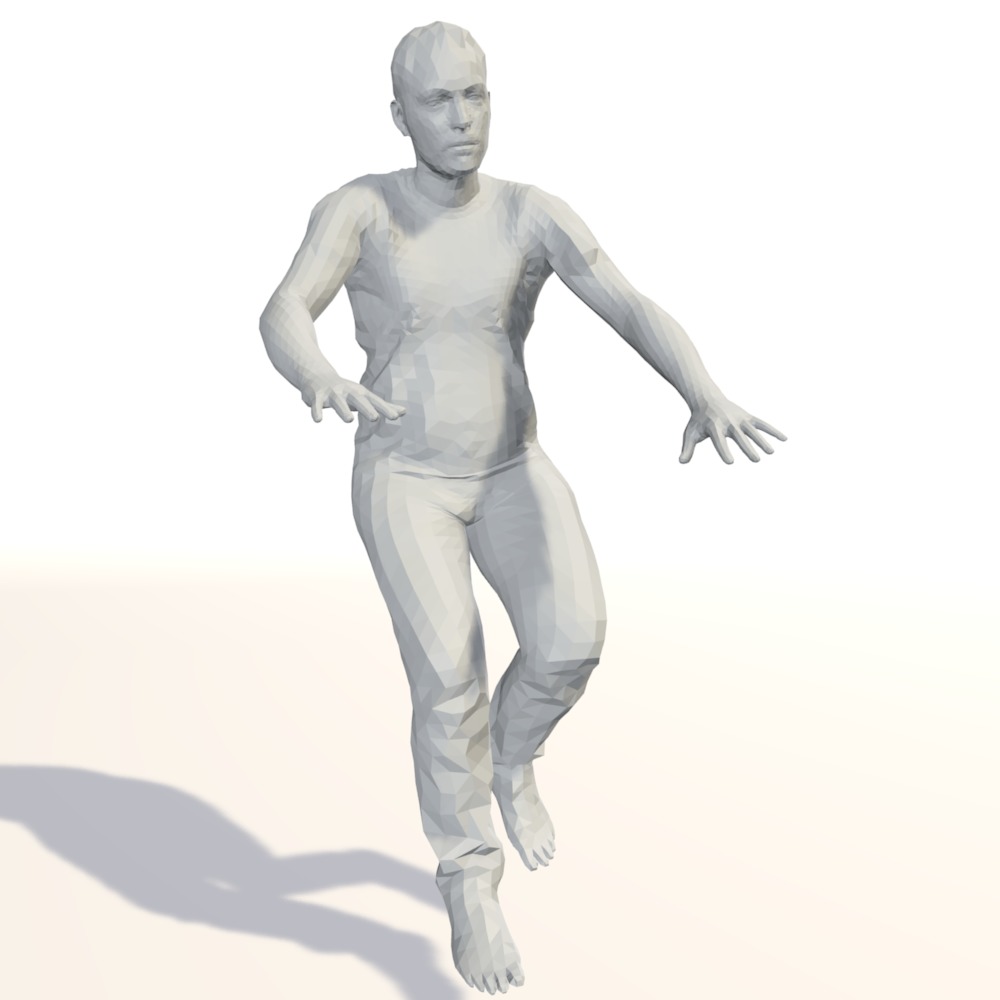} \\

\includegraphics[width=0.14\linewidth]{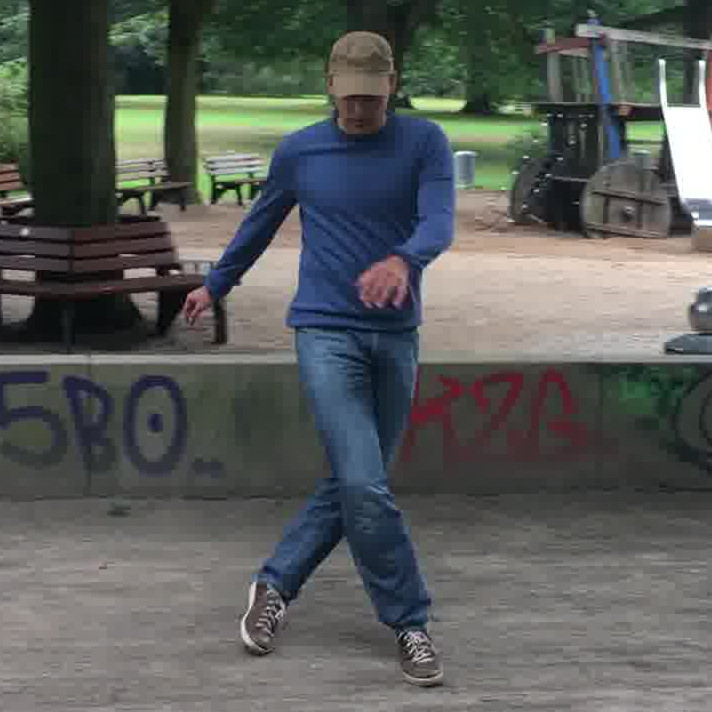} &
\includegraphics[width=0.14\linewidth]{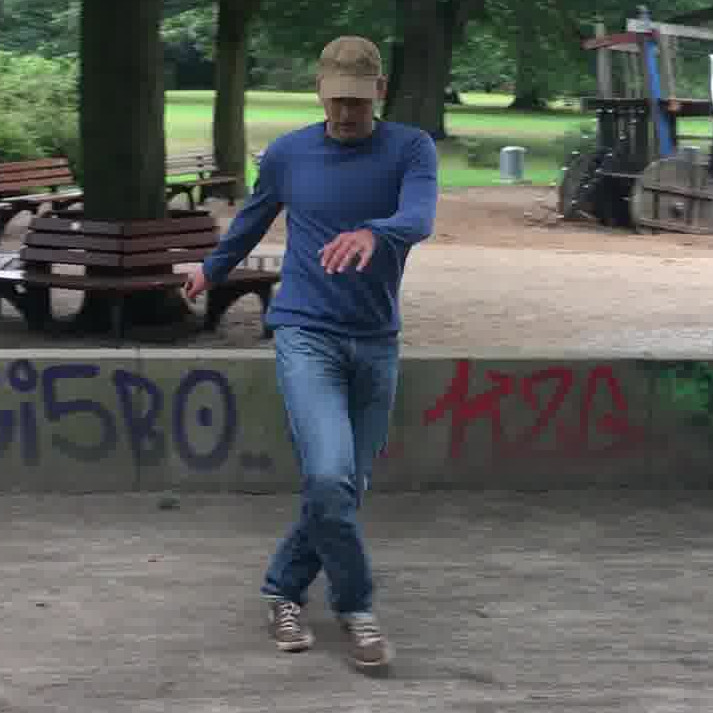} &
\includegraphics[width=0.14\linewidth]{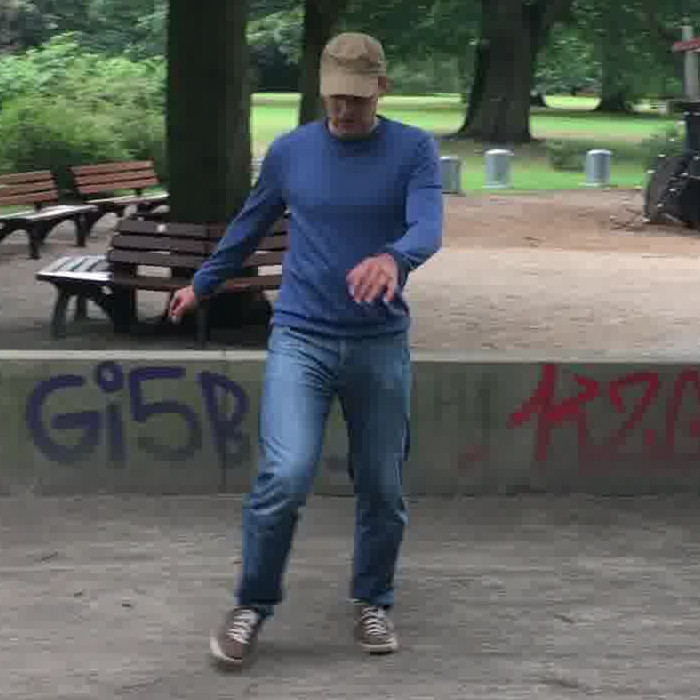} &
\includegraphics[width=0.14\linewidth]{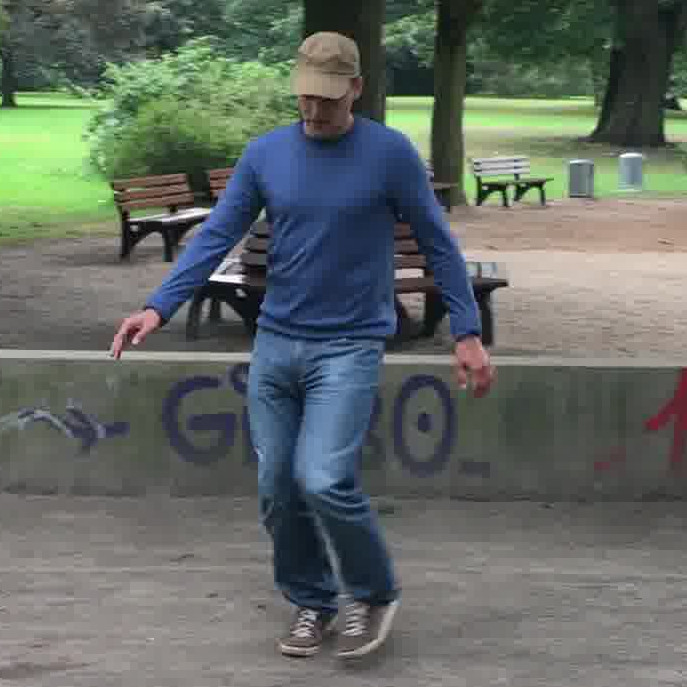} &
\includegraphics[width=0.14\linewidth]{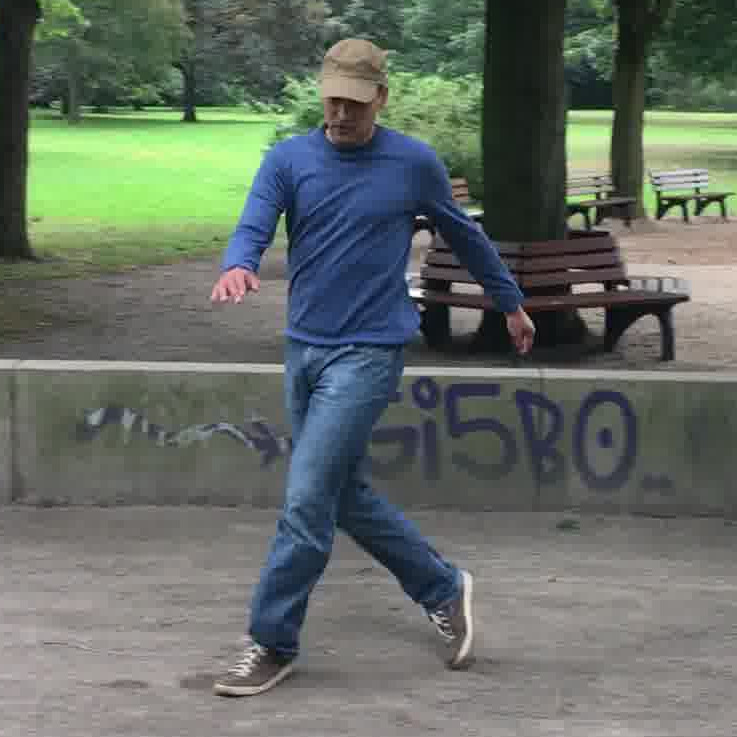}\\
\includegraphics[width=0.14\linewidth]{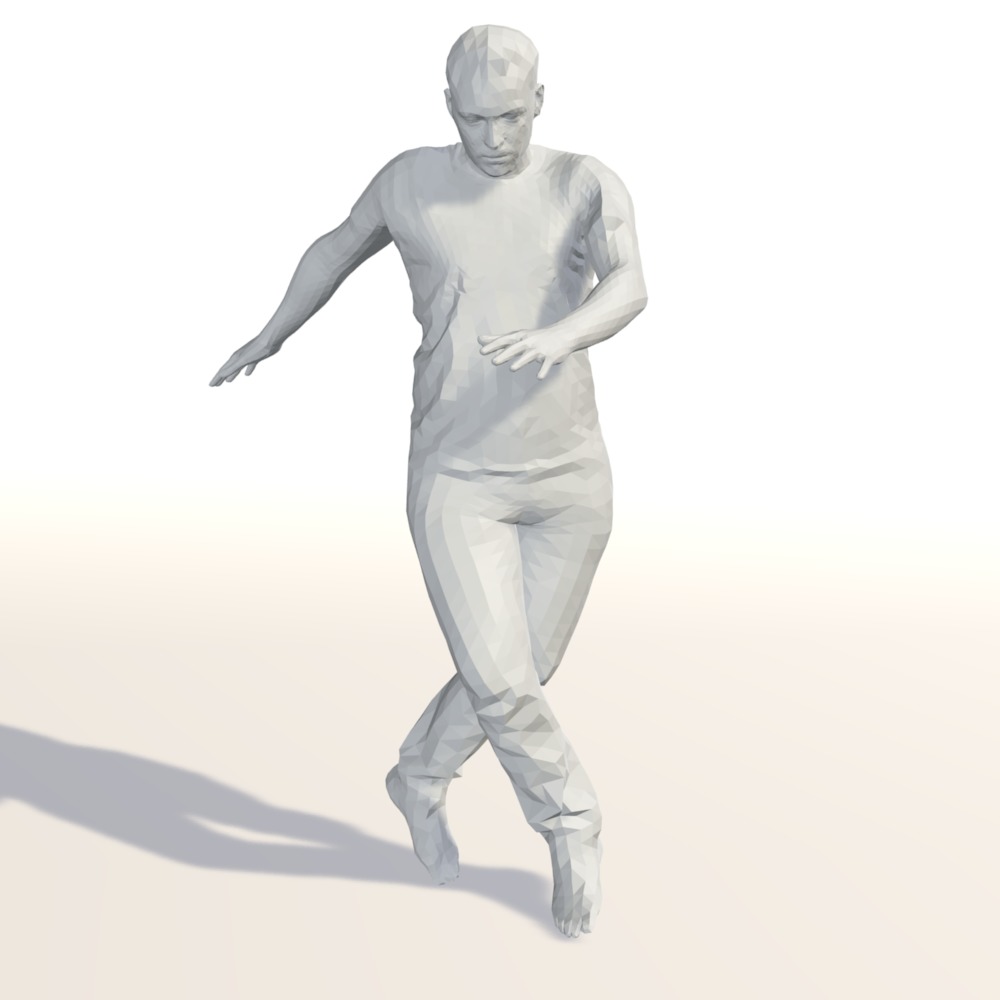} &
\includegraphics[width=0.14\linewidth]{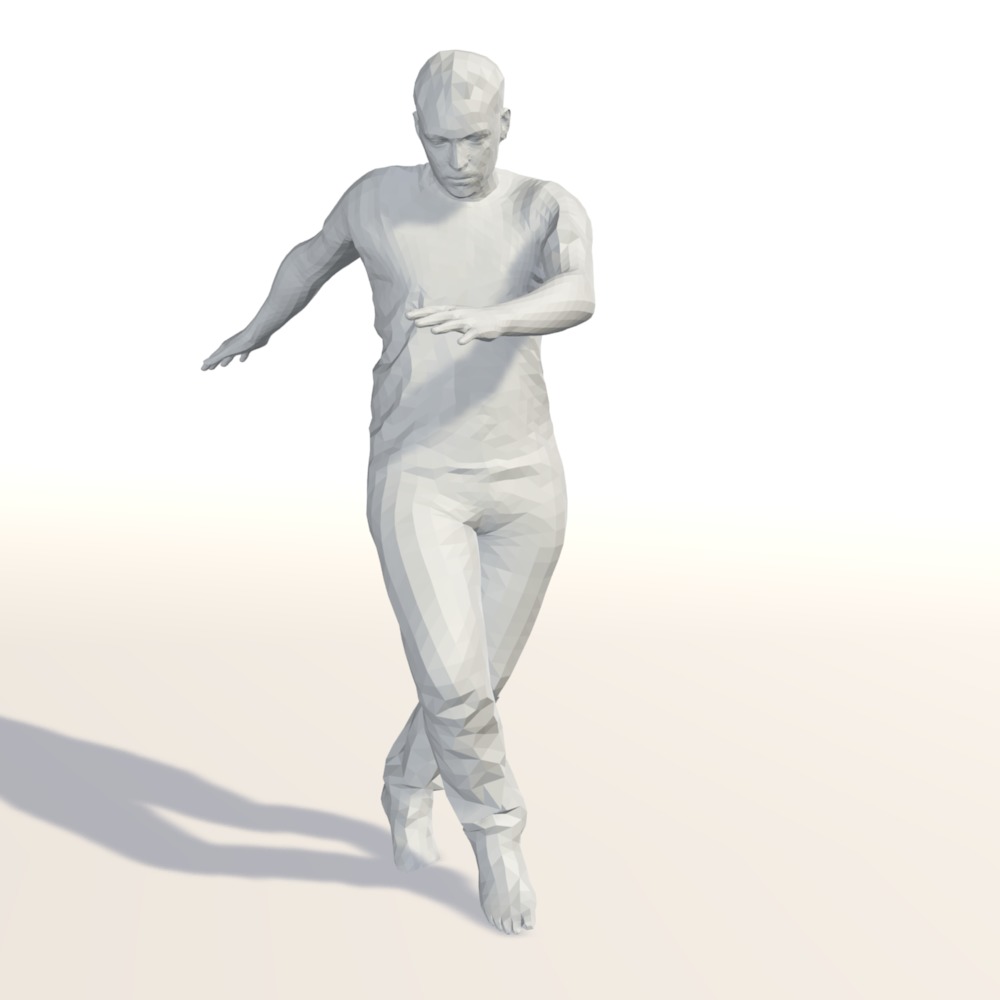} &
\includegraphics[width=0.14\linewidth]{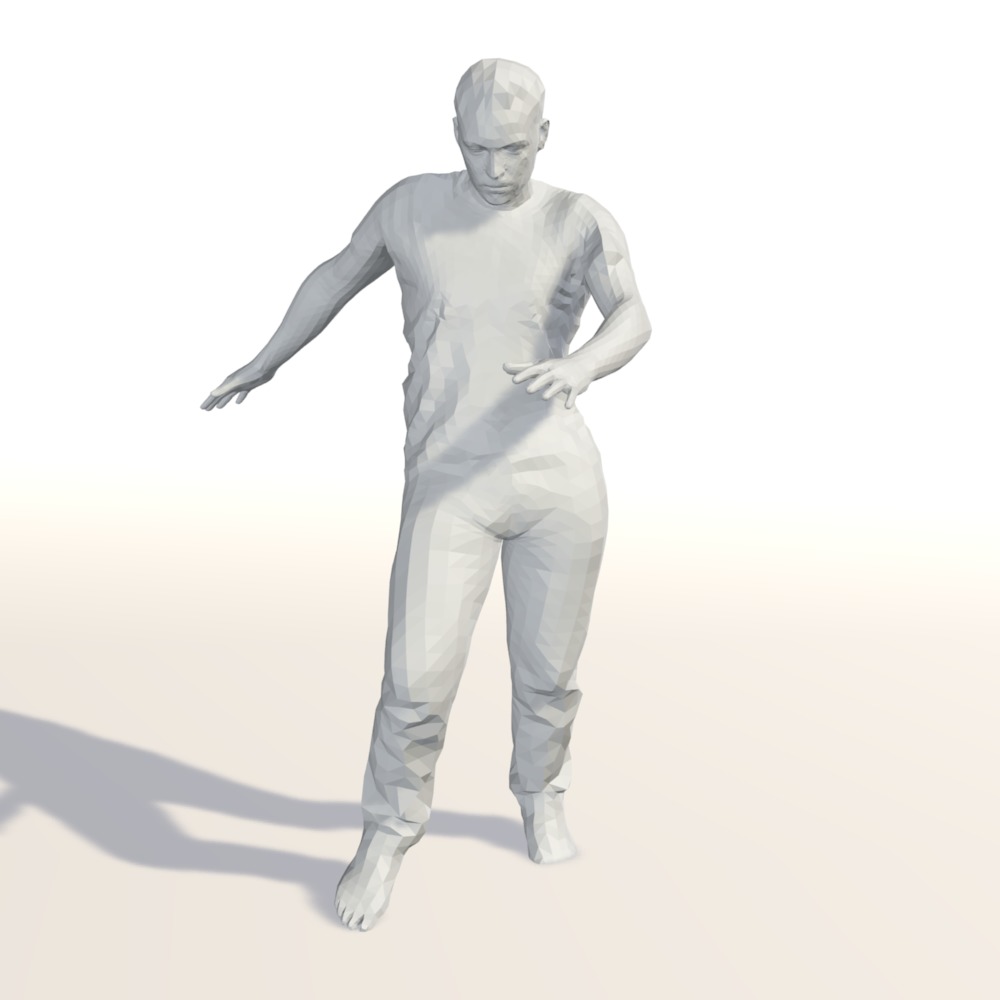} &
\includegraphics[width=0.14\linewidth]{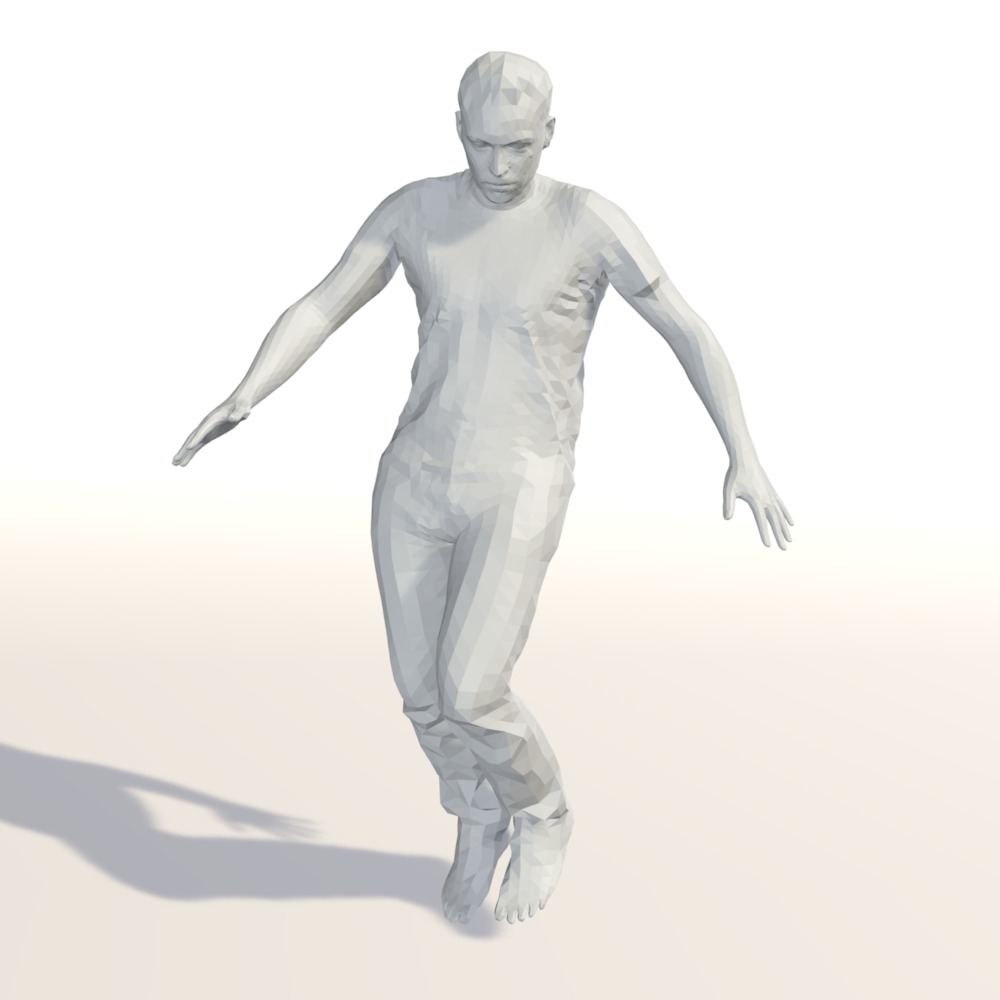} &
\includegraphics[width=0.14\linewidth]{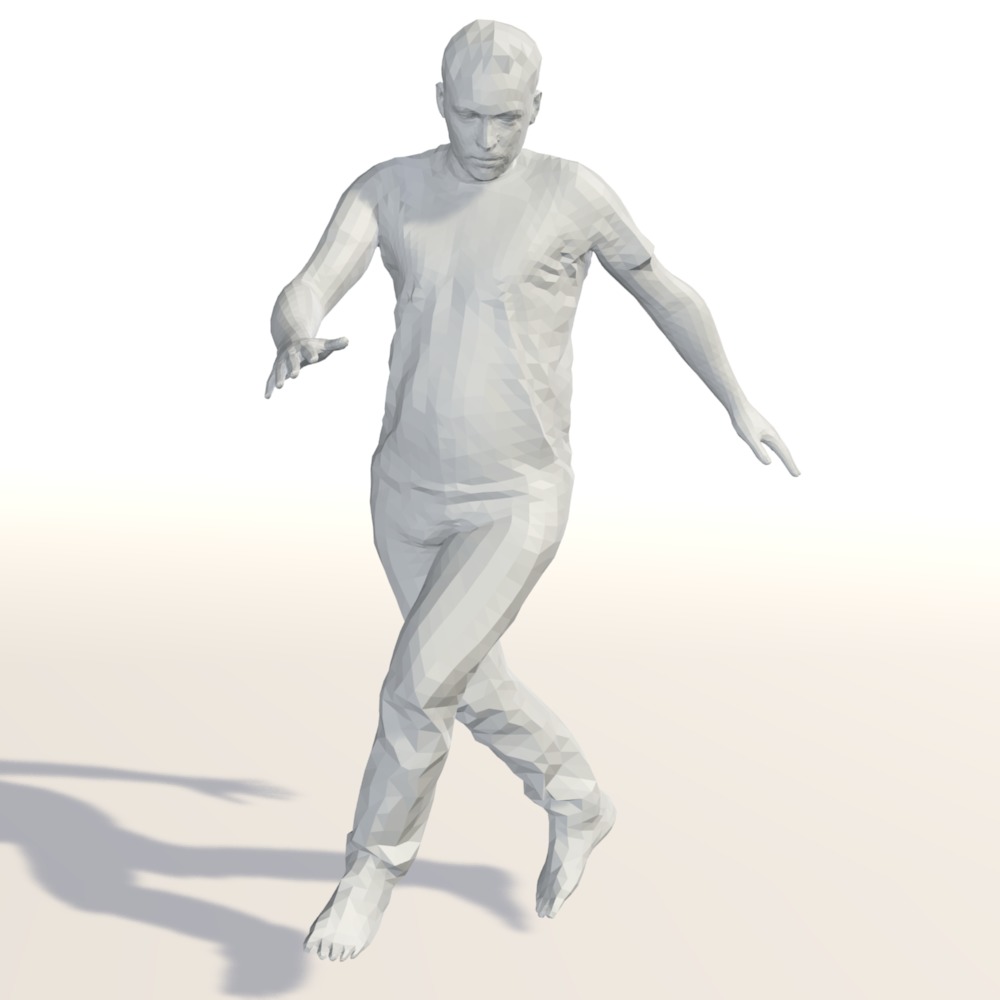} \\
\end{array}$
\end{center}
\caption{We show several results obtained using SIP: For most of the cases SIP successfully recovers the full human pose. This will enable to capture people performing everyday activities
in a minimally intrusive way. Results are best seen in the supplemental video.}
\label{fig:outdoorRecordings}
\end{figure*}
In \figref{fig:outdoorRecordings} we show several tracking results for challenging outdoor motions, such as jumping over a wall, warming exercises, biking and climbing. For all cases, our proposed SIP approach is able to successfully track the overall motion. For most of the cases, the recovered poses are visually accurate using only 6 IMUs.
Finally, in \figref{fig:whiteboardDrawing} we demonstrate that SIP is capable of reconstructing the handwriting on a whiteboard. For this experiment, we attached IMUs to the lower legs, wrists, back and chest and recorded IMU data while the actor was writing ``Eurographics'' on a white board. The resulting wrist motion clearly resembles the hand writing.
%
%
%
%
%
%
%
%

%% file: sections/conclusions.tex
SIP provides a new method for estimating the pose from sparse inertial sensors. SIP makes this possible by exploiting a
statistical body model and jointly optimizing pose over multiple frames to fit both orientation and acceleration data. 
We further demonstrate that the approach works even with approximate body models obtained from a few body word ratings. 
Quantitative evaluation shows that SIP can accurately reconstruct human pose accurately, with orientation errors of \added{13.32} degrees and positional errors of \added{3.9} cm.  

This technology opens up many directions for future research. 
\added{While SIP is able to track the full-body pose without drift, global position estimates still suffer from drift over time.} To that end, we plan to integrate simple physical constraints into the optimisation such as centre of mass preservation and ground contacts. \added{Exploiting laws of conservation of energies is very involved whereas modeling ground contacts is comparably easier: ground contacts produce high peaks in the accelerometer
signal which are easy to detect. Temporally fixing the position of body model points is straightforward to integrate in the proposed cost function and will compensate drift. However, modeling ground contacts depends on the motion to be tracked and assumes static friction \cite{andrews2016real}. Other options we will explore to compensate drift are integrating GPS measurements 
(e.g.~from a cell carried phone on the body), or visual data from a body mounted camera~\cite{rhodin2016egocap,shiratori2011motion}}.

Our current solution can not accurately capture \added{wrist and ankle joint parameters due to the IMU placement on the body, see \figref{fig:sensorPlacement} and \sectref{sec:imuPlacement}. While these unobserved parameters are also optimized within the anthropometric prior, we plan to} incorporate constraints derived from the 3D world geometry. \added{Also, instead of using static joint limits in the anthropometric term one could also incorporate pose-conditioned joint angle limits \cite{akhter2015pose} to obtain physically plausible poses.} We \added{further} plan to learn human motion models from captured data in every day situations.

\added{Finally, we would like to extend SIP to capture not only articulated motion but also soft-tissue motion by leveraging models
of human shape in motion such as \cite{Dyna}.}
SIP provides the technology to capture human motion with as few as 6 IMUs which is much less intrusive than existing technologies. 
There are many potential applications for this such as virtual reality, sports analysis, monitoring for health assessment, or recording of 
movement for psychological and social studies.\\

%% file: sections/acknowledgments.tex
\newpage
\textbf{Acknowledgments.} This work is partly funded by the DFG-Project RO 2497/11-1. Authors gratefully acknowledge the support. We thank Timo Bolkart,
Laura Sevilla, Sergi Pujades, Naureen Mahmood, Melanie Feldhofer and Osman Ulusoy for proofreading, Bastian Wandt and Aron Sommer for help with motion recordings,
Talha Zaman for voice recordings, Alejandra Quiros for providing the bodies from words and Senya Polikovsky, Andrea Keller and Jorge Marquez for technical support.